\documentclass[10pt,twocolumn,letterpaper]{article}
\usepackage[accsupp]{axessibility}  % Improves PDF readability for those with disabilities.
\usepackage[pagenumbers]{cvpr} % To force page numbers, e.g. for an arXiv version

\usepackage{graphicx}
\usepackage{amsmath}
\usepackage{amssymb}
\usepackage{booktabs}

\usepackage[pagebackref,breaklinks,colorlinks]{hyperref}

\usepackage{times}
\usepackage{epsfig}

\usepackage{url}            % simple URL typesetting
\usepackage{amsfonts}       % blackboard math symbols
\usepackage{nicefrac}       % compact symbols for 1/2, etc.
\usepackage{microtype}      % microtypography
\usepackage{caption}
\usepackage{color}

\usepackage{algorithm}
\usepackage{algorithmic}
\usepackage{array}
\usepackage{enumitem}
\newtheorem{defn}{Definition}

\newcommand{\TODO}[1]{\textcolor{blue}{}\textcolor{blue}{\emph{#1}}}

\newcommand{\x}[1]{\mathbf{x}_{#1}}

\newcommand{\Nx}[1]{\hat{\mathbf{x}}_{#1}} %normalized output
\def\eg{\emph{e.g.}}
\def\Eg{\emph{E.g.}}
\def\ie{\emph{i.e.}}

\def\vs{\emph{vs}}
\def\wrt{w.r.t.}

\def\etal{\emph{et al.}}
\def\SM{Appendix}
\def\ES{estimation shift}
\def\ESM{estimation shift magnitude}

\def\cvprPaperID{5015} % *** Enter the CVPR Paper ID here
\def\confName{CVPR}
\def\confYear{2022}

\begin{document}

%\title{beyond normalization module towards neural architecture design}
%\title{An Investigation into the Estimation Shift of Batch Normalization}
%\title{Towards Relieve Estimation Shift of BN in Neural Architecture Design}
\title{Delving into the Estimation Shift of Batch Normalization in a Network}

\author{Lei Huang$^{1}$\thanks{Corresponding author. E-mail: \textit{huangleiAI@buaa.edu.cn}} \quad Yi Zhou$^{2}$ \quad  Tian Wang$^{1}$ \quad Jie Luo$^{1}$ \quad  Xianglong Liu$^{1}$\\
	$^{1}$SKLSDE, Institute of Artificial Intelligence,  Beihang University, Beijing, China\\
	$^{2}$MOE Key Laboratory of Computer Network and Information Integration, Southeast University, China\\
	%{\tt\small $\{$huanglei, xlliu, blonster, langbo$\}$@nlsde.buaa.edu.cn, dacheng.tao@sydney.edu.au}
%$^{2}$University of Electronic Science and Technology of China, Chengdu, China\\
%	{\tt\small $\{$lei.huang, yi.zhou,  fan.zhu, li.liu, ling.shao$\}$ @inceptioniai.org}
}

%% omit the following lines up until the closing ``}''.
%% Additional authors and addresses can be added with ``\and'',
%% just like the second author.
%% To save space, use either the email address or home page, not both
%\and
\maketitle
%\thispagestyle{empty}
%%%%%%%%% ABSTRACT
\begin{abstract}
Batch normalization (BN) is a milestone technique in deep learning. It normalizes the activation using mini-batch statistics during training but the estimated population statistics during inference. This paper focuses on investigating the estimation of population statistics. We define the estimation shift magnitude  of BN to quantitatively measure the difference between its estimated population statistics and expected ones. Our primary observation is that the estimation shift can be accumulated due to the stack of BN in a network, which has detriment effects for the test performance. We further find a batch-free normalization (BFN) can block such an accumulation of estimation shift. These observations motivate our design of XBNBlock that replace one BN with BFN in the bottleneck block of residual-style networks. Experiments on the ImageNet and COCO benchmarks show that XBNBlock consistently improves the performance of different architectures, including ResNet and ResNeXt, by a significant margin and seems to be more robust to distribution shift. 
\end{abstract}

%the expected population statistics of BN and the estimation shift that is ill-defined and define its expected onquantitatively
%The  inconsistent operations of BN potentially affects its performance in varies of scenarios.   This paper seek to quantitatively investigate the estimation of populations statistics. 
%We investigate how the depth of a network affects the estimation of BN.bWe design new architectures to improve the performance. We proposed XBN-BLock significant improve the performance of the ResNets. 
%%%%%%%%% BODY TEXT
\section{Introduction}
\label{sec:intro}
Input normalization is extensively used in training neural networks for decades~\cite{1998_NN_LeCun} and shows good theoretical properties in optimization for linear models~\cite{1990_NeurIPS_LeCun,2011_NIPS_Wiesler}. 
It uses population statistics for normalization  that can be  calculated directly from the available training data.  A natural idea is to extend normalization for the activation in a network. However, normalizing  activation is more challenging since the distribution of internal activation varies, which leads to the estimation of population statistics for normalization inaccurate~\cite{2012_NN_Gregoire,2015_NeurIPS_Desjardins}.  A network with activation normalized by the population statistics shows the  training instability~\cite{2020_arxiv_Huang}.

Batch normalization (BN)~\cite{2015_ICML_Ioffe} addresses itself to normalize the activation using mini-batch statistics during training, but the estimated population statistics during inference/test. 
BN ensures the normalized mini-batch output standardized over each iteration, enabling stable training, efficient optimization ~\cite{2015_ICML_Ioffe,2016_CoRR_Ba,2018_NIPS_shibani,2018_CVPR_Huang} and potential generalization~\cite{2018_CVPR_Huang,2018_NIPS_Bjorck,2018_ECCV_Wu}. It has been extensively used in varieties of architectures~\cite{2015_CVPR_He,2015_CoRR_Szegedy,2016_CoRR_Zagoruyko,2016_CoRR_Szegedy,2016_CoRR_Huang_a,2017_CVPR_Xie}, and successfully proliferated throughout various areas~\cite{2015_IJCV_ImageNet,2014_ECCV_COCO,2020_arxiv_Huang}. 

\begin{figure}[t]
	\centering
	\vspace{-0.08in}
	\hspace{-0.15in}	\subfloat[Network with BN in each layer]{
		\begin{minipage}[c]{1.0\linewidth}
			\centering
			\includegraphics[width=7.8cm]{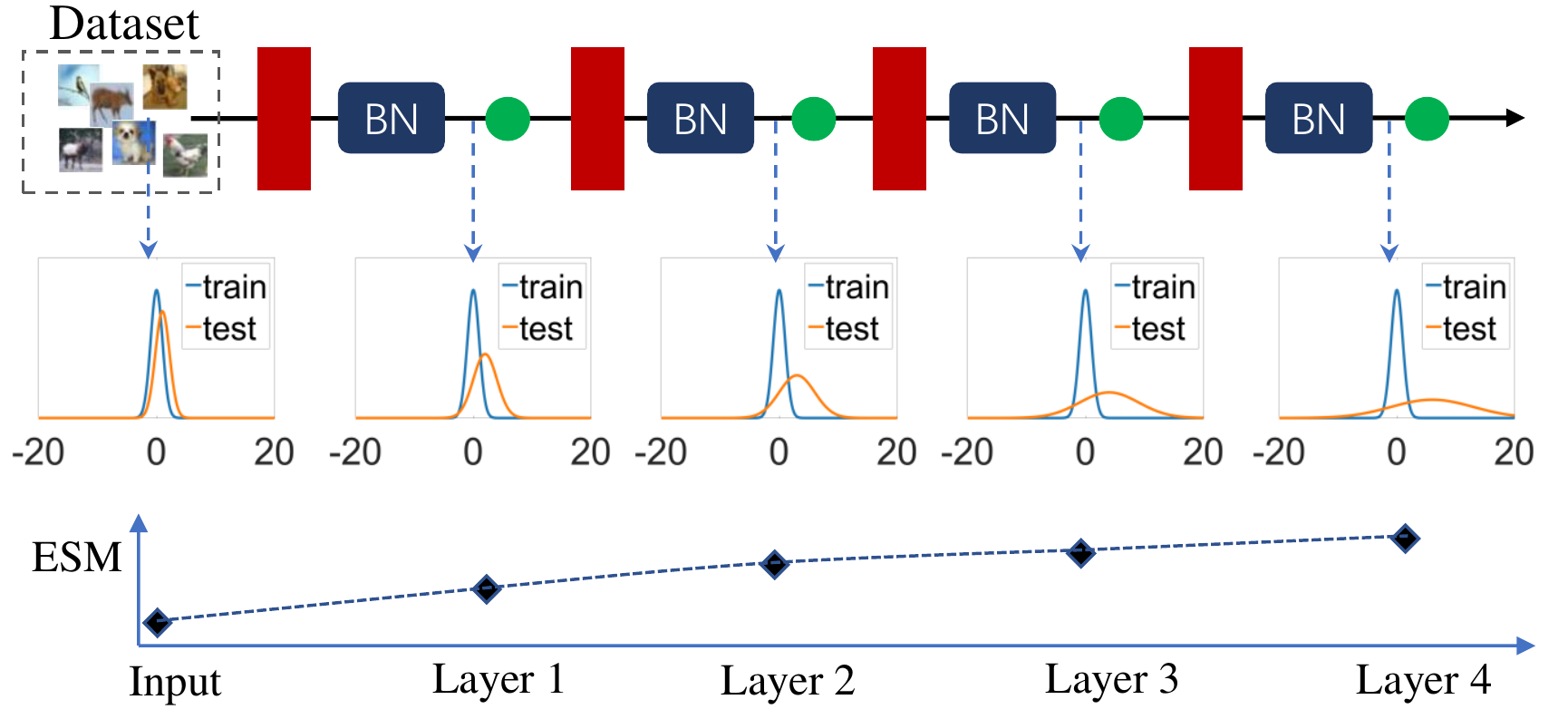}
		\end{minipage}
	}\\
	\hspace{-0.15in}		\subfloat[Network with BN and BFN mixed in different layers]{
		\begin{minipage}[c]{1.0\linewidth}
			\centering
			\includegraphics[width=7.8cm]{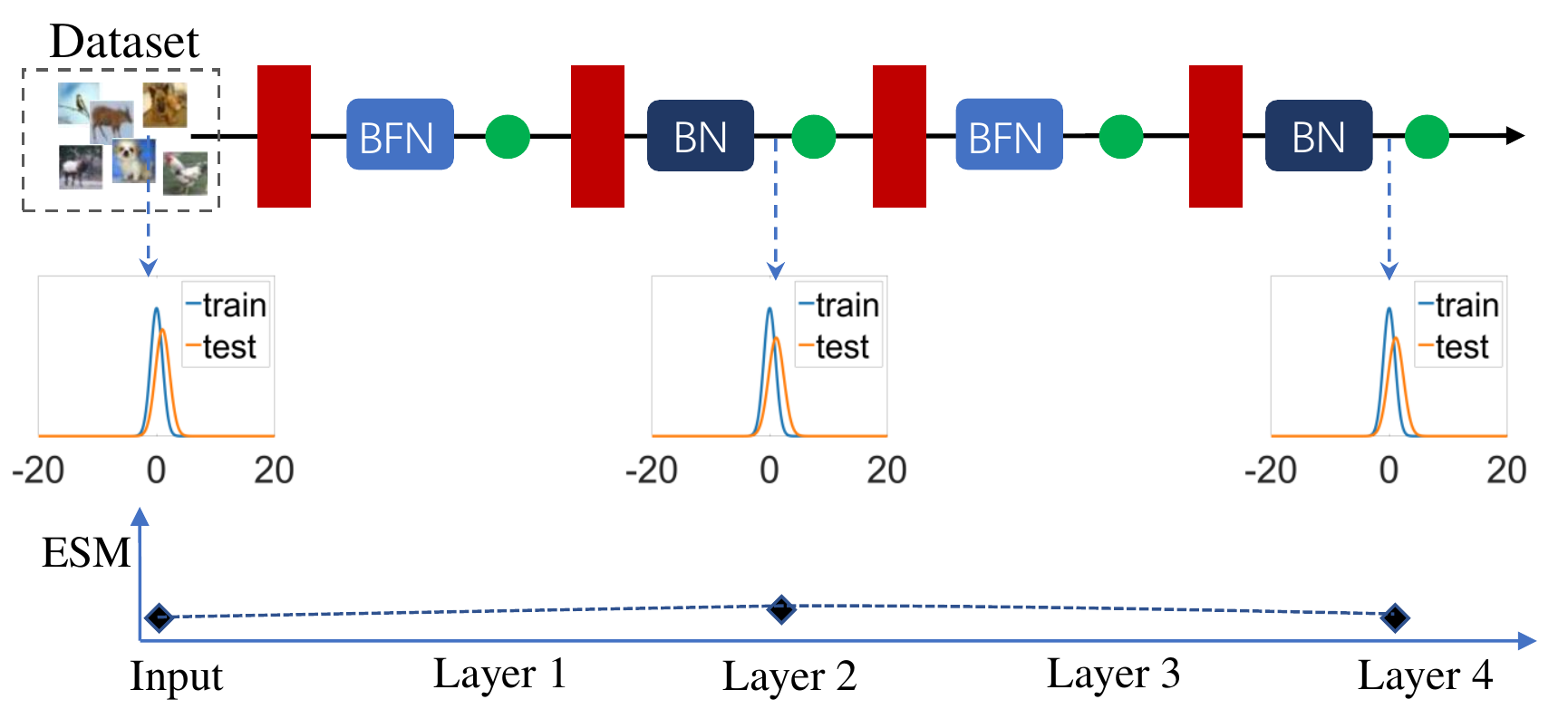}
		\end{minipage}
	}
 \vspace{-0.1in}
	\caption{Illustration of the main observations. The red rectangle and green round represent a linear and non-linear transformation, respectively. Given the training and test data with distribution shift, we show the distributions of normalized output after each BN layer during training and test, and calculate the magnitude of difference between the estimated population statistics and expected ones (refer to as ESM, and see Section~\ref{sec:ES} for details). }
	\label{fig:intro}
	\vspace{-0.17in}
\end{figure}

%Firstly, the noireoty small-batch-size problem of BN;s ***. It hevalyise affects the BN's applicaiton in smal size trianing; 2)
Despite the common success of BN, it still suffers from problems when applied in certain scenarios~\cite{2020_arxiv_Huang,2021_ICML_Brock}. One notorious limitation of BN is its small-batch-size problem --- BN's error increases rapidly as the batch size becomes smaller~\cite{2018_ECCV_Wu,2019_ICCV_Singh}. Besides, a network with a naive BN gets significantly degenerated performance, if there exists covariate shift  between the training and test data~\cite{2017_arxiv_Li,2020_NIPS_Schneider,2020_arxiv_Nado,2021_WACV_Benz}.
% (\eg for domain adaption  and defending image corruptions tasks).  
While these problems raise across different scenarios and contexts, the estimated population statistics of BN used for inference seems to be the link between them: 1) the small-batch-size problem of BN can be relieved if its estimated populations statistics are corrected during test~\cite{2019_ICCV_Singh,2020_ICLR_Summers}; 2) and a model is more robust for unseen domain data (corrupted images) if the estimated population statistics of BN are adapted based on the available test data~\cite{2017_arxiv_Li,2020_NIPS_Schneider,2021_WACV_Benz}. 
%  preivous work py esitmiant the positn eo after traing to efeive ***. 

This paper investigates the estimation  of population statistics in a systematic way. We introduce expected population statistics of BN, considering the ill-defined population statistics of the activation with a varying distribution during training (see Section~\ref{sec:ES} for details).  We refer to as estimation shift of BN if its estimated population statistics do not equal to its expected ones, and design experiments to quantitatively investigate how the estimation shift affects a batch normalized network. 
% quantlive evialut the magnitedof of estiamtion shift. We quantlilaive evalute the estimaiotns shift by desing experiemtns. 
 
 Our primary observation is that the estimation shift of BN can be accumulated in a network (Figure~\ref{fig:intro} (a)). This observation provides clues to explain why a network with BN has significantly degenerated performance under small-batch-size training, and why the population statistics of BN need to be adapted if there exists distribution shift for input data during test. 
 We further find that a batch-free normalization (BFN)---normalizing each sample independently without across batch dimension---can block the accumulation of the estimation shift of BN. This relieves the  performance degeneration of a network if a distribution shift occurs. 
 %our observation provides a new view for explaining the recent successes of  methods combining BN with other normalization methods.
 
 These observations motivate our design of XBNBlock that replaces one BN with BFN in the bottleneck of residual-style networks~\cite{2015_CVPR_He,2017_CVPR_Xie}. We apply the proposed XBNBlock to ResNet~\cite{2015_CVPR_He} and ResNeXt~\cite{2017_CVPR_Xie} architectures and conduct experiments on the ImageNet~\cite{2015_IJCV_ImageNet} and COCO~\cite{2014_ECCV_COCO} benchmarks. 
 XBNBlock consistently improves the performance for both architectures, with absolute gains of $0.6\% \sim 1.1\%$ in top-1 accuracy for ImageNet,   $0.86\% \sim 1.62\%$ in bounding box AP for COCO using Faster R-CNN~\cite{2015_NIPS_Ren}, and  $0.56\% \sim 2.06\%$ ( $0.22\% \sim 1.18\%$) in bounding box AP (mask AP) for COCO using Mask R-CNN~\cite{2017_ICCV_He}. Besides, XBNBlock seems to be more robust to the distribution shift.
  %Results show that XBNBlock consistently improves the performance of different architectures by a significant margin and is more robust to distribution shift. 

\vspace{-0.05in}
\section{Related Work}
\label{sec_pre}
%\vspace{-0.05in}
\paragraph{Estimating and exploiting population statistics.}
Batch normalization (BN) suffers from small-batch-size problem,  since the estimation of population statistics could be inaccurate.  
 %is believed the main causes of its small-batch-size  problem.
  To address this issue, a variety of batch-free normalization (BFN) are proposed~\cite{2016_CoRR_Ba,2018_ECCV_Wu,2019_NIPS_Li}, \eg, layer normalization (LN)~\cite{2016_CoRR_Ba} and group normalization (GN)~\cite{2016_CoRR_Ba}. These works perform the same normalization operation for each sample during training and inference. Another way to reduce the discrepancy between training and inference is to combine the estimated population statistics with mini-batch statistics for normalization during training~\cite{2017_NIPS_Ioffe,2019_NeurIPS_Chiley,2020_ICLR_Yan,2020_ICML_Shen,2020_ECCV_Yong,2021_CVPR_Yao}.
 % These work can gradually adjust the weights between the estimated population statistics and mini-batch statistics, for inconsistent normalization operation.
   These work may outperform BN  trained with a small batch size, where estimation is the main issue\cite{2015_ICML_Ioffe,2018_UAI_Izmailov,2019_ICLR_Luo}, but they usually have inferior performance  when the batch size is moderate.  

%, and can work well on small-batch-size training, but they usually have inferior performance than BN under the moderate btch.

Some works focus on estimating corrected normalization statistics during inference only, either for domain adaptation~\cite{2017_arxiv_Li}, corruption robustness~\cite{2020_NIPS_Schneider,2020_arxiv_Nado,2021_WACV_Benz}, or  small-batch-size training ~\cite{2019_ICCV_Singh,2020_ICLR_Summers}. These strategies do not affect the training scheme of the model.
Li \etal~\cite{2017_arxiv_Li}  propose adaptive batch normalization (AdaBN) for domain adaptation, where the estimation of BN statistics for the available target domain is modulated during test.  This idea is further exploited to improve robustness under covariate shift of the input data with corruptions~\cite{2020_NIPS_Schneider,2021_WACV_Benz}. 
Another line of works correct the normalization statistics for small-batch-size training by optimizing ~\cite{2019_ICCV_Singh,2020_ICLR_Summers}  the
 sample weight  during inference, seeking for that the normalized output by population statistics are similar to those observed using mini-batch statistics during training.
 Besides, there are works considering the prediction-time batch settings~\cite{2019_arxiv_Song,2020_arxiv_Nado} for deep generative model~\cite{2019_arxiv_Song} and preventing covariate shift of the test data~\cite{2020_arxiv_Nado}, where the mini-batch statistics from the test data are used for inference. 
%A similar idea is also exploited in \cite{2020_ICLR_Summers}, where the sample weights are viewed as hyperparameters, which are optimized on a validation set. 

 %Other work consider to adjust the estimated BN for domain adaption\TODO{ref}, or for corruption robust\TODO{ref}. 

Compared to the works shown in above, our work focuses on investigating the estimation shift of BN in a network. Our observation, that the estimation shift of BN can be accumulated  in a network, provides clues to explain why a network with stacked BNs  has significantly degenerated performance under small-batch-size training, and why the population statistics of BN in each layer needs to be adapted if there exists covariate shift for input data during test. 
Besides, we design XBNBlock with BN and BFN mixed to block the accumulation of estimation shift of BNs.
% inserted   showns significantly suporot for the smiall batch-size problem, e.g, the esitaiton shift can be blocked in a network. These is the potentional problme of smoall-batchsize traning problem of BN. Our obsrevation also illustioanr why when a distibuton shift, the network has significantly degnereatd perofmarnce (Please see Section ** for detials), these work suport the devieaotn of why ADaBN, works for ***, **and ***. . Furemore, our pay more attenton on the netwokr deishg, and the proposed methods singinfajtly better than the orignal BN. 

\vspace{-0.15in}
\paragraph{Combining BN with other normalization methods.}
Researches have also be conducted to build a normalization module in a layer by combining different normalization strategies. 
Luo \etal~propose switchable normalization (SN)~\cite{2019_ICLR_Luo}, which switches among the different normalization methods  by learning their importance weights, computed by a softmax function. This idea is further extended by introducing the sparsity constraints~\cite{2019_CVPR_Shao}, whitening operation~\cite{2019_ICCV_Pan}, and dynamic calculation of the importance weights~\cite{2020_CVPR_Zhang}.  Other methods address the combination of normalization methods in specific scenarios, including image style transfer~\cite{2018_NeurIPS_Nam}, image-to-image translation~\cite{2020_ICLR_Kim}, domain generalization~\cite{2020_ECCV_Seo} and meta-learning scenarios~\cite{2020_ICML_Bronskill}. Different from these methods which aim to build a normalization module in a layer, our proposed XBNBlock is a building block with BN and BFN mixed in different layers. Furthermore, our observation, that a BFN  can block the accumulation of estimation shift of BNs in a network, provides a new view to explain the successes of above methods combining BN with other normalization methods. 
 %Other work consider to comb

Our work is closely related to IBN-Net~\cite{2018_ECCV_Pan}, which carefully integrates instance normalization (IN)~\cite{2016_arxiv_Ulyanov} and BN as building blocks, and  can be wrapped into several deep networks to improve their performances.
Note that IBN-Net carefully designs the position of an IN and its channel number, while the design of our XBNBlock is simplified. Moreover, IBN-Net is motivated by that IN can learn style-invariant features~\cite{2016_arxiv_Ulyanov} thus benefiting generalization, while our XBNBlock is motivated by that a BFN can relieve the estimation shift of BN, thus avoiding its degenerated test performance if inaccurate estimation exists.
 Here, we highlight  our observation that a BFN (\eg, IN) can block the accumulation of estimation shift of BNs also provide a reasonable explanation to the success of IBN-Net in its test  performance, especially in the scenarios with distribution shift~\cite{2018_ECCV_Pan}. %(\eg, domain adaptation and transfer learning tasks).

%Rather than designing a combinational normalization module, Pan \etal proposed IBN-Net, which carefully integrates IN and BN as building blocks, and  can be wrapped into several deep networks to improve their performances. Qiao \etal introduced batch-channel normalization (BCN)~\cite{2019_arxiv_Qiao2}, which integrates BN and channel-based normalizations (e.g., LN and GN) sequentially as a wrapped module.

%
%\TODO{The BCNwork}

%(a) shows the training and test errors with respect to the training epochs/iterations; (b) shows the ESM$_{\mu}$ of BN in different layers; (c) shows the ESM$_{\sigma}$ of BN in different layers.
\begin{figure*}[t]
	\centering
		\vspace{-0.1in}
	\hspace{-0.15in}	\subfloat[Training and test errors]{
		\begin{minipage}[c]{.30\linewidth}
			\centering
			\includegraphics[width=5.4cm]{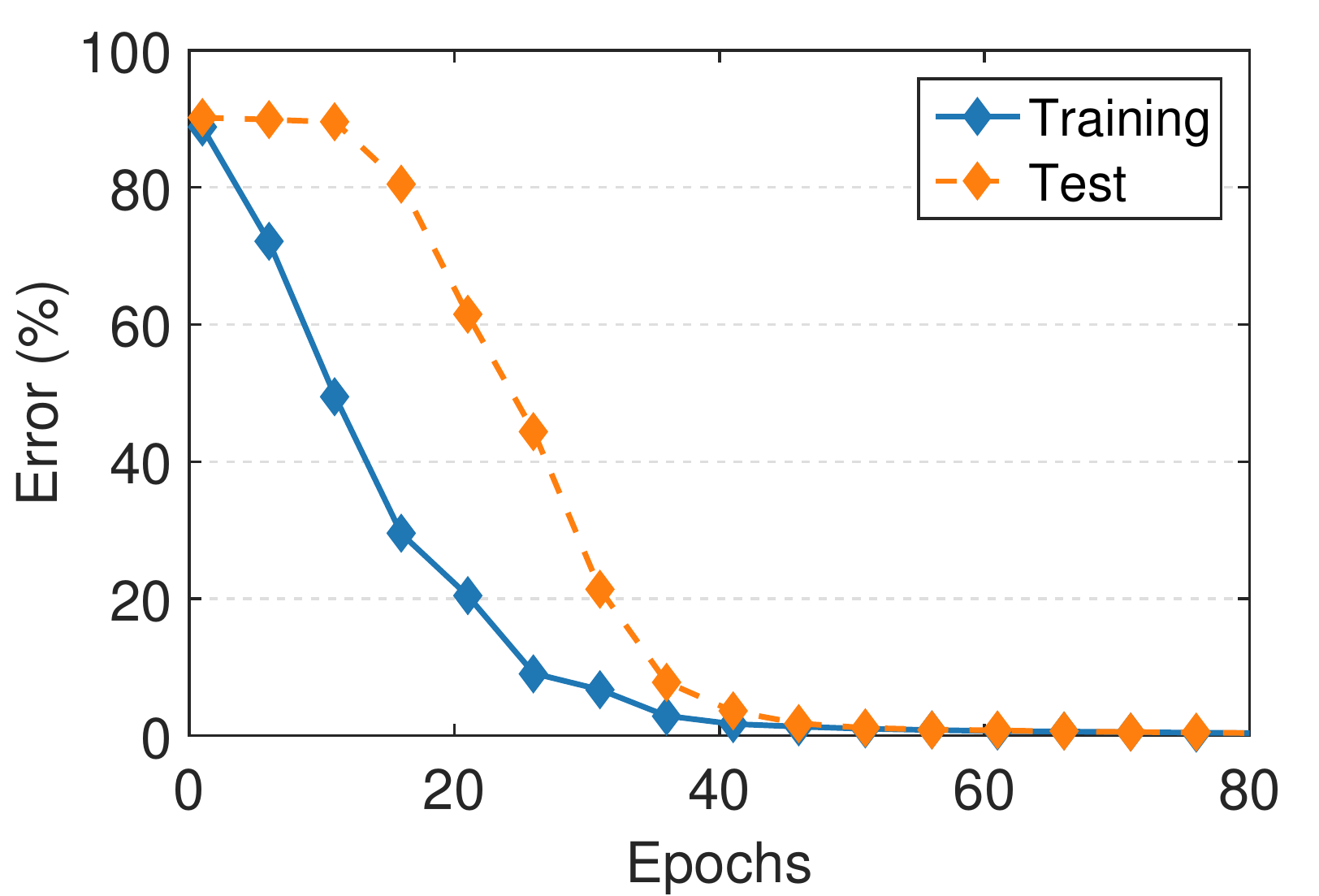}
		\end{minipage}
	}
	\hspace{0.15in}		\subfloat[ESM$_\mu$ of BN in different layers]{
		\begin{minipage}[c]{.30\linewidth}
			\centering
			\includegraphics[width=5.4cm]{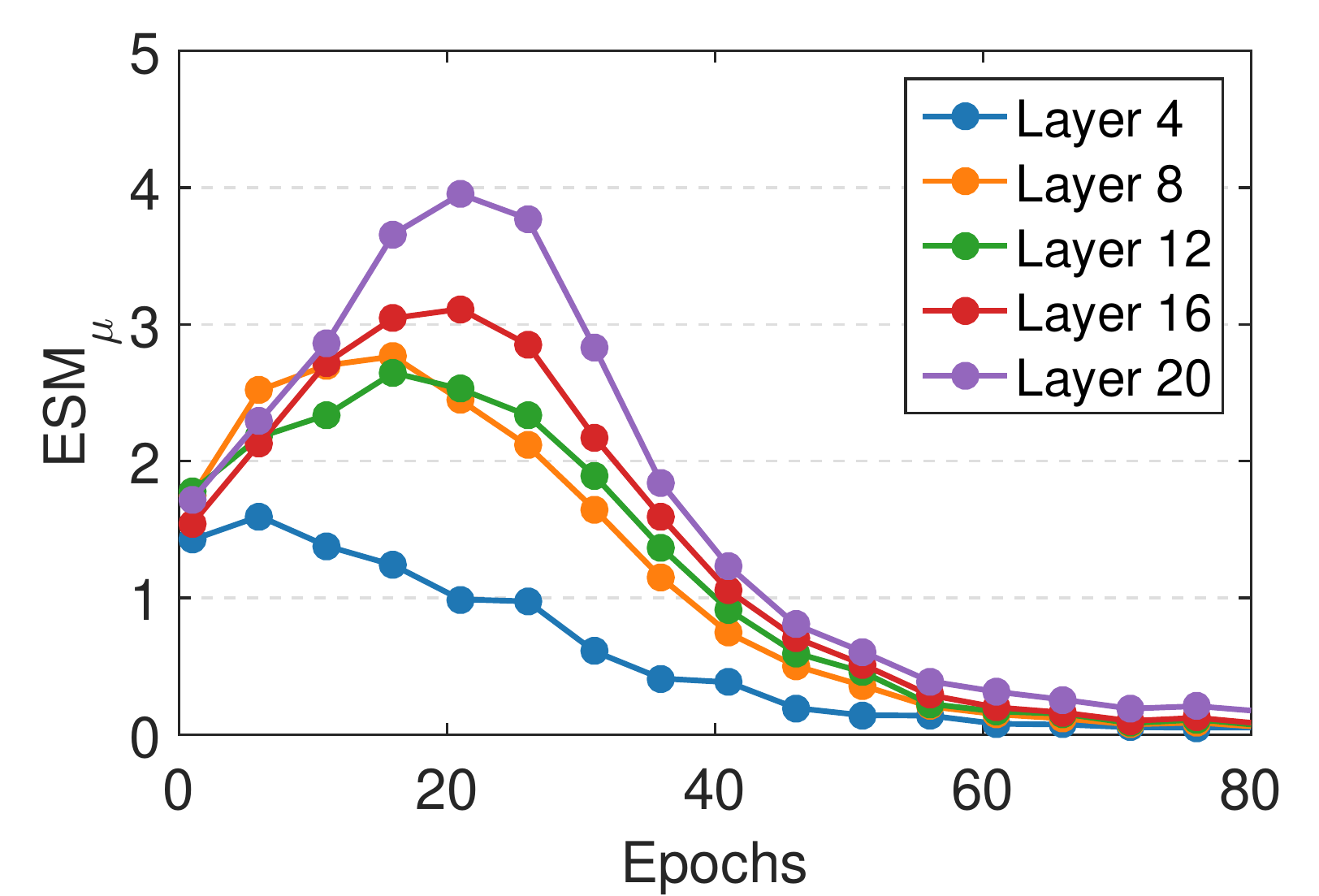}
		\end{minipage}
	}
	\hspace{0.15in}		\subfloat[ESM$_\sigma$ of BN in different layers]{
		\begin{minipage}[c]{.30\linewidth}
			\centering
			\includegraphics[width=5.4cm]{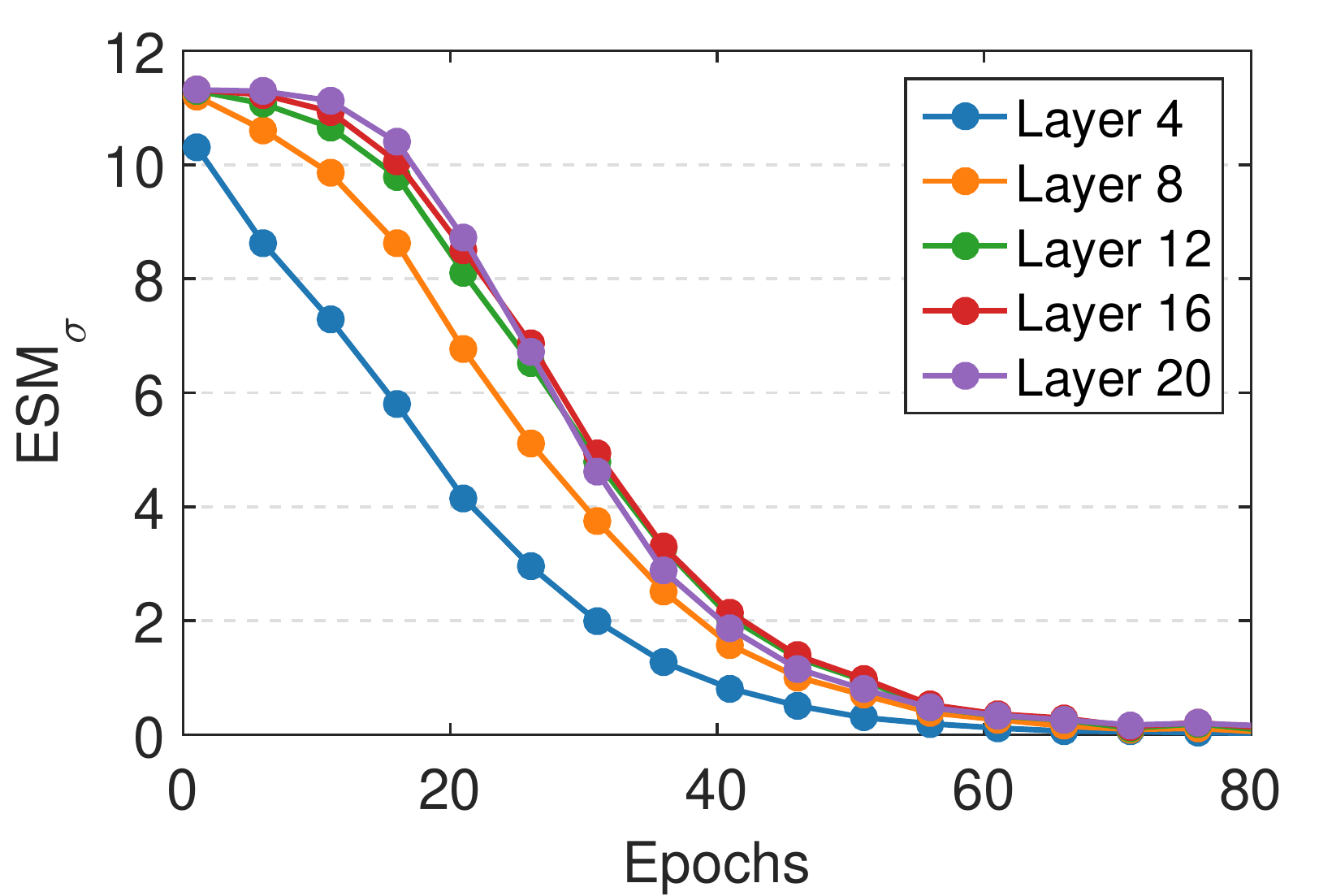}
		\end{minipage}
	}
	\vspace{-0.1in}
	\caption{Experiments with the training set $\mathbf{S}$ equaling to the test set  $\mathbf{S'}$.  We train a 20-layer MLP with 128 neurons in each layer for MNIST classification. $\mathbf{S}$ and $\mathbf{S'}$ are the original test set of MNIST with 10,000 samples. We use full-batch gradient descent to train 80 epochs (iterations) with a learning rate of 0.1. The estimated population statistics of BN are calculated by the commonly used running average (Eqn.~\ref{eqn:BN-inf}) with update factor $\alpha=0.9$.  We also try other configurations (\eg, varying the learning rate, update factor $\alpha$ and depth of the network), and further conduct experiments on convolutional neural networks (CNNs). We obtain similar results (see \TODO{\SM} ~\ref{supsec:investigation} for details). }
	\label{fig:exp1_valval}
	\vspace{-0.1in}
\end{figure*}

\begin{figure*}[t]
	\centering
	%\vspace{-0.08in}
	\hspace{-0.15in}	\subfloat[Training (solid lines) and test (dashed) errors]{
		\begin{minipage}[c]{.30\linewidth}
			\centering
			\includegraphics[width=5.4cm]{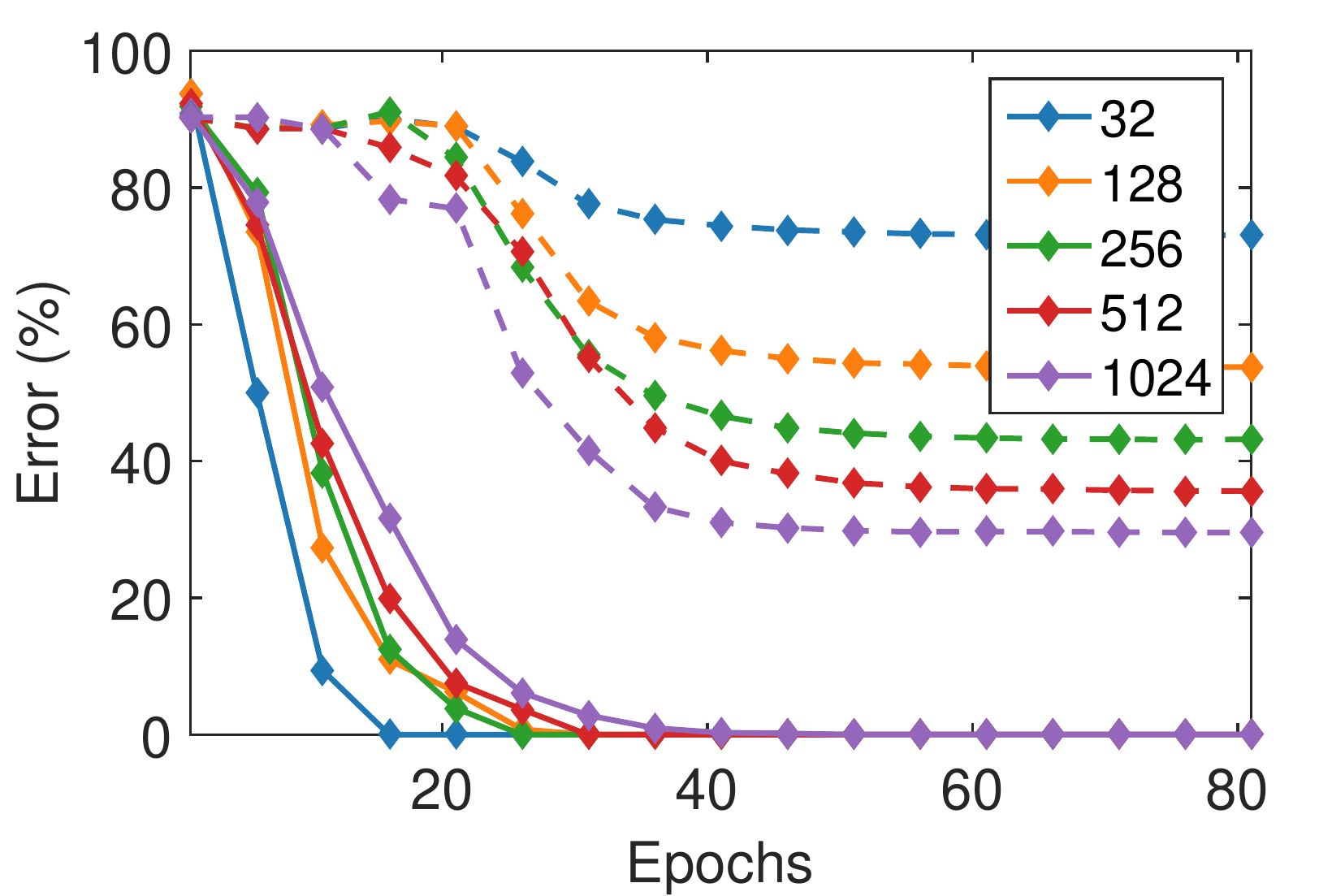}
		\end{minipage}
	}
	\hspace{0.15in}		\subfloat[ESM$_\sigma$ of BN at the end of training]{
		\begin{minipage}[c]{.30\linewidth}
			\centering
			\includegraphics[width=5.4cm]{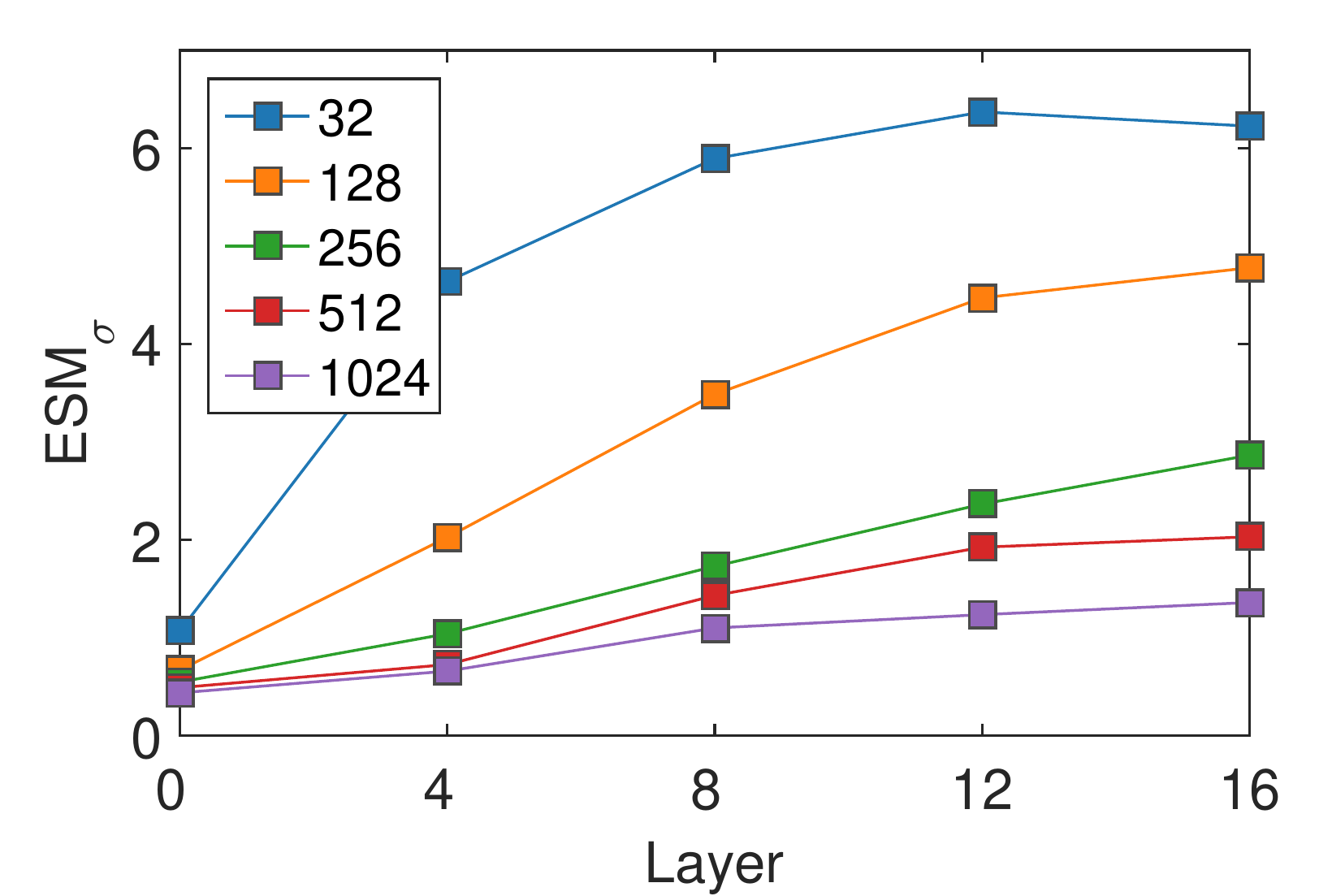}
		\end{minipage}
	}
	\hspace{0.15in}		\subfloat[ESM$_\sigma$ of BN in an arbitrary (the 16-th) layer ]{
		\begin{minipage}[c]{.30\linewidth}
			\centering
			\includegraphics[width=5.4cm]{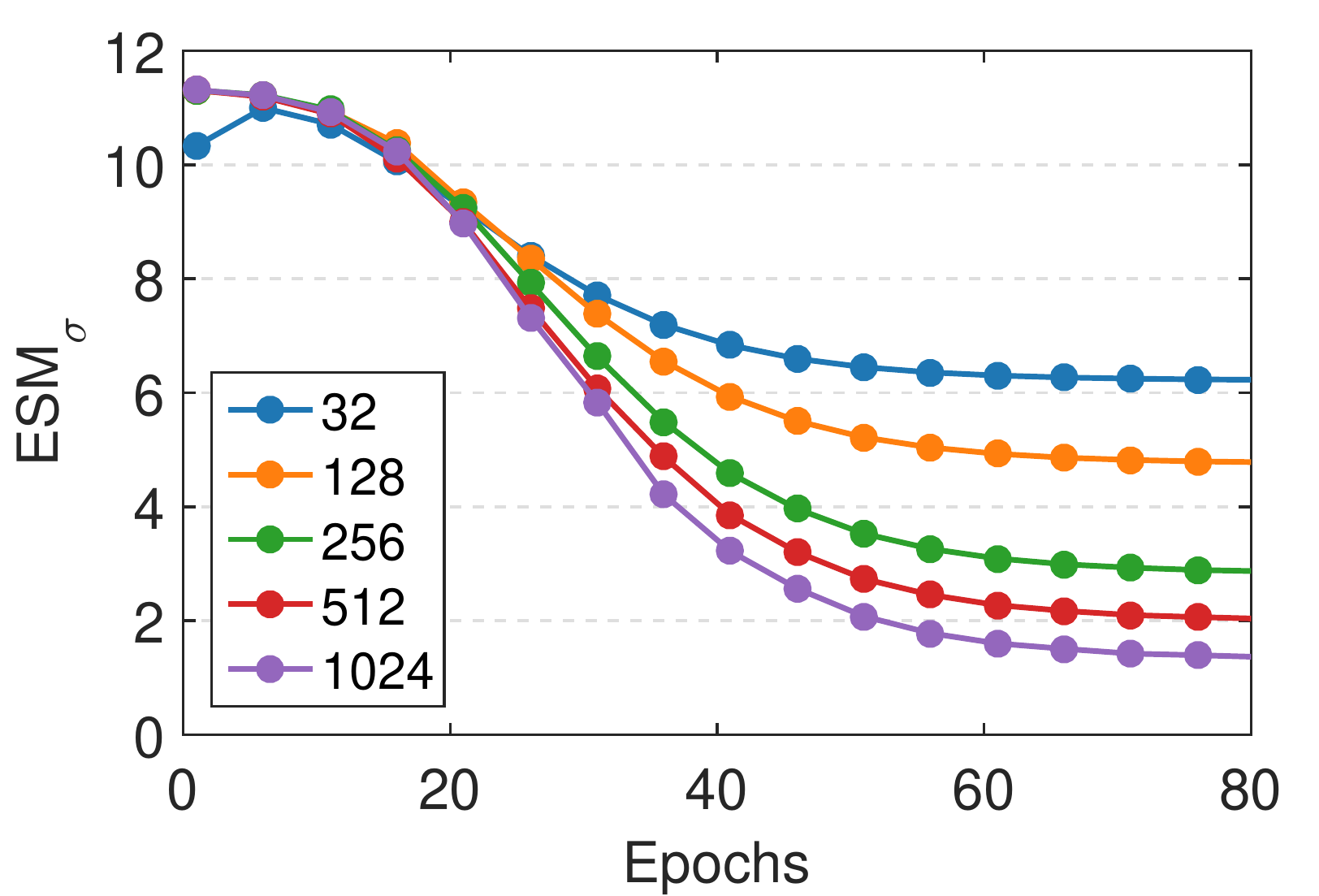}
		\end{minipage}
	}
	\vspace{-0.1in}
	\caption{Experiments using the training set $\mathbf{S}$ sampled from the test set  $\mathbf{S'}$. We follow the same experimental setup  in Figure~\ref{fig:exp1_valval} except that we use the training set with varying size $|\mathbf{S}|=\{32, 128, 256, 512, 1024\}$.  We use $\|\sqrt{\sigma^2_{train}} - \sqrt{\sigma^2_{test}}\|_2$ to evaluate the distribution shift of the input (referred to as $ESM_{\sigma}$ \wrt~ layer 0), where $\sigma^2_{train}$ ($\sigma^2_{test}$) is the variance of the training (test) set.}
	%Note that we calculate the distribution shift between training and test set using the $L^2$-norm between their variance (ESM$_\sigma$ with respect to layer 0) with a scaled constant of 128/1024, since the dimension of the input is 1024 while the hidden layer 128.
	\label{fig:exp2_valval_BS}
	\vspace{-0.17in}
\end{figure*}
% (a) shows the training (solid lines) and test (dashed lines) errors with respect to training epochs; (b) ESM$_{\sigma}$ of BN in an arbitrary layer (the 16-th layer) with respect to training epochs; (c) ESM$_{\sigma}$ of BN in different layers after the model is trained. 

\vspace{-0.03in}
\section{Preliminary}

\paragraph{Batch normalization.}
Let $\x{} \in \mathbb{R}^d$ be the $d$-dimensional input to a given layer of multi-layer perceptron (MLP). During training, batch normalization  normalizes each neuron/channel within $m$ mini-batch data by\footnote{BN usually uses extra learnable scale and shift parameters~\cite{2015_ICML_Ioffe}, and we  omit them as they are not relevant to the discussion of normalization.}
 \begin{small}
%	{\setlength\abovedisplayskip{0.05in} 
%	\setlength\belowdisplayskip{0.04in}
	%	 \vspace{-0.2in}
\begin{equation}
	\label{eqn:BN}
	\Nx{j}=BN_{train}(\x{j})= \frac{\x{j} - \mu_j}{\sqrt{\sigma^2_j +\epsilon}}, ~~j=1,2, ..., d,
\end{equation}
\end{small}
\hspace{-0.04in}where $\mu_j=\frac{1}{m}  \sum_{i=1}^{m}  \x{j}^{(i)}$ and $\sigma^2_j = \frac{1}{m}   \sum_{i=1}^{m} (\x{j}^{(i)}-\mu_j)^2 $ are the mini-batch mean and variance for each neuron, respectively, and $\epsilon$ is a small number to prevent numerical instability.
During inference/test, the population mean $\tilde{\mu}$ and variance $\tilde{\sigma}^2$ of the layer input are required for BN to make a deterministic prediction~\cite{2015_ICML_Ioffe} as:
 \begin{small}
%	\setlength\abovedisplayskip{0.05in} 
%	\setlength\belowdisplayskip{0.04in}
	%	 \vspace{-0.2in}
\begin{equation}
	\label{eqn:BN-inf}
	\Nx{j}=BN_{inf}(\x{j})= \frac{\x{j} - \tilde{\mu}_j}{\sqrt{\tilde{\sigma}^2}}, ~~j=1,2, ..., d.
\end{equation}
\end{small}
\hspace{-0.04in}Even though the population statistics $\{\tilde{\mu}, \tilde{\sigma}^2\}$ of the layer input are ill-defined (illustrated in Section~\ref{sec:expected_PS}), their estimation  $\{\hat{\mu}, \hat{\sigma}^2\}$ are usually used in Eqn.~\ref{eqn:BN-inf} by calculating the running average of mini-batch statistics over different  training iterations $t$ with an update factor $\alpha$ as follows:
 \begin{small}
%	\setlength\abovedisplayskip{0.05in} 
%	\setlength\belowdisplayskip{0.04in}
	%	 \vspace{-0.2in}
\begin{equation}
	\label{eqn:running-average}
	\begin{aligned}
		\begin{cases}
			\quad \hat{\mu}^t=  (1-\alpha) \hat{\mu}^{t-1} + \alpha \mu^{t-1},\\
			\quad (\hat{\sigma}^t)^2=  (1-\alpha) (\hat{\sigma}^{t-1})^2 + \alpha (\sigma^{t-1})^2.
		\end{cases}
	\end{aligned}
\end{equation}
\end{small}
\hspace{-0.04in}The discrepancy of BN during training and inference limits its usage in recurrent  neural network~\cite{2016_CoRR_Ba}, or harms the performance for small-batch-size training~\cite{2018_ECCV_Wu}, since the estimation of population statics can be inaccurate. 
%We  note that other methods for estimating $\{\hat{\mu}, \hat{\sigma}^2\}$ also exist. \Eg, $\{\hat{\mu}, \hat{\sigma}^2\}$ are calculated after the model is trained~\cite{2015_ICML_Ioffe,2019_ICLR_Luo,2020_arxiv_Nado}, especially for the small-batch-size training~\cite{2019_ICCV_Singh,2020_ICLR_Summers}; or adapted using the available test data for domain adaption~\cite{2017_arxiv_Li} and improving corruption robustness~\cite{2020_NIPS_Schneider}.   
\vspace{-0.15in}
\paragraph{Batch-free normalization.}
There exists batch-free normalization for avoiding normalization along the batch dimension, and thus avoiding the estimation of population statistics.
These methods use consistent operations during training and inference. 
One representative method is layer normalization (LN)~\cite{2016_CoRR_Ba} that  standardizes the layer input within the neurons for each training sample, as:
 \begin{small}
	\begin{equation}
		\label{eqn:LN}
		\Nx{j}=LN(\x{j})= \frac{\x{j} - \mu}{\sqrt{\sigma^2 +\epsilon}}, ~~j=1,2, ..., d,
	\end{equation}
 \end{small}
\hspace{-0.04in}where $\mu=\frac{1}{d}  \sum_{i=1}^{d}  \x{j}$ and $\sigma^2 = \frac{1}{d}   \sum_{i=1}^{d} (\x{j}-\mu)^2 $ are the  mean and variance for each sample, respectively. LN is further generalized by group normalization (GN)~\cite{2018_ECCV_Wu}  that divides the neurons into groups and performs the standardization within the neurons of each group independently. By changing the group number, GN is more flexible than LN, enabling it to achieve good performance on visual tasks limited to small-batch-size training (\eg, object detection and segmentation~\cite{2018_ECCV_Wu}).
While these BFN methods can work well on certain scenarios, they cannot match the performance of BN in most situations and are not commonly used in CNN architectures.

%how accurate the estimated statistics relative to its expected one, and 
%BN introduces inconsistent normalization operations during training (using mini-batch statistics, and inference (using population statistics estimated in). This means that the upper layers are trained on representations different from those computed during inference. Th
%we are not aware of any prior work investigating its usefulness for deep learning under covariate shift..
%Previous work 
%The inconsistent operation of BN between training (Eqn. \TODO{ref}) and inference (Eqn. \TODO{ref}) make the usage of BN challenge, heavily affects the usage of BN in the small batch size training or domain adaptation, defending input noise. While prevous work attribute the small batch problem of BN to its inconsistent operation. Here we address that 
%While previous work naturally use the estimated population statistics for inference, we note that no work appareently investigate how the estimated population can be well estimating the expected population, and what factors affects the estimation. This work will address on this estimation problem. 

\vspace{-0.03in}
\section{Estimation Shift of Batch Normalization}
%The discrepancy of BN between training and inference means that the representations in lower layers used for training the upper layers might be different from those computed during inference. While previous work naturally use the estimated statistics for inference, we are not aware of any prior work investigating the estimated statistics of BN in a systematic manner.   
We begin with illustrating the ill-defined population statics of BN, and then design  comprehensive experiments  for investigating the estimation shift of BN. 
\subsection{Expected Population Statistics of BN}
\label{sec:expected_PS}
%\vspace{-0.05in}
Let $\mathbf{S}$ be the training set and $\{S^t\}_{t=1}^{T}$  the mini-batch data sampled from $\mathbf{S}$ during training. Considering a neural network with a  BN $F_{\psi, \theta}(S)=F^{post}_{\psi}(BN(F^{pre}_{\theta}(S)))$, we denote $X=F^{pre}_{\theta}(S)$ and $\widehat{X}=BN(X)$. The population statistics of the certain training set $\mathbf{S}$ are well-defined and they can be well estimated straightforwardly using the mini-batch statistics of  $\{S^t\}_{t=1}^{T}$. However, the population statistics of the activation $\mathbf{X}=F^{pre}_{\theta}(\mathbf{S})$ are ill-defined, because $\mathbf{X}$ is varying during training due to the update of parameter $\theta$ in each iteration. Indeed, the mini-batch samples of $\mathbf{X}$ are $ X^{t} = F^{pre}_{\theta^{t}} (S^t),$ for$~ t=1, ..., T$, which depends not only on the mini-batch input $S^t$, but also on the model sequences $\{F^{pre}_{\theta^t} (\cdot)\}_{t=1}^{T}$. Therefore, the  population statistics of  $\mathbf{X}$ should be a function of the training set $\mathbf{S}$ and the varying model sequences $\{F^{pre}_{\theta^t} (\cdot)\}_{t=1}^T$ during training. Even though it is difficult to explicitly define the population statistics of  $\mathbf{X}$ from the statistical view,
we note that the mini-batch input $\widehat{X}^t$  of sub-network $F^{post}_{\psi} (\cdot) $ is always a standardized distribution for each iteration. Therefore, the ideal population statistics of $\mathbf{X}$ should ensure the normalized output standardized  over the test set.
We  implicitly define the expected population statistics of BN as follows. 
\vspace{-0.02in}
%	\begin{small}
	\begin{defn}
		\label{def1:EPS}
		Let $F_{\tilde{\psi},\tilde{\theta}}(\cdot)$ be the trained model on training set $\mathbf{S}$. Given the test set $\mathbf{S'}$, we refer to $\{\tilde{\mu}, \tilde{\sigma}^{2}\}$ are the \textbf{expected population statistics} of BN, where $\tilde{\mu}$ ($\tilde{\sigma}^{2}$) is the mean (variance) of BN's input $\mathbf{X}=F^{pre}_{\tilde{\theta}}(\mathbf{S'})$.

	\end{defn}
	%	\end{small}
 Note that the expected population statistics of BN are defined on the trained model $F_{\tilde{\psi},\tilde{\theta}}(\cdot)$ conditioned on the input from the test set $\mathbf{S'}$ rather than the training set $\mathbf{S}$, because the population statistics of $\mathbf{X}=F^{pre}_{\tilde{\theta}}(\mathbf{S})$ consider only the last trained model $F^{pre}_{\tilde{\theta}}(\cdot)$ rather than the model sequences $\{F^{pre}_{\theta^t} (\cdot)\}_{t=1}^{T}$. Indeed, the population statistics of $\mathbf{X}=F^{pre}_{\tilde{\theta}}(\mathbf{S})$ can be readily calculated once the model is trained, as introduced in ~\cite{2015_ICML_Ioffe,2019_ICLR_Luo,2021_arxiv_Wu}. However, they usually have worse generalization performance than the one used by running average shown in Eqn.~\ref{eqn:running-average}.
 % We will further illustrate them in Section~\ref{sec:accumulation_ES}.
 
 \subsection{An Investigation into the Estimation Shift}
 \label{sec:ES}

Given the expected population statistics of the BN defined, we refer to as \textit{\textbf{\ES}} of BN   if its estimated population statistics do not equal to its expected ones. It is important to investigate how the \ES~of BN affects the performance of batch normalized network. We thus seek to  quantitatively measure the magnitude of the difference between estimated  statistics and its expected ones. 
%by the following definition.   

\vspace{-0.02in}
%	\begin{small}
\begin{defn}
	\label{def2:ESM}
	Let $\tilde{\mu}$ ($\tilde{\sigma}^2$) is the expected population mean (variance) of BN and  $\hat{\mu}$ ($\hat{\sigma}^2$) is the estimated one. We define the \textbf{\ESM} (ESM) as the $L^2$-norm of their difference. \Eg, $ESM_{\mu}= \|\hat{\mu} -\tilde{\mu}\|_2$ and $ESM_{\sigma}= \|\sqrt{\hat{\sigma}^2} - \sqrt{\tilde{\sigma}^2}\|_2$.
\end{defn}
%	\end{small}
In the following sections, we design experiments to investigate how the \ES~ of BN affects the performance of batch normalized network and how it can be rectified.

%It is interesting to investigate how the \ES of BN affects the performance of batch normalized network. Especially, how the depth of a neural network affects the estimation and whether such an estimation can be accumulated by the stack of multiple layers with batch normalization. We conduct two toy experiments that show ***.
\vspace{-0.05in}
\subsubsection{Accumulation of Estimation Shift in a Network}
\label{sec:accumulation_ES}
\vspace{-0.05in}
%Here, we investigate how the how the depth of a neural network affects the estimation of BN.
We consider two experimental setups: 1) in setup one, we use the training set $\mathbf{S}$  equaling to the test set $\mathbf{S'}$ for investigating  \ES~of BN under the scenario without distribution shift of the input data; 2) in setup two, the training set $\mathbf{S}$ is sampled from the test set $\mathbf{S'}$. We vary the size of $\mathbf{S}$ to modulate the distribution shift between training and test set.
% For both setting, we use the full-batch gradient descent to simplify the  analyses. 
\vspace{-0.15in}
\paragraph{Setup one.} The details of experimental setup and the results are shown in Figure~\ref{fig:exp1_valval}. 
 We observe that there are significant gaps between the training and test errors in the first 30 epochs.
 Note that the training and test errors in this setup should be the same  over iterations if BN adopts the same operation during training and inference. 
 %This observation implies that the inconsistent operations of BN may downgrade the test performance even if it has good training performance. 
 In Figure~\ref{fig:exp1_valval} (b) and (c), $ESM_{\mu}$ and $ESM_{\sigma}$ of BN in certain layers are significantly larger than zero in the first 30 epochs and then gradually converge to zero. This phenomenon clearly shows that the error gaps between training and test are mainly caused by the inaccurate estimation of the population statistics of BN. 
 
 One important observation is that the $ESM_{\mu}$ and $ESM_{\sigma}$ of BN in deeper layers  have potentially  higher values during the first 30 epochs. This observation implies that the estimation of BN in lower layers will affect the one in upper layer. The estimation shift of BN in upper layer will be amplified if the BN in lower layer suffers from estimation shift which causes a distribution shift of the input into upper layer between training and test. 
 Therefore, the inaccurate estimation of population statistics can be potentially accumulated/compounded due to the stack of BN layers.
% because the estimation shift in lower layer can lead to a distribution shift during training and test for the input of the upper layer. 
 %which will probably downgrade the test performance. 
 %Thus, if we can break such an accumulation, it is possible to relieve the estimation problem as we will discussed in Section **.  
%\TODO{Address We also run the experiments with different architectures, different learning rate, and also on other dataset, please see the SM for details. }
\vspace{-0.15in}
\paragraph{Setup two.} In this setup, the training set $\mathbf{S}$ is sampled from the test set $\mathbf{S'}$ and we vary the size of training set $|\mathbf{S}|$ to modulate the distribution shift between the training and test set. We expect to see how the varying distribution shift affects the estimation of BN's population statistics in a network.  The details of experimental setup and results are shown in Figure~\ref{fig:exp2_valval_BS}.
 % We use $\|\sqrt{\sigma^2_{train}} - \sqrt{\sigma^2_{test}}\|_2$ to evaluate the distribution shift between the  training  and test set, where $\sigma^2_{train}$ ($\sigma^2_{test}$) is the variance of the training (test) set, and we report the results in   Figure~\ref{fig:exp2_valval_BS} (c) (the $ESM_{\sigma}$ with respect to the Layer 0). 
  We find that the distribution shift  can be potentially larger when decreasing the size of sampled training set from Figure~\ref{fig:exp2_valval_BS} (b).  Furthermore, the  $ESM_{\sigma}$ of all the BN layers are significantly larger than zero, and a BN layer in a model trained with fewer samples has higher $ESM_{\sigma}$. Besides, in Figure~\ref{fig:exp2_valval_BS} (a),  we observe that all the models can be trained with an zero training error, while the test error is significantly higher if a model is trained on the training set with fewer samples. 
  %By looking into the  $ESM_{\sigma}$ in certain layers after the training is finished (Figure ~\ref{fig:exp2_valval_BS} (b)), we observe that the  $ESM_{\sigma}$ of all the BN layers are significantly larger than 0, and a BN layer in a model trained with fewer samples has higher $ESM_{\sigma}$.
These observations imply that the distribution shift of the input between the training and test set can cause the estimation shift of BN, which has a detriment effect on the test performance. \Eg, we find that the model without BN obtains a test error of $57.73\%$  when using 32 training samples, compared to the model with BN having a test error of $73.02\%$.

One important observation is that $ESM_{\sigma}$ of BN in deeper layers  have potentially  higher value at the end of training.
%, and this observation applies all the models trained on different size of training set. 
%It seems that the 
%This observation shows that the estimation shift of BN could be accumulated over layers in a network. 
This observation shows remarkable evidences to support that the estimation shift of BN can be accumulated due to the stack of BN layers. Moreover, the estimation shift is graver if the model is trained with fewer training samples and stronger distribution shift of the input data.

 %Besides, the $ESM_{\sigma}$ of BN  gradually converges to a stable value (Figure ~\ref{fig:exp2_valval_BS} (c)) 
Here, we highlight that it is important to define the expected population statistics of BN on $F_{\tilde{\theta}}(\mathbf{S'})$
 rather than $F_{\tilde{\theta}}(\mathbf{S})$. We note that the $ESM_{\sigma}$ of BN  gradually converges to a stable value (Figure ~\ref{fig:exp2_valval_BS} (c)) in this experiment, which suggests that the estimation used by the running average (Eqn.~\ref{eqn:BN-inf})  converges to the estimation on the trained model over the training set~\cite{2015_ICML_Ioffe,2019_ICLR_Luo} (\ie, $F_{\tilde{\theta}}(\mathbf{S})$).  $ESM_{\sigma}$ will be zero if  $ESM_{\sigma}$ is define on $F_{\tilde{\theta}}(\mathbf{S})$. This is not what we expect, because it provides no information to diagnose the degenerated test performance of a model trained on the training set with fewer samples that suffers larger distribution shift over the test set, as shown in this experiment. 
%  This suggests that the ESM  provides no information  be the same if we use the population of training set for the expected population statistics, however it is not what we expected, becasue it provides no information to diognoaize. 
%%, which might be caused by the distribution shift of the input between training and test set.  
%even though  ESM does not change unpon the trained is finished, witch suggested the estimation used by the running average Eqn.\TODO{ref} will converged to the estimation based on the trained model over training set (that is $F_{\tilde{\theta}}(\mathbf{S})$). This suggests that the ESM will be the same if we use the training set for the expected population statistics, however it is not what we expected, because the test accuracy are significantly different if we use different training set. This experiments suggested the importance we define the ESM over test set, because it can more cover the situation where the trianing and test set has ditraitns shift. 

In summary, according to the experiments above, we argue that estimation shift of BN can be potentially accumulated in a network with stacked BNs, which probably has a detriment effect on the test performance of the network, especially with the distribution shift occurred.

% Therefore, the performance of network could be improved if we can block the accumulation of estimation shift of BN.
%  the inaccurate estimation can be accumulated considering BN in each layer, and the distribution shift between the training set and the test set can also be potentially propagated, which will affects the performance of the network. 

%Note: The inconsistency problem of BN during training and inference is not equal to the estimation shift problem. We thus decomposition the inconsistency between training of inference of BN into two components: 1) the difference of normalization operation; 2) the estimation shift problem

\begin{figure}[t]
	\centering
	\vspace{-0.1in}
\hspace{0.15in}	\subfloat[Training and test errors]{
		\begin{minipage}[c]{.48\linewidth}
			\centering
			\includegraphics[width=4.2cm]{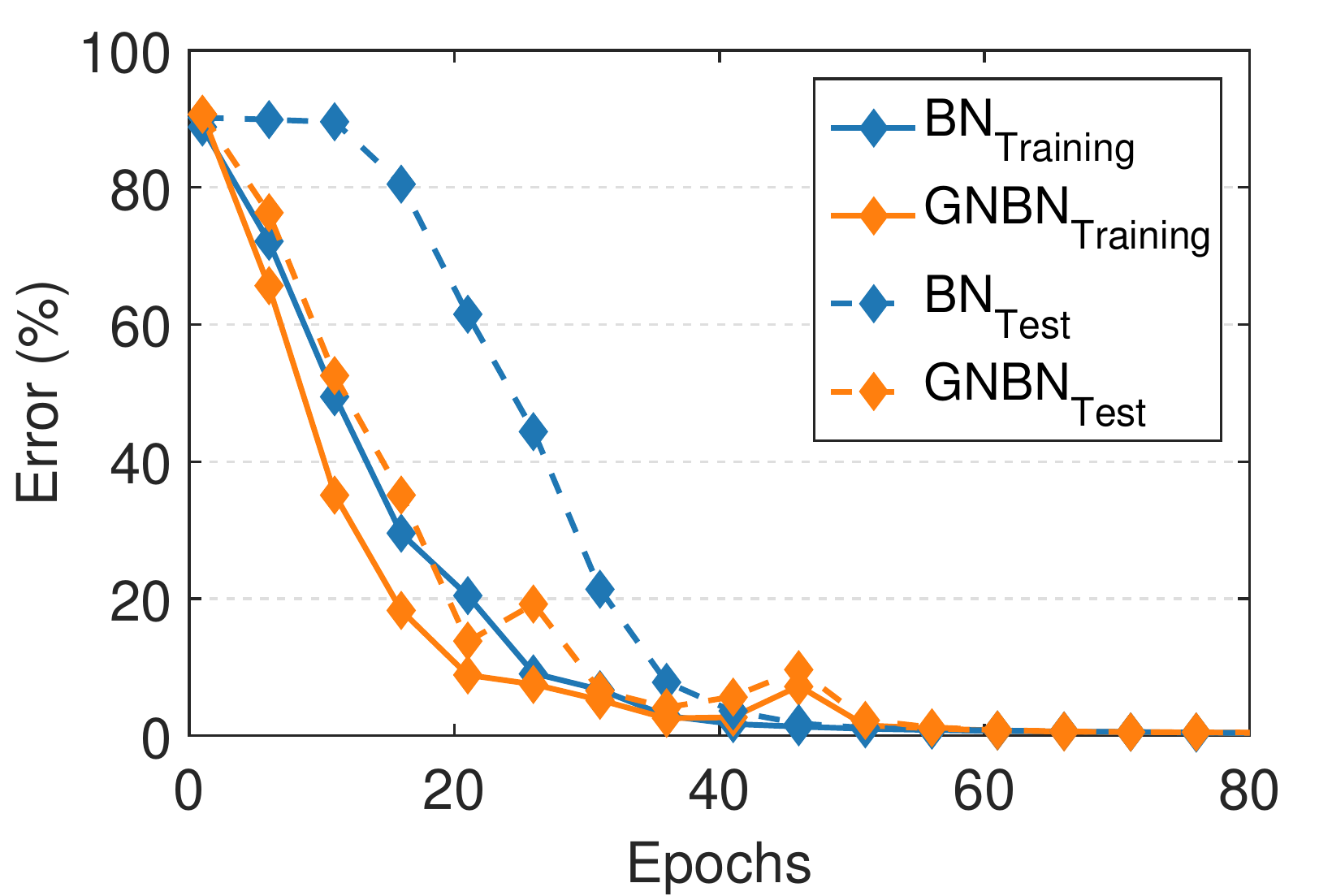}
		\end{minipage}
	}
	\subfloat[ESM$_\sigma$ of BN in different layers]{
		\begin{minipage}[c]{.48\linewidth}
			\centering
			\includegraphics[width=4.2cm]{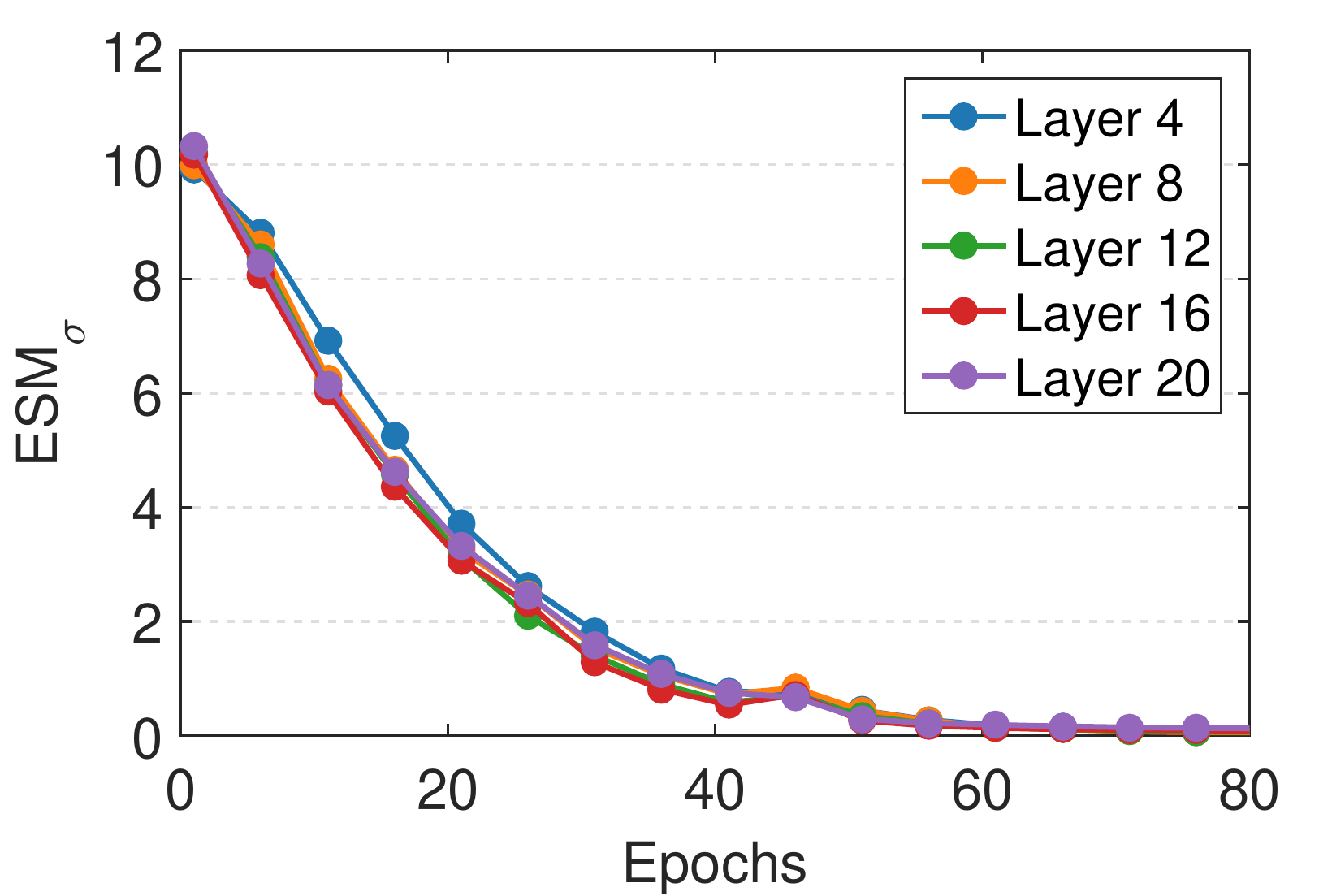}
		\end{minipage}
	}
	\vspace{-0.1in}
	\caption{Experiments on a network with BN and GN mixed. We follow the same setup shown in Figure~\ref{fig:exp1_valval}, except that we replace the BNs of the odd layers with GNs in the network (referred to as `GNBN').  Here, we use GN with a group number of 4. We also try different group numbers and obtain similar observations (see \TODO{\SM} ~\ref{supsec:investigation} for details).}
	\label{fig:exp1_valval_GNvsBN}
	\vspace{-0.17in}
\end{figure}
%(a) shows the training and test errors with respect to the epochs; (b) shows the ESM$_{\sigma}$ of BN in different layers.

\begin{figure}[t]
	\centering
   \vspace{-0.1in}

	\hspace{0.15in}	\subfloat[ESM$_\sigma$ of BN at the end of training]{
		\begin{minipage}[c]{.48\linewidth}
			\centering
			\includegraphics[width=4.2cm]{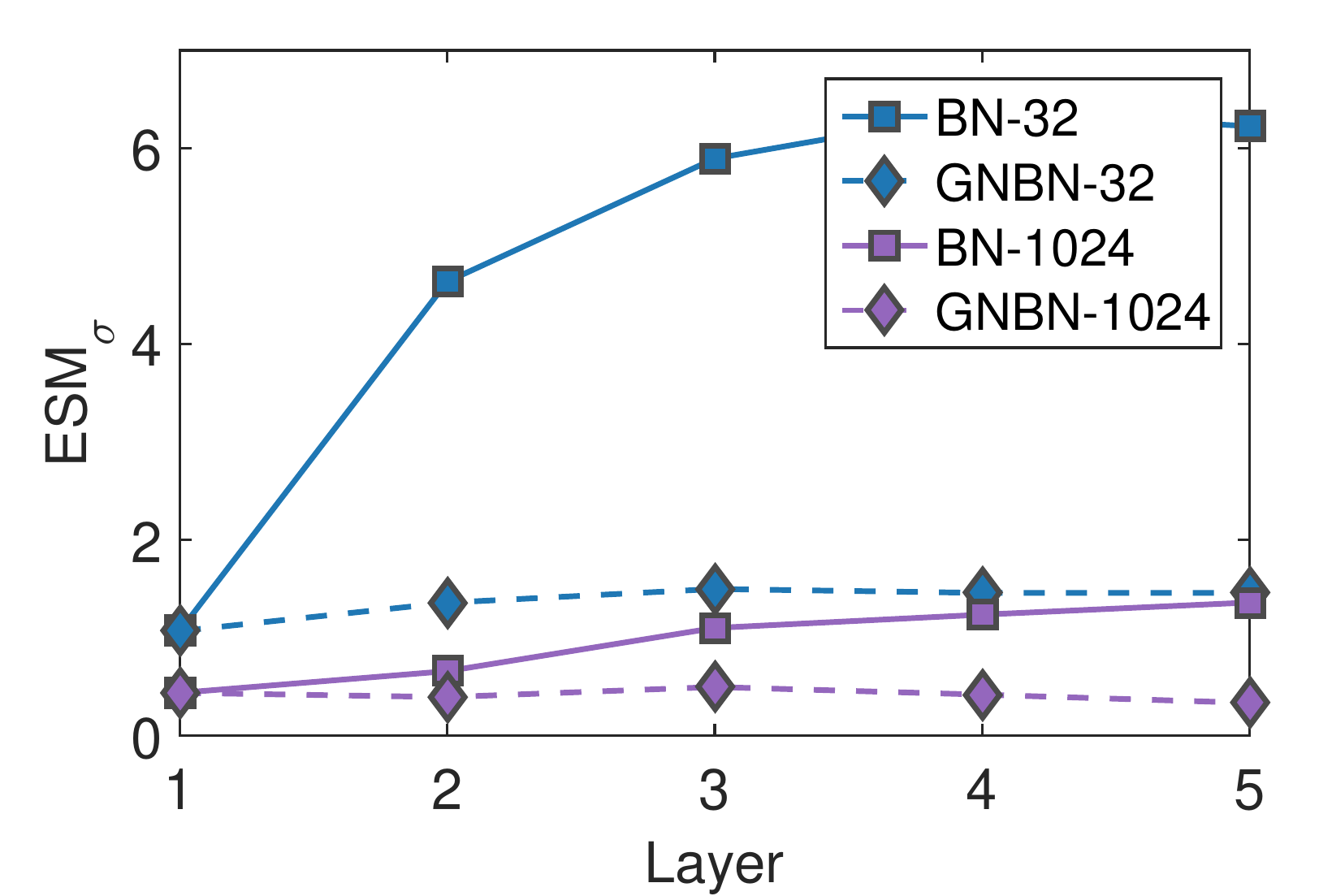}
		\end{minipage}
	}
	\subfloat[Test errors]{
	\begin{minipage}[c]{.48\linewidth}
		\centering
		\includegraphics[width=4.2cm]{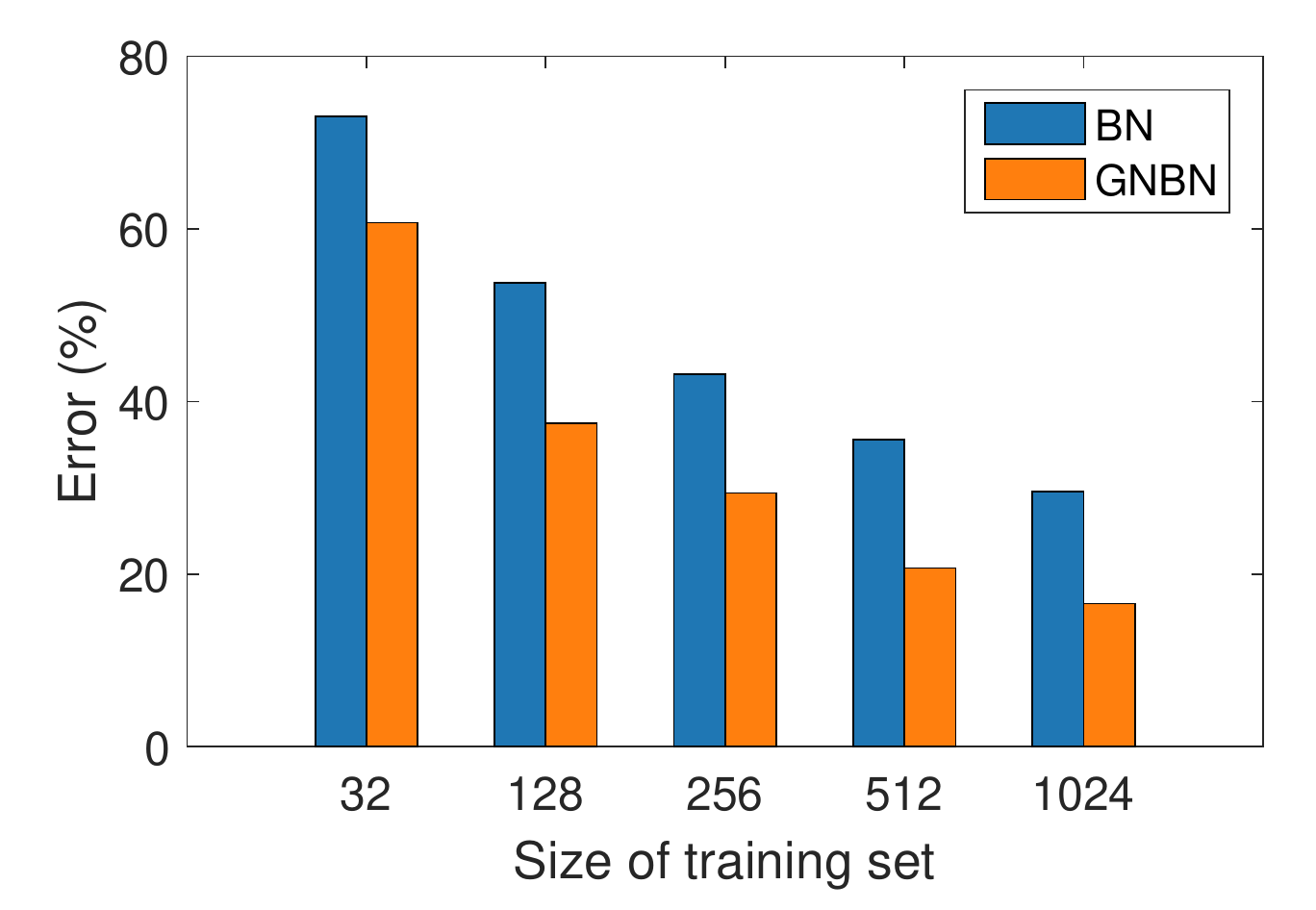}
	\end{minipage}
}
	\vspace{-0.1in}
	\caption{Experiments on a network with BN and GN mixed. We follow the same setup shown in Figure~\ref{fig:exp2_valval_BS}, except that we replace the BNs of odd layers with GNs in the network.  Here, `-N' indicates that the model is trained on the training set with N samples.}
	\label{fig:exp2_valval_GNvsBN}
	\vspace{-0.17in}
\end{figure}
%(a) shows ESM$_\sigma$ of BN at the end of training, and `-N' indicates that the model is trained on the training set with N samples. (b) shows the test errors for different training set;}
	\vspace{-0.05in}
\subsubsection{Blocking the Accumulation of Estimation Shift}
\label{sec:accumulation_block_ES}
	\vspace{-0.05in}

%While these BFN methods ensure consistent training and inference operation and can work well on certain scenarios, the disadvantages of these methods is that their improved conditioning is not well demonstrated and is problems to make the examples has nearly the same representations, which make them not commonly used in CNN architectures. 

We experimentally show that the accumulation of estimation shift of BN can be relieved if a BFN is inserted in a network.  We replace the BNs of the odd layers with GNs, and refer to this network as `GNBN'.  
% design a network that its BN in the odd layer 
%while  the batch-free normalized applied in certain layer of a batch normalized network. These batch-free normalized methods can effectively relives the accumulated estimation problem. 
We follow the previous two experimental setups shown in Section~\ref{sec:accumulation_ES} and show the results in Figure~\ref{fig:exp1_valval_GNvsBN} and~\ref{fig:exp2_valval_GNvsBN}, respectively. 
%except that we use the network with the BN replaced  odd layers (2n-1)
%Figure 2 shows the results on Setting-up 1 where we replace certain $2n-th$ layer with Group Normalization (We refer to as BN-GN).
 We find that the error gaps between the training and test are significantly reduced in the first 30 epochs from Figure~\ref{fig:exp1_valval_GNvsBN} (a). Importantly, we observe that  $ESM_{\sigma}$ of BN among all layers are nearly the same during training from Figure~\ref{fig:exp1_valval_GNvsBN} (b). This implies that the GN in the odd layer potentially blocks the accumulation of estimation shift of BNs in its two adjacent  layers. 
 %in the GNBN architecture is potentially not affects by other layers. \eg, we can see the $ESM_{\sigma}$ in each layer nearly has the same curvature. 

In Figure~\ref{fig:exp2_valval_GNvsBN}(a), we observe that $ESM_{\sigma}$ of BNs in the `GNBN' is significant lower than the original network (`BN'). Furthermore, there is no remarkable difference for  $ESM_{\sigma}$ of BN among different layers at the end of training. These observations further corroborate that GN can  block the accumulation of estimation shift of BNs in its  two adjacent layers. We attribute this to the consistent operation of GN between training and inference (for each sample) which ensures that the input of later layers have nearly the same  distribution. The blocked accumulation of estimation shift ensures a significantly improved performance for a network, as shown in the comparison of `GNBN' to `BN' in Figure~\ref{fig:exp2_valval_GNvsBN}(b). 

 According to the experiments above, we argue that a BFN (\eg, GN) can  block the accumulation of estimation shift of BN in a network, which can relieve the  performance degeneration of a network if distribution shifts exist.

%\begin{table}[t]
%	\centering
%%	\vspace{-0.1in}
%	\begin{footnotesize}
%	\begin{tabular}{c|ccccc}
%		\bottomrule[1pt]
%		Methods & BN & GN& P1 & P1 & P3    \\
%		\hline
%		Baseline (BN)  &76.29 &76.29&76.29&76.29 &76.29 \\
%		\toprule[1pt]
%	\end{tabular}
%	\vspace{-0.1in}
%	\caption{Results of positions when applying a GN in XBNBlock. We evaluate the top-1 validation accuracy.}
%	\vspace{-0.15in}
%	\label{table:Postion}
%	\end{footnotesize}
%\end{table}

\vspace{-0.05in}
\section{Evaluation on Visual Recognition Tasks}
\vspace{-0.05in}
\label{sec_experiments}

%Based on our analysis, we believe that a network with BN and BFN mixed can improve its BN-only counterpart, since a BFN can relieve the accumulation of \ES~of BN in a network. 
% of the original BN newtwork for its reliveae in estimation shif of BN and robust to distribution shift.
In this section, we first design a kind of convolution block, and  then validate its effectiveness on ImageNet classification~\cite{2015_IJCV_ImageNet}, as well as COCO detection and segmentation~\cite{2014_ECCV_COCO}. 

%  the effeicteve of the propose block in the  ImageNet classificiaont and the object detections. 

%We investigate the effectiveness of our proposed method  
\subsection{Proposed XBNBlock}
\label{sec:XBNBlock}
\begin{figure}[t]
	\centering
	\vspace{-0.1in}
	\hspace{-0.15in}	\subfloat[Original bottleneck]{
		\begin{minipage}[c]{.42\linewidth}
			\centering
			\includegraphics[width=1.8cm]{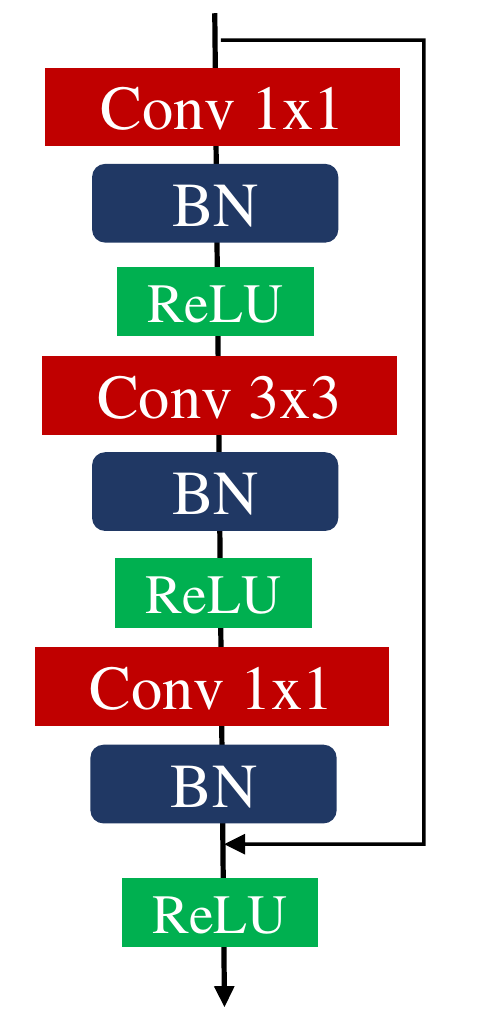}
		\end{minipage}
	}
	\hspace{0.15in}		\subfloat[XBNBlock-P2]{
		\begin{minipage}[c]{.42\linewidth}
			\centering
			\includegraphics[width=1.8cm]{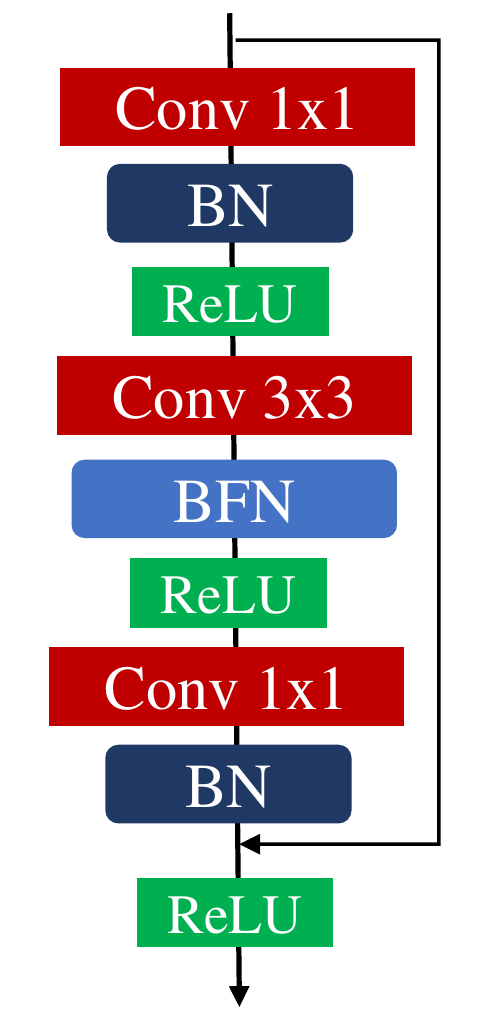}
		\end{minipage}
	}
	\vspace{-0.1in}
	\caption{Bottleneck \vs~ our `XBNBlock-P2' that replaces the second BN of a bottleneck with a  BFN.}
	\label{fig:arch}
	\vspace{-0.1in}
\end{figure}

We design XBNBlock that replaces one BN\footnote{We experimentally find that replacing two BNs with BFNs in the bottleneck usually has worse performance.}  with  BFN in the bottleneck  (Figure~\ref{fig:arch} (a)) which is widely used in the residual-style networks~\cite{2015_CVPR_He,2017_CVPR_Xie}. Figure~\ref{fig:arch} (b) shows the proposed `XBNBlock-P2' in which we replace the second BN layer with BFN. We also consider other positions and compare their performance in Section~\ref{sec:exp_ablation}. 

%Consider the BFN for the convolutional layer.  
 %based on the resent block. We claims one BFN layer be insert into the network may improve the performance of the network. 
%\vspace{-0.1in}
%\paragraph{Convolutional layer.}
For the convolutional input $\mathbf{X} \in \mathbb{R}^{d \times m \times H \times W}$, where $H$ and $W$ are the height and width of the feature maps, BN and BFN used in CNNs both calculate the mean/variance over the $H$ and $W$ dimensions. This paper mainly uses GN as BFN (referred to as XBNBlock$_{GN}$), considering GN is more flexible to control the constraints on the distribution of normalized output by changing its group number~\cite{2021_CVPR_Huang}. We also experiments with IN which calculates the mean/variance only over the $H$ and $W$ dimensions for each channel of a sample, and provides stronger constraints on the normalized output. \Eg, IN ensures the distribution of each channel standardized, while GN ensures the distribution of each group (multiple channels) standardized. 
%\TODO{Show the contrarian number of IN in math.}
%
% consider each spatial position in a feature map as a sample~\cite{2015_ICML_Ioffe} and normalize over the unrolled input $\mathbf{X} \in \mathbb{R}^{d \times m  H  W}$. In contrast, LN and GN  view each spatial position in a feature map as a neuron~\cite{2018_ECCV_Wu} and normalize over the unrolled input $\mathbf{X} \in \mathbb{R}^{d  H  W \times m }$. Following GN, GW also views each spatial position as a neuron, \ie, GW operations (Eqns.~\ref{eqn:GW-1},~\ref{eqn:GW} and~\ref{eqn:GW-2}) are performed for each sample with unrolled input $\mathbf{x} \in \mathbb{R}^{dHW}$.
\begin{table}[t]
	\centering
	%	\vspace{-0.1in}
	\begin{footnotesize}
		\begin{tabular}{c|cc}
			\bottomrule[1pt]
			Methods & Accuracy (\%)     \\
			\hline
			Baseline (BN)  &76.29  \\
			GN &      75.73      \\
			XBNBlock$_{GN}$-P1 &77.08  \\
			XBNBlock$_{GN}$-P2 &\textbf{77.40}  \\
			XBNBlock$_{GN}$-P3 &76.76 \\
			\toprule[1pt]
		\end{tabular}
		\vspace{-0.13in}
		\caption{Results of different positions when applying a GN in XBNBlock. We evaluate the top-1 validation accuracy.}
		\vspace{-0.15in}
		\label{table:Postion}
	\end{footnotesize}
\end{table}

%\vspace{-0.13in}
\subsection{ImageNet Classification}
\label{subsec_imagenet}
\vspace{-0.03in}
We conduct experiments on the ImageNet dataset with 1,000 classes ~\cite{2015_IJCV_ImageNet}.
We use the official 1.28M training images as a training set, and evaluate the top-1 accuracy  on a single-crop of 224$\times$224 pixels in the validation set with 50k images. % We adopt the data augmentation   the \cite{2016_TorchImageNet}
 Our implementation is based on PyTorch~\cite{2017_NIPS_pyTorch}.
We mainly apply our XBNBlock in the ResNet~\cite{2015_CVPR_He} and ResNeXt~\cite{2017_CVPR_Xie} models to validate its effectiveness. Please refer to \TODO{\SM}~\ref{supsec:other-arch} for more results on other architectures.

\begin{figure}[t]
	\centering
	\vspace{-0.1in}
\includegraphics[width=7.0cm]{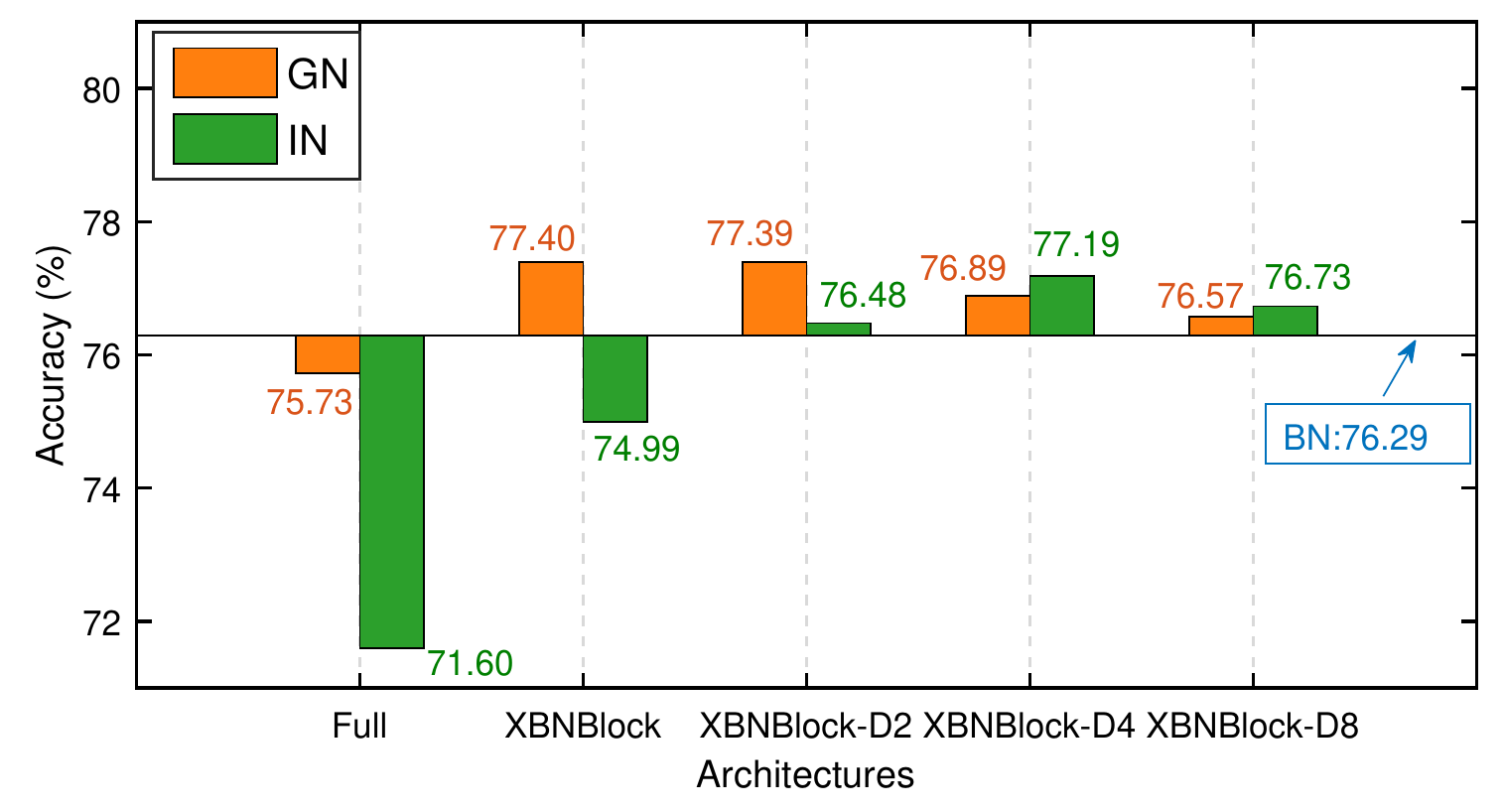}
	\vspace{-0.14in}
	\caption{Top-1 validation accuracy of different positions when applying a XBNBlock in a network. `Full' indicates a network with all the BNs replaced with GN/IN.}
	\label{fig:exp5_Arch}
	\vspace{-0.14in}
\end{figure}

\begin{figure}[t]
	\centering
%	\vspace{-0.08in}
	\includegraphics[width=7.0cm]{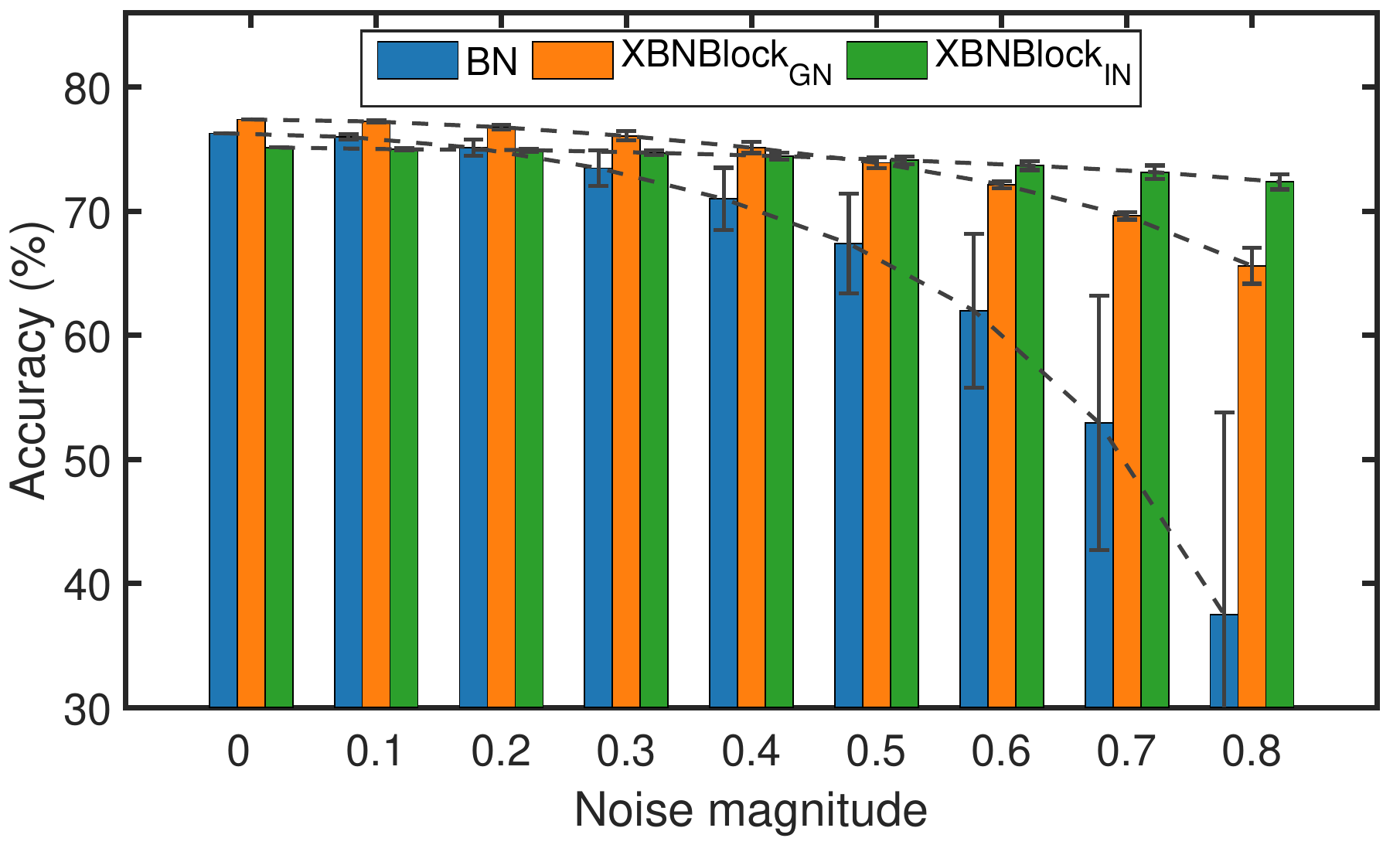}
	\vspace{-0.14in}
	\caption{Top-1 validation accuracy with different noise magnitudes imposed on the estimated population statistics. The results are averaged over 5 random seeds. We refer to a bottleneck/XBNBlock as `disturbed block' if its first BN uses  $\{ \hat{\mu}_{\delta}, \hat{\sigma}^2_{\delta}\}$ for normalization during inference. Here the first six blocks of ResNet-50 are  `disturbed block'. We also perform  experiments using other blocks as `disturbed block' and obtain similar observations (see \TODO{\SM}~\ref{sup_sec:experiments} for details).}
	\label{fig:exp6_RandomPerturb}
	\vspace{-0.17in}
\end{figure}

\begin{table*}[t]
	\vspace{-0.12in}
	\centering
	\begin{small}
	\begin{tabular}{l|cccc}
		\bottomrule[1pt]
		Method     & ResNet-50   &  ResNet-101 &  ResNeXt-50 &  ResNeXt-101\\
		\hline
		Baseline (BN)~\cite{2015_ICML_Ioffe} & ~~76.29 & ~~77.65 & ~~77.06 & ~~79.17  \\
		GN~\cite{2018_ECCV_Wu} & ~~75.73 & ~~77.18 & ~~75.67 & ~~78.02  \\
		IBN-Net*~\cite{2018_ECCV_Pan} & ~~\textbf{77.46} & ~~78.61 & ~~-- & ~~79.12  \\
		SN*~\cite{2019_ICLR_Luo} & ~~76.90 & ~~77.99 & ~~-- & ~~--  \\	
		\hline
		XBNBlock$_{GN}$ (ours) & ~~77.40 & ~~78.21 & ~~\textbf{77.66} & ~~\textbf{79.84}  \\	
		XBNBlock$_{GN}$-D2 (ours) & ~~77.39 & ~~\textbf{78.63} & ~~77.63 & ~~79.72  \\				
		\toprule[1pt]
	\end{tabular}
\vspace{-0.13in}
	\caption{Top-1 accuracy ($\%$) on ResNets~\cite{2015_CVPR_He} and ResNeXts~\cite{2017_CVPR_Xie} for ImageNet. `*' indicates the results are  from the corresponding papers.}
	%Here, we apply GW and BW$_\Sigma$ in these models following the \textbf{S1-B2} architecture.}
\label{table:ImageNet-Res-Step}
\vspace{-0.17in}
	\end{small}
\end{table*}

\begin{table}[t]
	\centering
	\vspace{-0.05in}
	\begin{footnotesize}
		\begin{tabular}{c|cccc}
			\bottomrule[1pt]
			 &  LS  &     MixUp & COS& LS~+~MixUP~+~COS\\
			\hline
			 Baseline (BN)  &76.70 &76.75 &76.72 &77.16\\
			XBNBlock$_{GN}$ &\textbf{77.41} &\textbf{77.70}   &\textbf{77.60} &\textbf{78.22}  \\
			\toprule[1pt]
		\end{tabular}
		\vspace{-0.13in}
		\caption{Top-1 accuracy ($\%$) on ResNet-50 using advanced training strategies. `LS' indicates label smoothing and `COS' indicates cosine learning rate decay.}
		\vspace{-0.2in}
		
		\label{table:Advanced-Strategy}
	\end{footnotesize}
\end{table}

%\begin{table}[t]
%	\centering
%	\vspace{-0.05in}
%	\begin{footnotesize}
%	\begin{tabular}{c|cc}
%		\bottomrule[1pt]
%		Training strategies & Baseline (BN)    & XBNBlock$_{GN}$      \\
%		\hline
%		label smooth (LS)  &76.70 &\textbf{77.41} \\
%		MixUp &76.75 &\textbf{77.70}  \\
%		cosine learning (COS) &76.72 &\textbf{77.60}  \\
%		LS~+~MixUP~+~COS &77.16 &\textbf{78.22}  \\
%		\toprule[1pt]
%	\end{tabular}
%	\vspace{-0.1in}
%	\caption{Top-1 accuracy ($\%$) on ResNet-50 using advanced training strategies.}
%		\vspace{-0.2in}
%		
%	\label{table:Advanced-Strategy}
%\end{footnotesize}
%\end{table}

\vspace{-0.1in}
\subsubsection{Ablation Studies on ResNet-50}
\label{sec:exp_ablation}
\vspace{-0.05in}
%\paragraph{Experimental setup.}
We adopt the widely used training protocol to compare the performance of ResNets/ResNeXt for ImageNet classification~\cite{2015_CVPR_He}:  we apply stochastic gradient descent (SGD) using a mini-batch size of 256, momentum of 0.9 and weight decay of 0.0001. We train over 100 epochs. The initial learning rate is set to 0.1 and divided by 10 at 30, 60  and 90 epochs. Our baseline is the ResNet-50 trained with BN~\cite{2015_ICML_Ioffe}.

\vspace{-0.15in}
\paragraph{Positions of BFN in an XBNBlock.}
We investigate the position where to apply BFN in an XBNBlock. We use GN (with group number $g$=64)\footnote{We also try other group numbers shown in the \TODO{\SM}~\ref{sup_sec:experiments}.} as BFN. We consider three XBNBlock variants that replace the first, second and third BN in the bottleneck and refer to them as `XBNBlock-P1', `XBNBlock-P2' and `XBNBlock-P3', respectively. We substitute these XBNBlocks for all the bottlenecks of ResNet-50 and report the results in Table~\ref{table:Postion}.
We can  see that all the networks with XBNBlock outperform the baseline  by a clear margin. Note that the network in which all the BNs are replaced with GNs has only $75.63\%$ validation accuracy, which is worse than the baseline. This result implies that the estimation shift of BN probably exists due to the accumulation of multiple stacked BNs, even for training under a moderate batch-size. GN can block the accumulation of estimation shift of BN and thus improve the performance of the network with BN. 
We observe that `XBNBlock-P2' obtain the best performance, and we refer to `XBNBlock-P2' (Figure~\ref{fig:arch}) as our XBNBlock by default in the following experiments. 

% ResNet and ResNeXt are both composed primarily of a stem layer and multiple bottleneck blocks~\cite{2015_CVPR_He}. We consider:  1) replacing the BN in the stem layer with GW (referred to as `S1'); and 2) replacing the $1^{st}$, $2^{nd}$, $1^{st}~\&~2^{nd}$, and $3^{rd}$ BNs in all the bottleneck blocks, which are referred to as `B1', `B2', `B12' and `B3', respectively. We investigate five architectures, \textbf{S1}, \textbf{S1-B1}, \textbf{S1-B2}, \textbf{S1-B3} and \textbf{S1-B12}, which have 1, 17, 17, 17 and 33 GW modules, respectively.
%We also perform experiments using BW~\cite{2019_CVPR_Huang} and BW$_\Sigma$~\cite{2020_CVPR_Huang} (employing a covariance matrix to estimate the population statistics of BW) for contrast.

\vspace{-0.15in}
\paragraph{Positions of XBNBlock in a network.}
We also investigate the positions where to apply XBNBlock in a network. There are 16 bottlenecks in ResNet-50, and we consider three variants to alternatively substitute  XBNBlocks for the bottlenecks: (1) XBNBlock-D2: the $\{2n, n=1,2,...,8\}$-th bottlenecks are replaced with XBNBlocks; (2)  XBNBlock-D4:  the $\{4n, n=1,2,3,4\}$-th bottlenecks are replaced with XBNBlocks; (3) XBNBlock-D8:  the $\{8n, n=1,2\}$-th bottlenecks are replaced with XBNBlocks. We investigate GN and IN in the XBNBlock and refer to as `XBNBlock$_{GN}$' and `XBNBlock$_{IN}$'. The results are shown in Figure~\ref{fig:exp5_Arch}. 
We observe that all the `XBNBlock$_{GN}$' models have better validation accuracy than the baseline, and a network with fewer XBNBlock$_{GN}$ has worse performance. 
We also find that `XBNBlock$_{IN}$-D4' obtains a  validation accuracy of $77.19\%$,  better than the baseline ( $76.29\%$) while  XBNBlock$_{IN}$ has only  $74.99\%$. We attribute this phenomenon to that IN provides stronger constraints on the normalized output, which can affect the representation ability of the model, \eg, XBNBlock$_{IN}$ has only a training accuracy of $77.93\%$, significantly lower than the baseline with $80.29\%$ training accuracy.  In the following section, we show that such constraints make a model more robust. 

%give a more stable distribution to block the accumulation of estimation shift. However, such constraints
% considering that IN provides stronger constraints on the normalized output and 
% to see how the BFNs, that provide different magnitudes in constraints on the normalized output, affect the performance of a network. We use GN and IN 

\vspace{-0.15in}
\paragraph{Robustness to distribution shift.}
As discussed in Section~\ref{sec:accumulation_block_ES}, a BFN can block the accumulation of the estimation shift of BN, which suggests that a model with BFN could be more robust than the distribution shift. We design experiments to validate this arguments. We disturb the estimated mean and variance of BN as follows:
 \begin{small}
%	\setlength\abovedisplayskip{0.05in} 
%	\setlength\belowdisplayskip{0.04in}
	%	 \vspace{-0.2in}
\begin{equation}
	\label{eqn:BN-disturb}
	\begin{aligned}
		\begin{cases}
			\quad \hat{\mu}_{\delta} =  (1+\delta_{\mu})\hat{\mu},~~~~\delta_{\mu} \sim uniform (-\Delta, \Delta)\\
			\quad \hat{\sigma}^2_{\delta}=  (1+\delta_{\sigma}) \hat{\sigma}^2,  ~~~~\delta_{\sigma} \sim uniform (-\Delta, \Delta),
		\end{cases}
	\end{aligned}
\end{equation}
\end{small}
\hspace{-0.05in}where $\Delta$ represents  noise magnitude. 
%We mainly disturb the first BN of bottleneck/XBNBlock, and disturb the first 6 block so
Figure~\ref{fig:exp6_RandomPerturb} shows that the baseline (`BN') has significantly reduced validation accuracy as noise magnitude increases, while XBNBlock$_{GN}$/XBNBlock$_{IN}$ is more stable for such a disturbance. This suggests that the consistent normalization operations during training and inference of a BFN can potentially reduce the distribution shift in a layer and improve the robustness of models. We also note that XBNBlock$_{IN}$ is more robust than XBNBlock$_{GN}$, we  attribute this to that   IN indeed provides stronger constraints than GN on the normalized output, which gives a more stable distribution to prevent the distribution shift. 

% Here, we conduct experiment to validate the robustness of the proposed XBNBlock to distribution shift. We also investigate the effect of inserting a GW/BW layer after the last average pooling (before the last linear layer) to learn the decorrelated feature representations, as proposed in \cite{2019_CVPR_Huang}.  This can slightly improve the performance ($0.10\%$ on average) when using GW, though the net gain is smaller than using BW ($0.22\%$) or BW$_\Sigma$ ($0.43\%$). Please refer to the \TODO{\SM} for details.
%The results are show in Figure ~\ref{fig:exp_position} (b).

\begin{table*}[t]
	\centering
	\vspace{-0.18in}
	\begin{small}
	\begin{tabular}{c|ll|ll|ll|ll}
		\bottomrule[1pt]
		& \multicolumn{4}{c| }{ResNet-50} &    \multicolumn{4}{c}{ResNext-101}  \\
		& \multicolumn{2}{c| }{2fc head box} &    \multicolumn{2}{c|}{4conv1fc head box} & \multicolumn{2}{c| }{2fc head box} &    \multicolumn{2}{c}{4conv1fc head box}  \\
		\hline
		Method     & AP$^{bbox}$   & AP$^{mask}$ & AP$^{bbox}$& AP$^{mask}$  & AP$^{bbox}$   & AP$^{mask}$ & AP$^{bbox}$& AP$^{mask}$  \\
		\hline
		%BN*   &  36.90 & 58.60 & 40 &  36.85 & 57.73 & 39.93  \\
		$BN^{\dag}$  &37.40 &34.01 &37.51  &33.68  &42.13 &37.78 &42.24  &37.53  \\	
		GN  &37.55 &34.06 &39.02  &34.37  &41.47 &37.17 &42.18  &37.53  \\
		XBNBlock$_{GN}$  &\textbf{38.19} &\textbf{34.57} &\textbf{39.57}  &\textbf{34.86}  &\textbf{42.69} &\textbf{38.00} &\textbf{43.43}  &\textbf{38.68}  \\
		\toprule[1pt]
	\end{tabular}
	\vspace{-0.13in}
	\caption{Detection and segmentation results ($\%$) on COCO using the Mask R-CNN framework implemented in~\cite{massa2018mrcnn}. Models based on ResNet-50 backbone are trained by 1x lr scheduling (90k iterations), with a batch size of 16 on eight GPUs. Models based on ResNeXt-101 backbone are trained by 1x lr scheduling (180k iterations), with a batch size of 8 on eight GPUs. }
		\vspace{-0.1in}
	\label{table:Mask-RCNN}
	\end{small}
\end{table*}
\vspace{-0.1in}
\subsubsection{Experiments on Larger Models}
\label{sec:larger-model}
\vspace{-0.05in}
We validate the effectiveness of XBNBlock on ResNet-101~\cite{2015_CVPR_He}, ResNeXt-50 and ResNeXt-101~\cite{2017_CVPR_Xie}. 
The baselines are the original networks trained with BN, and we also train the models with GN.
The results are shown in Table~\ref{table:ImageNet-Res-Step}. We can see that our method consistently improves the baseline (BN) by a significant margin over all architectures. Our method obtains comparable performance to IBN-Net~\cite{2018_ECCV_Pan}. Note that IBN-Net carefully designs the position of IN in a network and its channel number, while the design of our XBNBlock is simplified. We argue  our observation, that a BFN (\eg, IN) can block the accumulation of estimation shift of BN, also provides a reasonable explanation to the success of IBN-Net in its good performance, especially in the scenarios with distribution shift (\eg, domain adaptation and transfer learning tasks.~\cite{2018_ECCV_Pan}).

\vspace{-0.15in}
\paragraph{Advanced training strategies.}
Besides the standard training strategy described in Section~\ref{sec:exp_ablation}, we also conduct experiments using more advanced training strategies: 1) cosine learning rate decay with 100 epochs trained~\cite{2017_ICLR_Loshchilov}; 2) label smoothing~\cite{2019_CVPR_He} with a smoothing factor of 0.1; 3) mixup~\cite{2018_ICLR_Zhang} training with a mix factor of 0.2. XBNBlock also consistently outperforms the baseline by a significant margin. Table~\ref{table:Advanced-Strategy} shows the results on ResNet-50 and please see \TODO{\SM}~\ref{sup_sec:experiments} for results on ResNet-101 and ResNeXt-50. 
% and GW again improves the baseline consistently. Please refer to the \TODO{\SM} for details.
%3) GN has worse perfomance than BN, especially on ResNeXt models.
\vspace{-0.15in}
\paragraph{Towards whitening.} Note that our method can also use the recently proposed group whitening (GW)~\cite{2021_CVPR_Huang} as a BFN. By applying GW in our design, our XBNBlock outperforms the state-of-the-art normalization (whitening) methods. \Eg, our method obtains  validation accuracy of $79.18\%$ on ResNet-101, compared to the baseline (BN) of $77.65\%$, with a gain of $1.53\%$. Please see the \TODO{\SM}~\ref{sup_sec:experiments}  for details.  
% to form a more beneficial results. By using group whitening into our design, our method can obtain comparable performance to the state-of-art hwitening methods,  

\begin{table}[t]
	\centering
	\vspace{-0.12in}
	\begin{footnotesize}
	\begin{tabular}{c|cc}
		\bottomrule[1pt]
		Method     & 2fc head box   & 4conv1fc head box    \\
		\hline
		$BN^{\dag}$  &36.31 &36.85  \\
		GN  &36.62 &37.86  \\
		XBNBlock$_{GN}$ &\textbf{37.17} &\textbf{38.47}  \\
		\toprule[1pt]
	\end{tabular}
\vspace{-0.15in}
	\caption{Detection results ($\%$) on COCO using the Faster R-CNN framework implemented in~\cite{massa2018mrcnn}. We use ResNet-50 as the backbone, combined with FPN. All models are trained by 1x lr scheduling (90k iterations), with a batch size of 16 on eight GPUs.}
	\vspace{-0.17in}
	\label{table:Faster-RCNN}
	\end{footnotesize}
\end{table}

%Here we denote the corsponding whitneing algorothima as ZCA$_{\CM{}}^L$
\vspace{0.1in}
\subsection{Detection and Segmentation on COCO}
\label{sec_Coco}
\vspace{-0.05in}
We conduct experiments for object detection and segmentation on the COCO benchmark~\cite{2014_ECCV_COCO}. 
 We use the Faster R-CNN \cite{2015_NIPS_Ren} and Mask R-CNN \cite{2017_ICCV_He} frameworks based on the publicly available codebase `maskrcnn-benchmark'~\cite{massa2018mrcnn}.
We train the models on the COCO $train2017$ set and evaluate on the COCO $val2017$ set. We report the standard COCO metrics of average precision (AP) for bounding box detection (AP$^{bbox}$) and instance segmentation (AP$^{mask}$)~\cite{2014_ECCV_COCO}.
We experiment with both fine-tuning from pre-trained models and training from scratch. 
\vspace{-0.1in}
\subsubsection{Fine-tuning from Pre-trained Models}
\vspace{-0.05in}
In this section, we fine-tune the models trained on ImageNet for object detection and segmentation on the COCO benchmark~\cite{2014_ECCV_COCO}. For BN, we use its frozen version (indicated by $BN^{\dag}$) when fine-tuning for object detection~\cite{2018_ECCV_Wu}.
%These computer vision tasks in general benefit from higher-resolution input, so the batch size tends to be small in common practice.
\vspace{-0.15in}
\paragraph{Object detection using Faster R-CNN.}
We use Faster R-CNN framework for object detection and use the  ResNet-50 models pre-trained on ImageNet (Table~\ref{table:ImageNet-Res-Step}) as the backbones, combined with the feature pyramid network (FPN)~\cite{2017_CVPR_Lin}. We consider two setups: 1) we use the  box head consisting of two fully connected layers (`2fc') without a normalization layer, as proposed in~\cite{2017_CVPR_Lin}; 2) following~\cite{2018_ECCV_Wu}, we replace the `2fc' box head with `4conv1fc' and apply GN to the FPN and box head for both `GN' and our `XBNBlock$_{GN}$'.
We use the default  hyperparameter configuration  from the training scripts provided by the codebase~\cite{massa2018mrcnn} for Faster R-CNN.
The results are reported in Table~\ref{table:Faster-RCNN}.
The XBNBlock pre-trained model consistently outperform  $BN^{\dag}$ and GN by a remarkable margin. \Eg, XBNBlock$_{GN}$ obtains $38.47\%$ AP under the setup of `4conv1fc' head box, compared to the baseline of $36.85\%$, with a gain of $1.62\%$.

\vspace{-0.15in}
\paragraph{Results on Mask R-CNN.} 
We use Mask R-CNN framework for object detection and instance segmentation. We use both the ResNet-50 and the ResNeXt-101~\cite{2017_CVPR_Xie} models pre-trained  on  ImageNet (Table~\ref{table:ImageNet-Res-Step}) as the backbones, combined with FPN. We consider both the `2fc' and `4conv1fc' setups.  We again use the default hyperparameter configuration from the training scripts provided by the codebase for Mask R-CNN~\cite{massa2018mrcnn}.
The results are shown in Table~\ref{table:Mask-RCNN}. 
The XBNBlock pre-trained model consistently outperforms  $BN^{\dag}$ and GN by a significantly margin, over both the backbones and setups.
% \Eg, XBNBlock obtains $38.47\%$ AP under the setup of `4conv1fc' head box, compared to baseline with $36.85\%$, a gains of $1.62\%$. 
%GW achieves $44.41\%$ in box AP and $39.17\%$ in mask AP, an improvement over $BN^{\dag}$  of $2.17\%$ and $1.64\%$, respectively.
%, \eg, we train the model for $180k$, with batch size of 8 on 8 GPUs
%Mainly benefits finetuning.
%
%\TODO{We also provides the experiments on ResNet-50 like Faster R-CNN in the \SM}
\vspace{-0.12in}
\subsubsection{Training from Scratch}

%One notorious problem of BN is its small-batch-size problem.
One main concern for XBNBlock  is that it cannot work well under small-batch-size training scenarios, due to the exist of BNs.
Here, we train  Faster R-CNN from scratch and use normal BN which is not frozen. We use ResNet-50 as the backbone and follow the same setup as in the previous experiment, except that:1) we vary the batch size (BS) in $\{2, 4, 8\}$ on each GPU; 2) we search the learning rate in $\{0.01, 0.02, 0.04\}$\footnote{The default learning rate is 0.02 and we do it for that the model with GN-only cannot obtain a reasonable result if the learning rate is not appropriate for certain BS, while the model using BN/XBNBlock has no such a problem.} considering that  BS varies, and report the best performance. Table~\ref{table:trainFromScratch} shows the results. We can see XBNBlock$_{GN}$ obtains significantly better performance than BN and GN under the batch size of 4 and 8. Under the batch size of 2,  even though XBNBlock$_{GN}$ has slightly worse  performance than GN, it significantly outperforms BN by a gain of $2.1\%$ AP. We believe that the small-batch-size problem of BN may consist of: 1) the inaccurate estimation between  training and inference distribution of a BN layer; 2) the accumulated estimation shift of BNs in a network. We argue that the GN in XBNBlock blocks the accumulation of estimation shift, thus mitigates the small-batch-size problem of the BNs in a network.

\begin{table}[t]
	\centering
	\vspace{-0.12in}
	\begin{footnotesize}
		\begin{tabular}{c|lll}
			\bottomrule[1pt]
			Method     & $BS=2$   & $BS=4$ & $BS=8$ \\
			\hline
			BN  & 25.35 & 29.33 & 29.56  \\
			GN  &  \textbf{28.19} & 27.36 & 28.22 \\
			XBNBlock$_{GN}$ & 27.45 & \textbf{30.51} & \textbf{30.58} \\
			%46.00$_{(\textcolor{red}{\downarrow 0.19})}$ & 37.54$_{(\textcolor{blue}{\uparrow 0.01})}$ & 60.18$_{(\textcolor{blue}{\uparrow 0.36})}$ & 39.99$_{(\textcolor{blue}{\uparrow 0.03})}$  \\	
			\toprule[1pt]
		\end{tabular}
		\vspace{-0.15in}
		\caption{Detection results ($\%$) on COCO by training from scratch. We use ResNet-50 as the backbone, combined with FPN. All models are trained by 1x lr scheduling (90k iterations) on eight GPUs, with a varying batch size (BS) in $\{2, 4, 8\}$ on each GPU. }
		%	\vspace{-0.1in}
		\label{table:trainFromScratch}
		%	\vspace{0.1in}
	\end{footnotesize}
	\vspace{-0.17in}
\end{table}

\vspace{-0.02in}
\section{Conclusion}
\vspace{-0.02in}
This paper found that the estimation shift of BN can be accumulated in a network, which can lead to a detriment effect for a network during test, and that a batch-free normalization can block such accumulation of estimation shift, which can relieve the  performance degeneration of a network if distribution shifts occur. These observations can potentially contribute to understanding the application of normalization in different scenarios, and designing architectures for better performance. 
We believe our designed XBNBlock is a practical method that has potentialities to be used in broader architectures and  applications. 

\noindent\textbf{Acknowledgement}
This work was partially supported  by the National Key Research and Development Plan of China under Grant 2021ZD0112901, National Natural Science Foundation of China (Grant No. 62106012, 61972016 and 62106043).

%%%%%%%%% REFERENCES
{\small
\bibliographystyle{ieee_fullname}
\bibliography{2GBN}
}

\appendix
\clearpage
\renewcommand{\thetable}{A\arabic{table}}
\setcounter{table}{0}

\renewcommand{\thefigure}{A\arabic{figure}}
\setcounter{figure}{0}

\section{Investigation of Estimation Shift on MNIST}
\label{supsec:investigation}
\begin{figure*}[h]
	\centering
	%	\vspace{-0.08in}
	\hspace{-0.15in}	\subfloat[Training and test errors]{
		\begin{minipage}[c]{.30\linewidth}
			\centering
			\includegraphics[width=5.4cm]{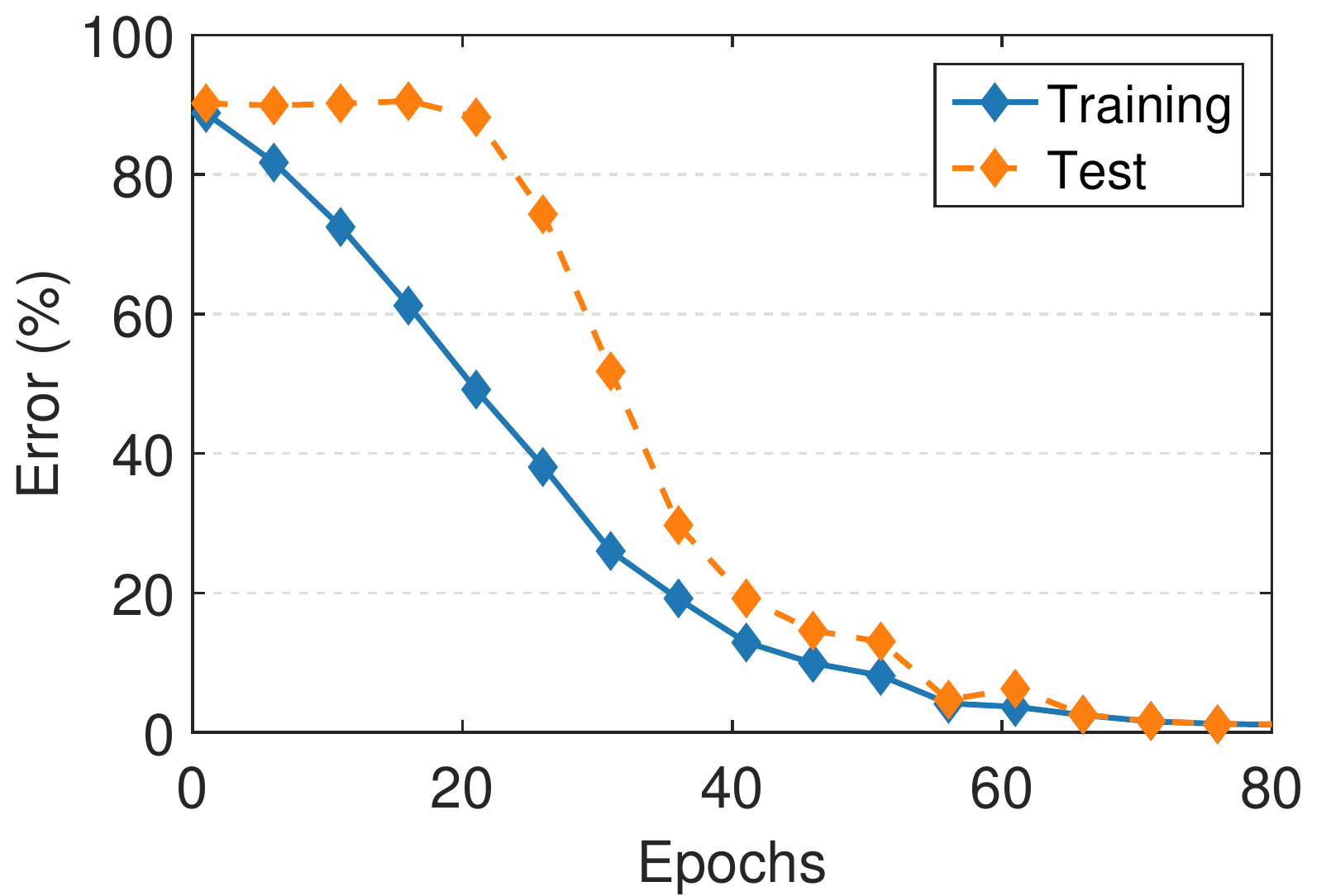}
		\end{minipage}
	}
	\hspace{0.15in}		\subfloat[ESM$_\mu$ of BN in different layers]{
		\begin{minipage}[c]{.30\linewidth}
			\centering
			\includegraphics[width=5.4cm]{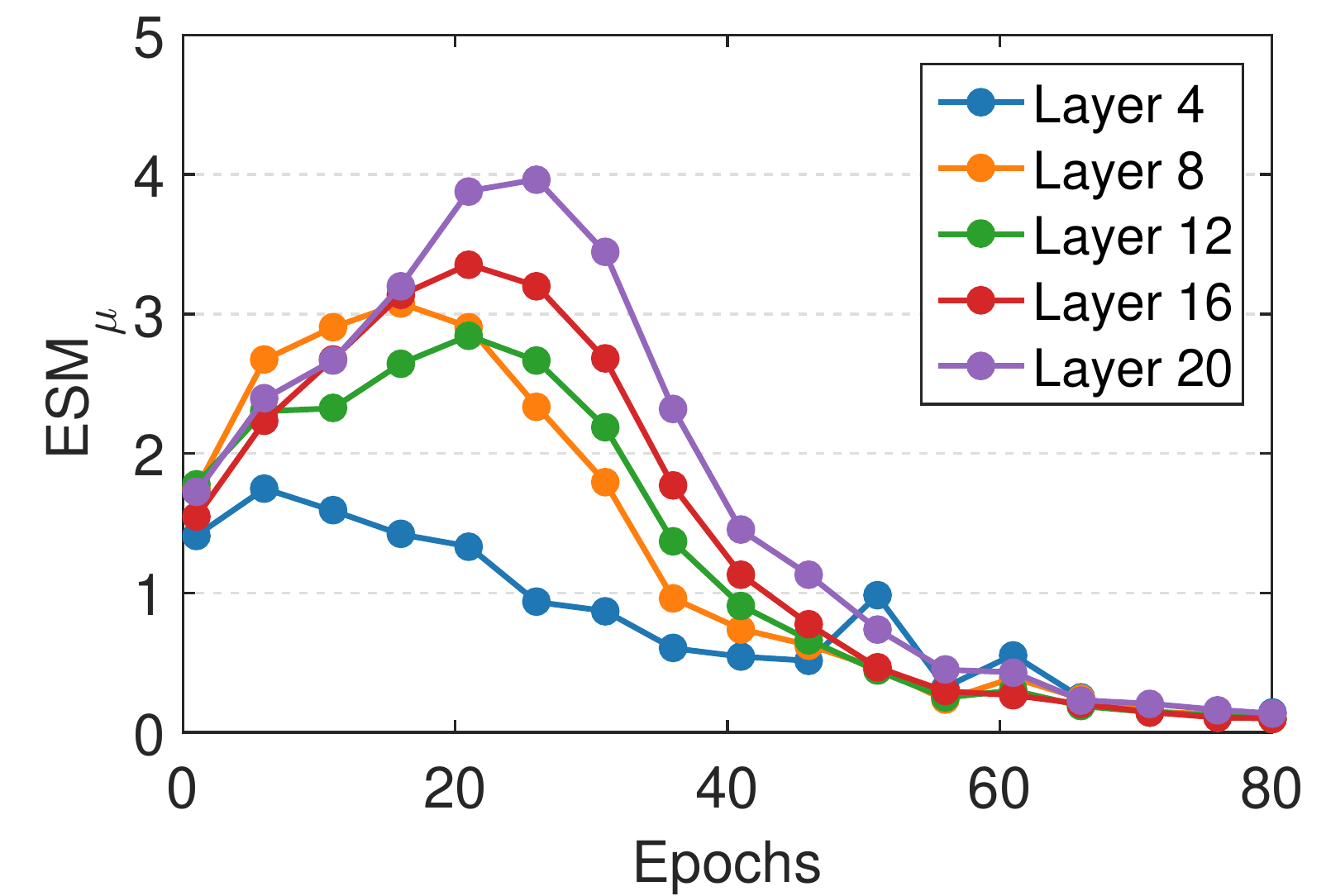}
		\end{minipage}
	}
	\hspace{0.15in}		\subfloat[ESM$_\sigma$ of BN in different layers]{
		\begin{minipage}[c]{.30\linewidth}
			\centering
			\includegraphics[width=5.4cm]{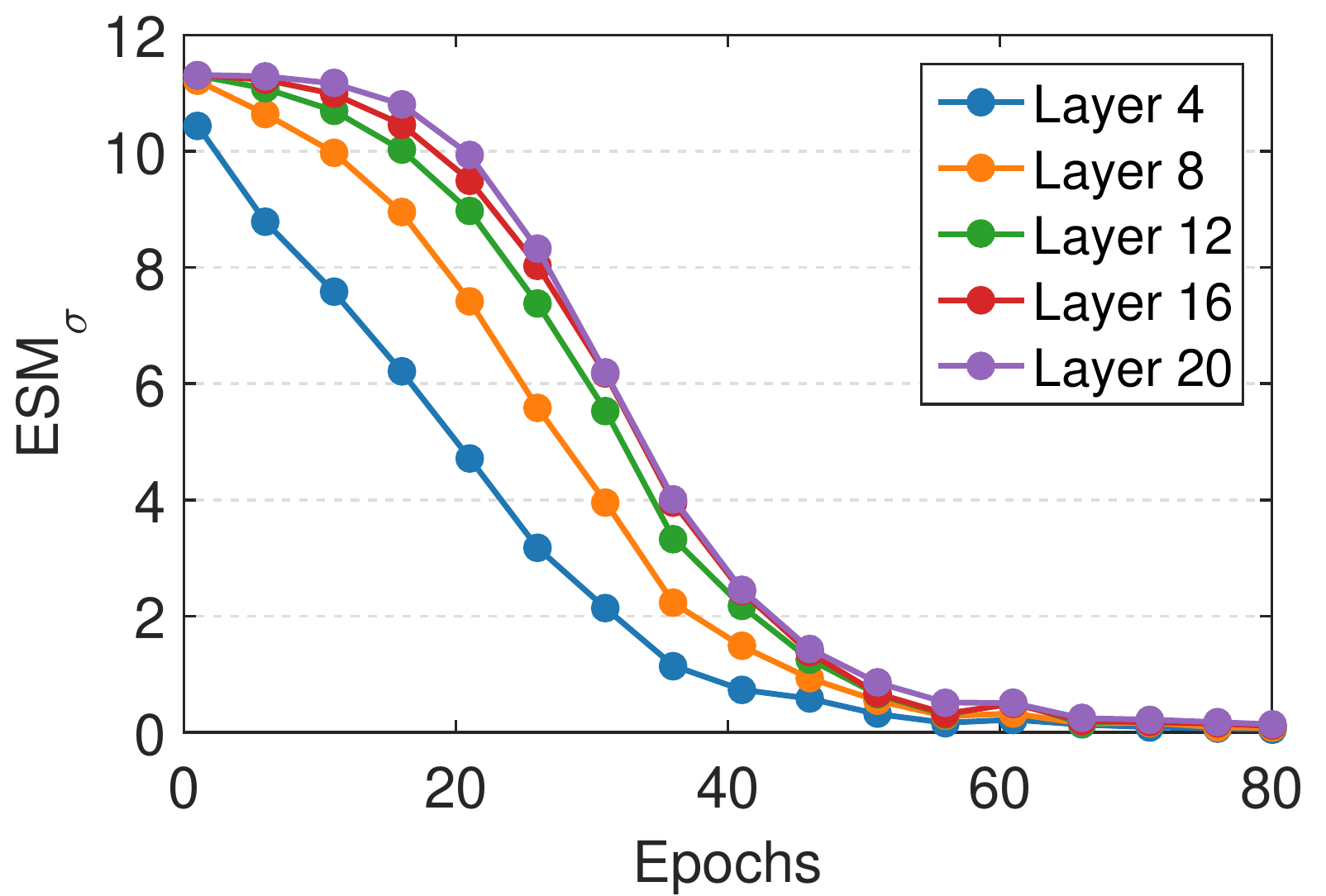}
		\end{minipage}
	}
	%\vspace{-0.08in}
	\caption{Experiments with training set $\mathbf{S}$ equaling to the test set  $\mathbf{S'}$. We follow the same experimental setup as the one in Figure~\ref{fig:exp1_valval} of the paper, except that we use  a learning rate of 0.05. }
	\label{sup_fig:exp1_valval_LR05}
	%	\vspace{-0.17in}
\end{figure*}

\begin{figure*}[h]
	\centering
	%	\vspace{-0.08in}
	\hspace{-0.15in}	\subfloat[Training and test errors]{
		\begin{minipage}[c]{.30\linewidth}
			\centering
			\includegraphics[width=5.4cm]{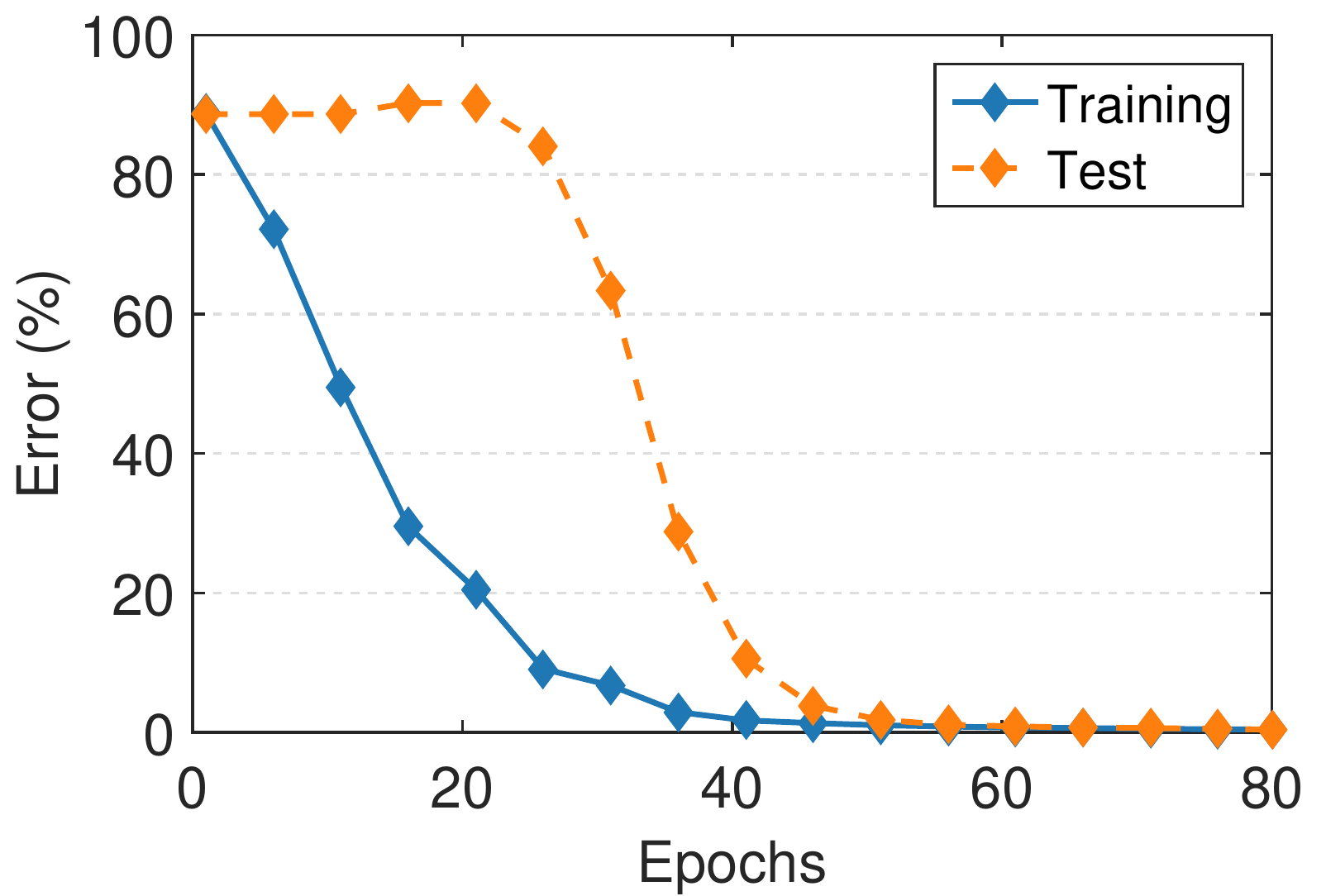}
		\end{minipage}
	}
	\hspace{0.15in}		\subfloat[ESM$_\mu$ of BN in different layers]{
		\begin{minipage}[c]{.30\linewidth}
			\centering
			\includegraphics[width=5.4cm]{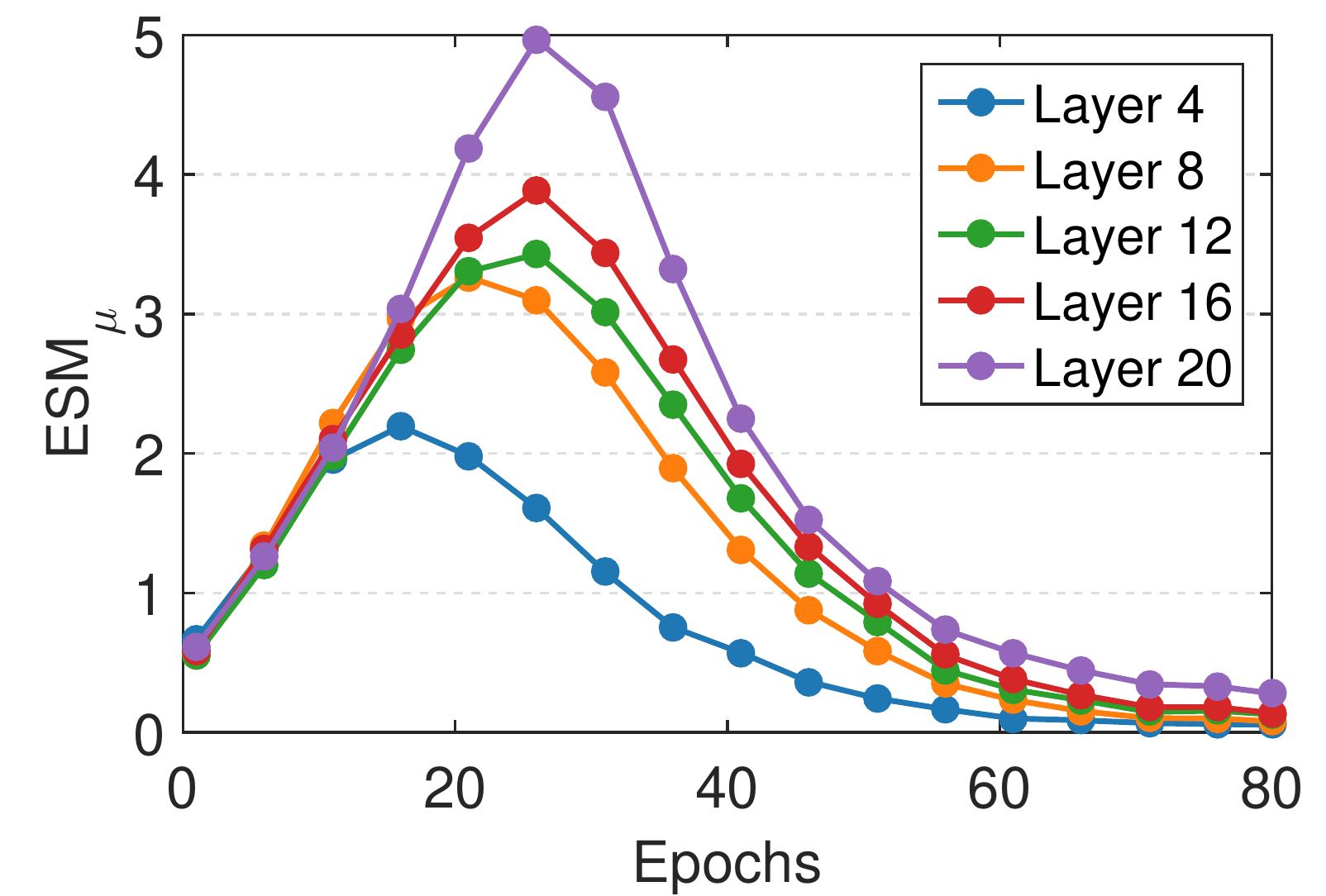}
		\end{minipage}
	}
	\hspace{0.15in}		\subfloat[ESM$_\sigma$ of BN in different layers]{
		\begin{minipage}[c]{.30\linewidth}
			\centering
			\includegraphics[width=5.4cm]{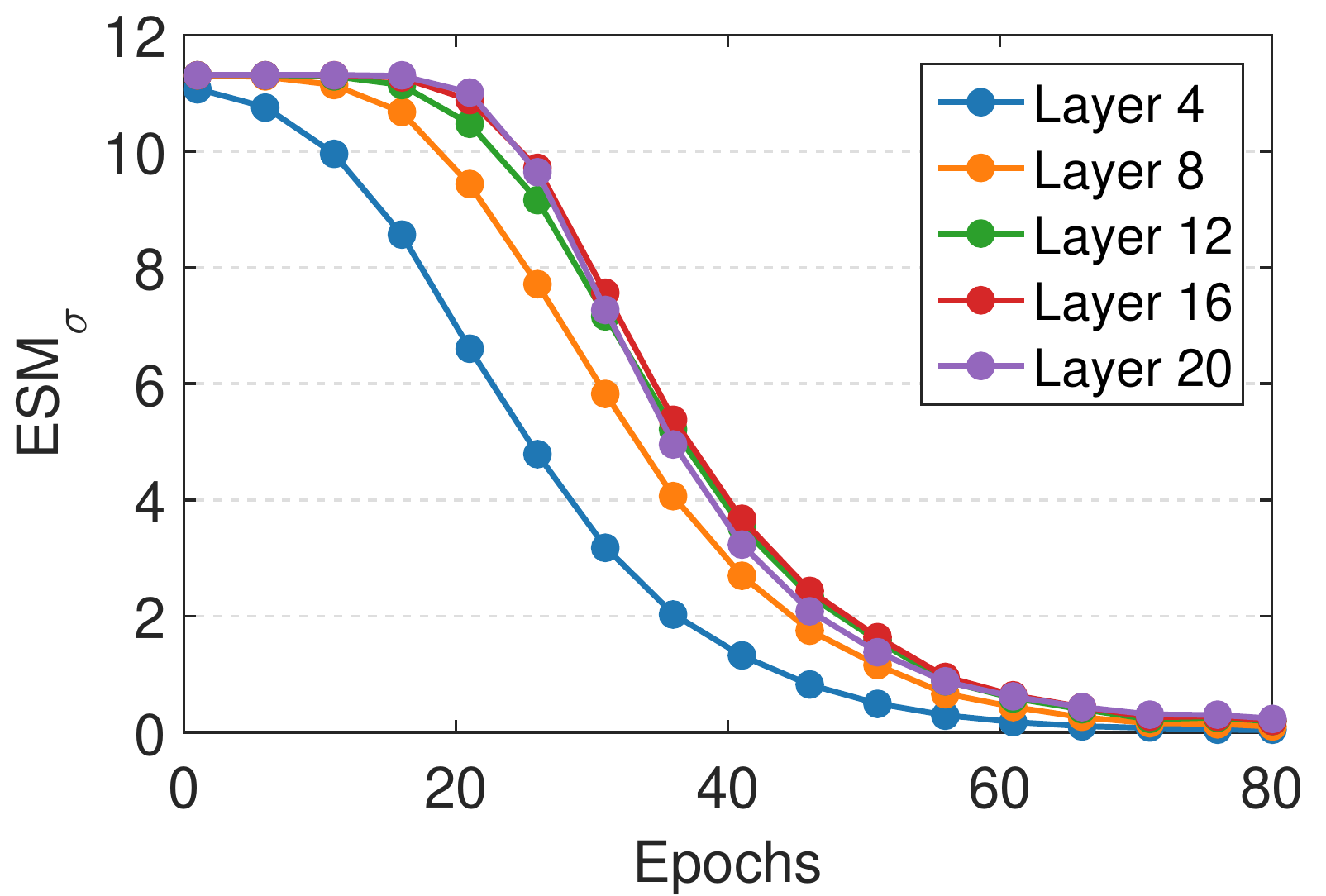}
		\end{minipage}
	}
	%\vspace{-0.08in}
	\caption{Experiments with training set $\mathbf{S}$ equaling to the test set  $\mathbf{S'}$. We follow the same experimental setup as the one in Figure~\ref{fig:exp1_valval} of the paper, except that we use an update factor of $\alpha=0.1$ for the running average in calculate population statistics of BN.}
	\label{sup_fig:exp1_valval_MM01}
	%	\vspace{-0.17in}
\end{figure*}

\begin{figure*}[h]
	\centering
	%	\vspace{-0.08in}
	\hspace{-0.15in}	\subfloat[Training and test errors]{
		\begin{minipage}[c]{.30\linewidth}
			\centering
			\includegraphics[width=5.4cm]{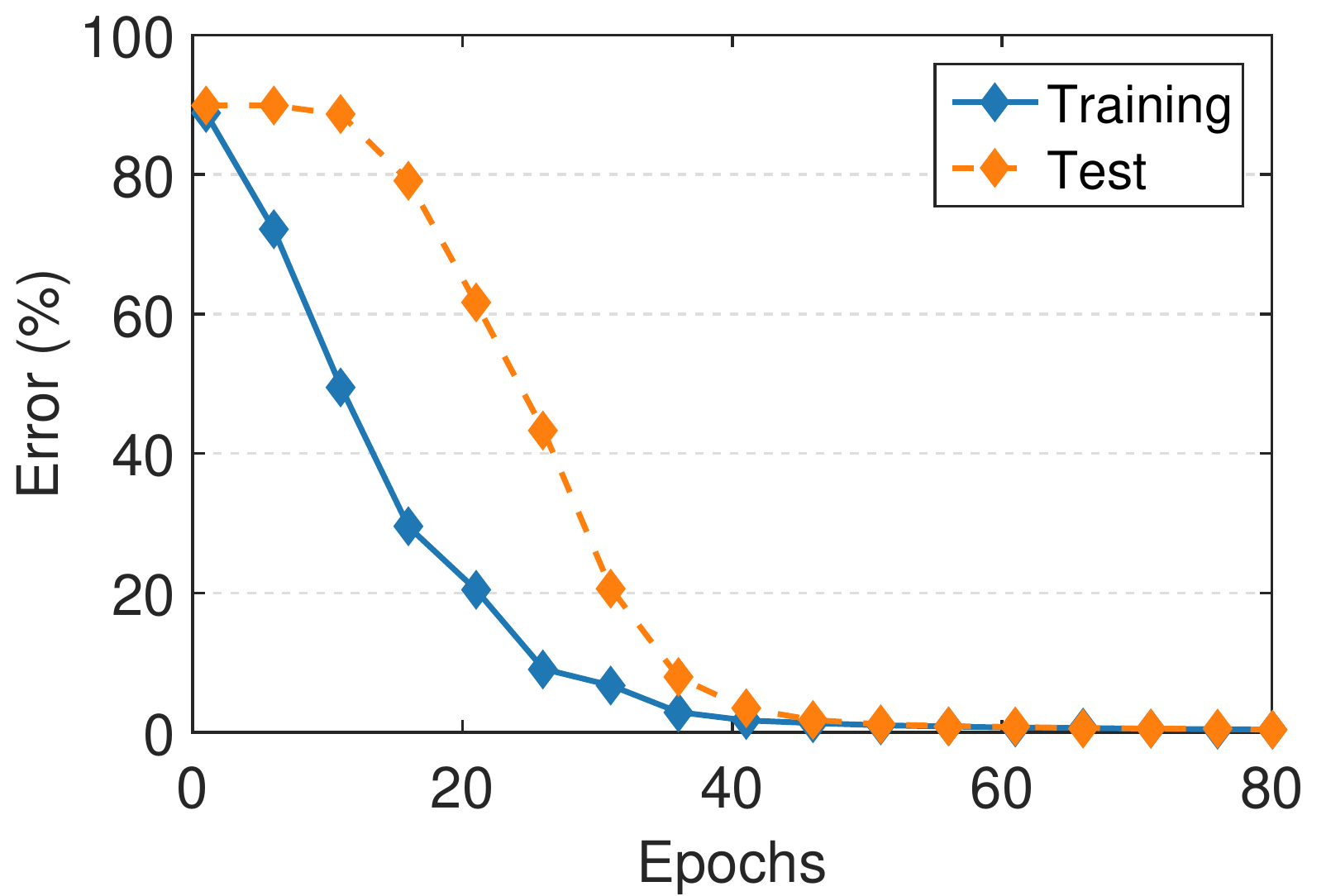}
		\end{minipage}
	}
	\hspace{0.15in}		\subfloat[ESM$_\mu$ of BN in different layers]{
		\begin{minipage}[c]{.30\linewidth}
			\centering
			\includegraphics[width=5.4cm]{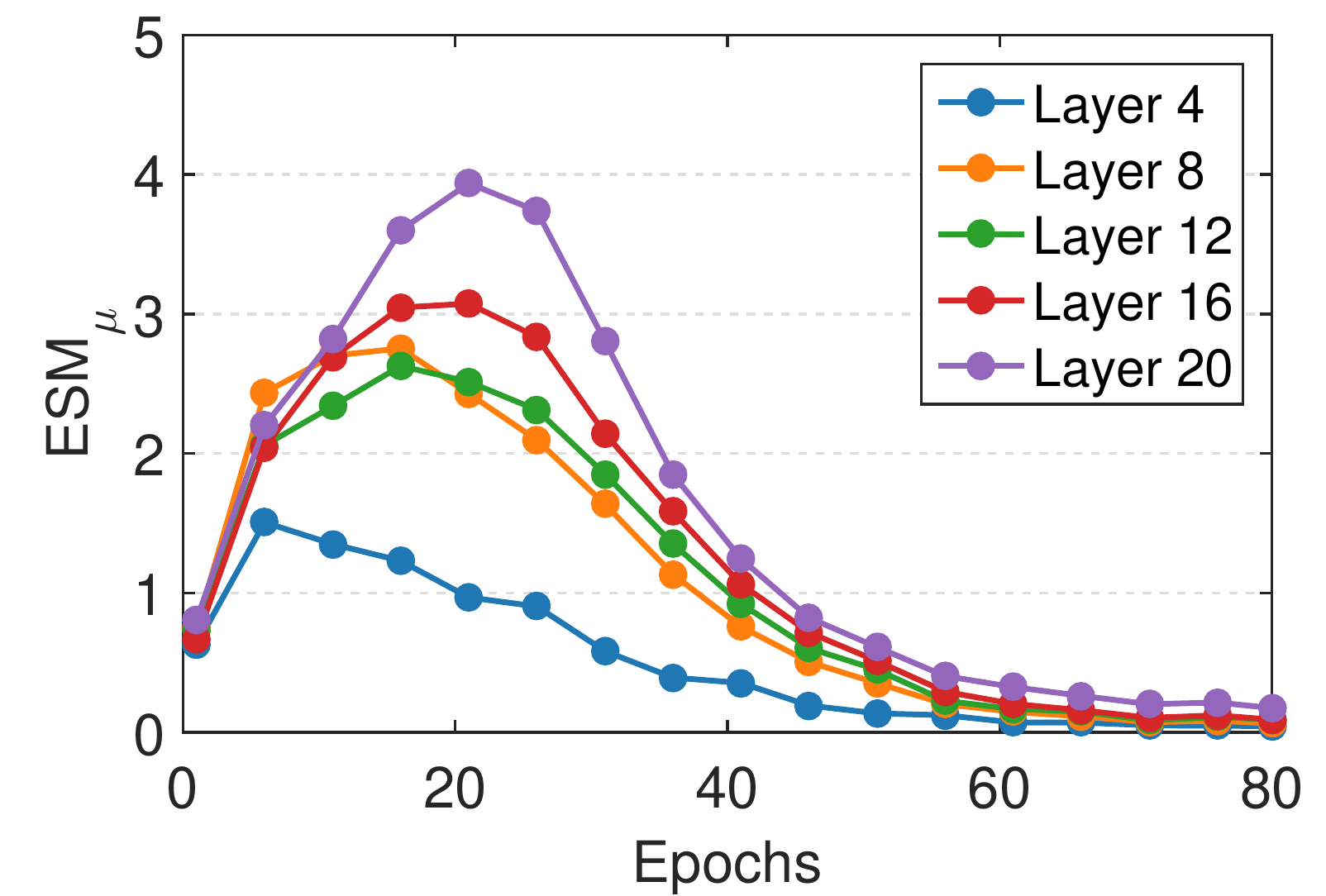}
		\end{minipage}
	}
	\hspace{0.15in}		\subfloat[ESM$_\sigma$ of BN in different layers]{
		\begin{minipage}[c]{.30\linewidth}
			\centering
			\includegraphics[width=5.4cm]{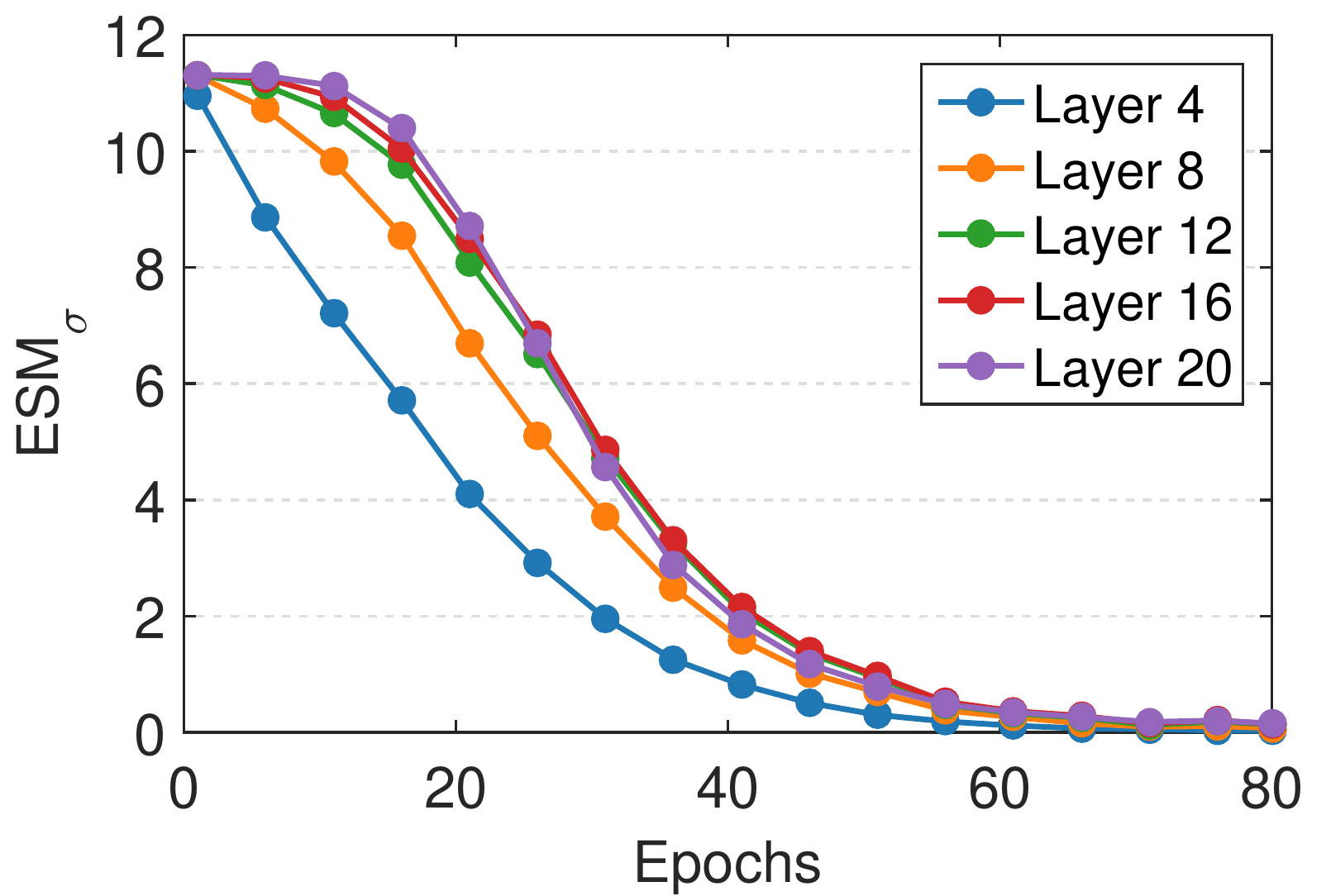}
		\end{minipage}
	}
	%\vspace{-0.08in}
	\caption{Experiments with training set $\mathbf{S}$ equaling to the test set  $\mathbf{S'}$. We follow the same experimental setup as the one in Figure~\ref{fig:exp1_valval} of the paper, except that we use an update factor of $\alpha=0.5$ for the running average in calculate population statistics of BN. }
	\label{sup_fig:exp1_valval_MM05}
	%	\vspace{-0.17in}
\end{figure*}

In Figure~\ref{fig:exp1_valval} of the paper, we show the results of setup one where the training set $\mathbf{S}$ equals to the test set  $\mathbf{S'}$. The experiments are conducted using a learning rate of 0.1 and an update factor $\alpha=0.9$, trained on the 20-layer multi-layer perceptron (MLP) architecture. 
%We train a 20-layer multi-layer perceptron (MLP) with 128 neurons in each layer for MNIST classification.
% We use full-batch gradient descent to train 80 epochs (iterations) with a learning rate of 0.1. 
% The estimated population statistics of BN are calculated by the commonly used running average with update factor $\alpha=0.9$.

\begin{figure*}[htb]
	\centering
	%	\vspace{-0.08in}
	\hspace{-0.15in}	\subfloat[Training and test errors]{
		\begin{minipage}[c]{.30\linewidth}
			\centering
			\includegraphics[width=5.4cm]{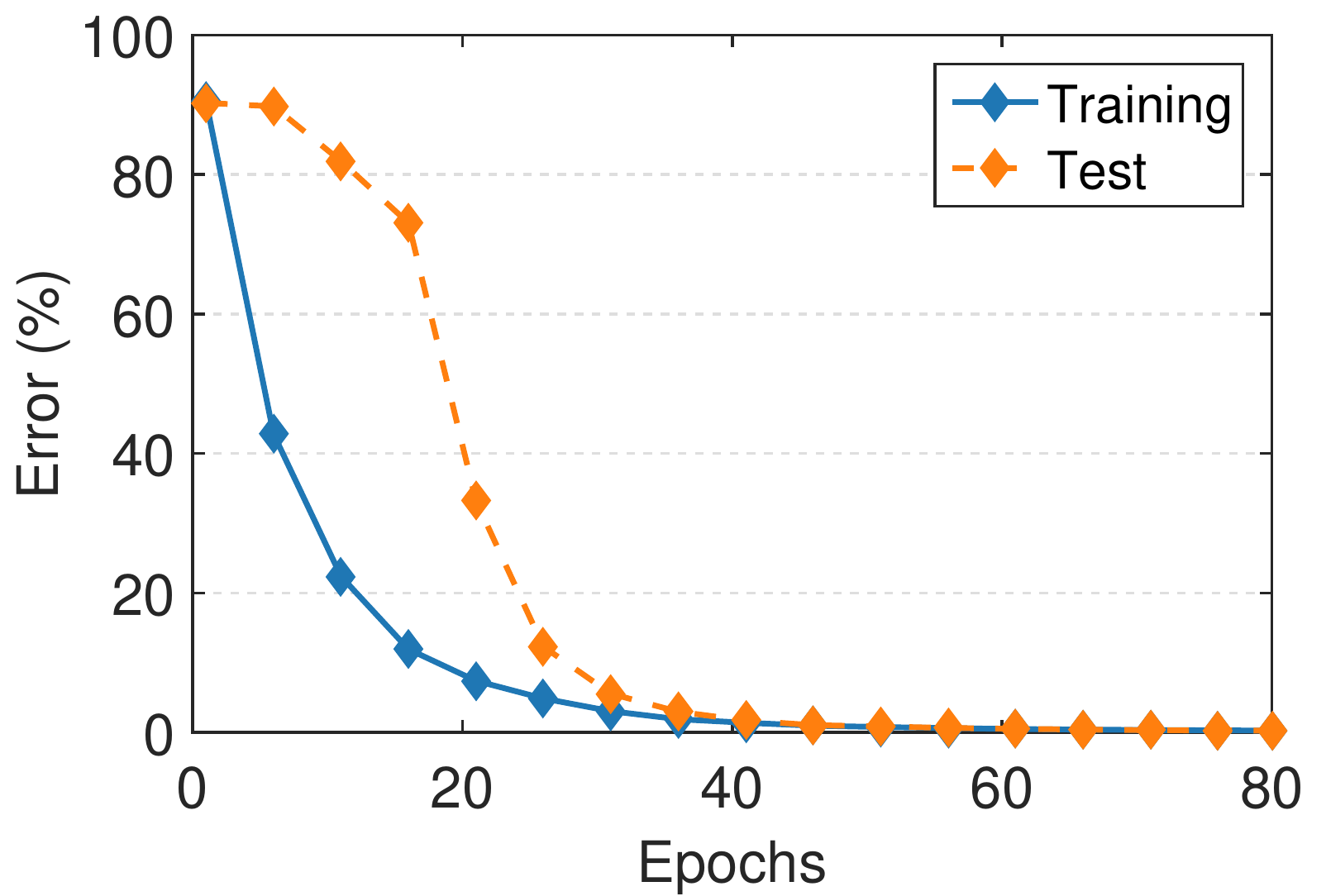}
		\end{minipage}
	}
	\hspace{0.15in}		\subfloat[ESM$_\mu$ of BN in different layers]{
		\begin{minipage}[c]{.30\linewidth}
			\centering
			\includegraphics[width=5.4cm]{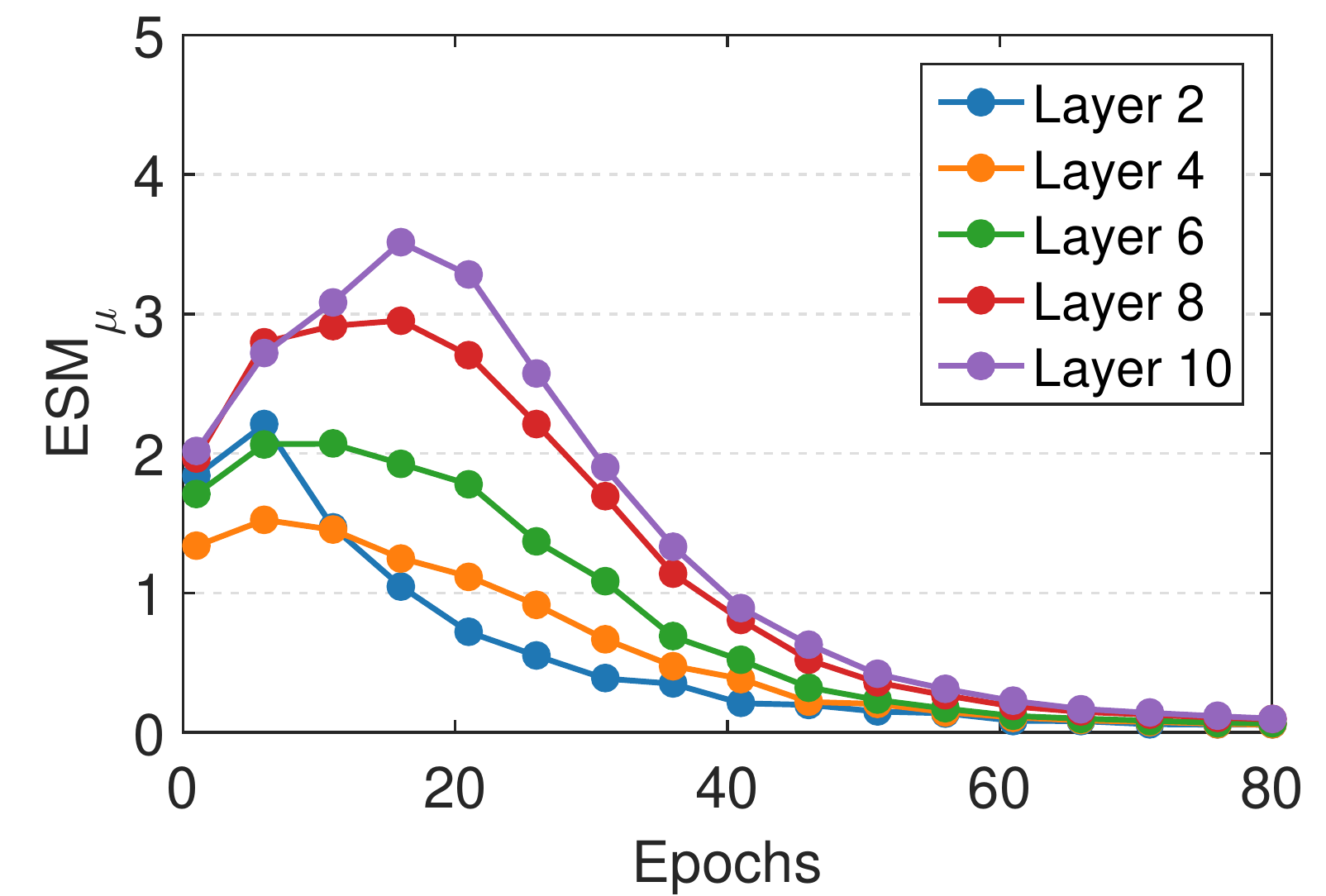}
		\end{minipage}
	}
	\hspace{0.15in}		\subfloat[ESM$_\sigma$ of BN in different layers]{
		\begin{minipage}[c]{.30\linewidth}
			\centering
			\includegraphics[width=5.4cm]{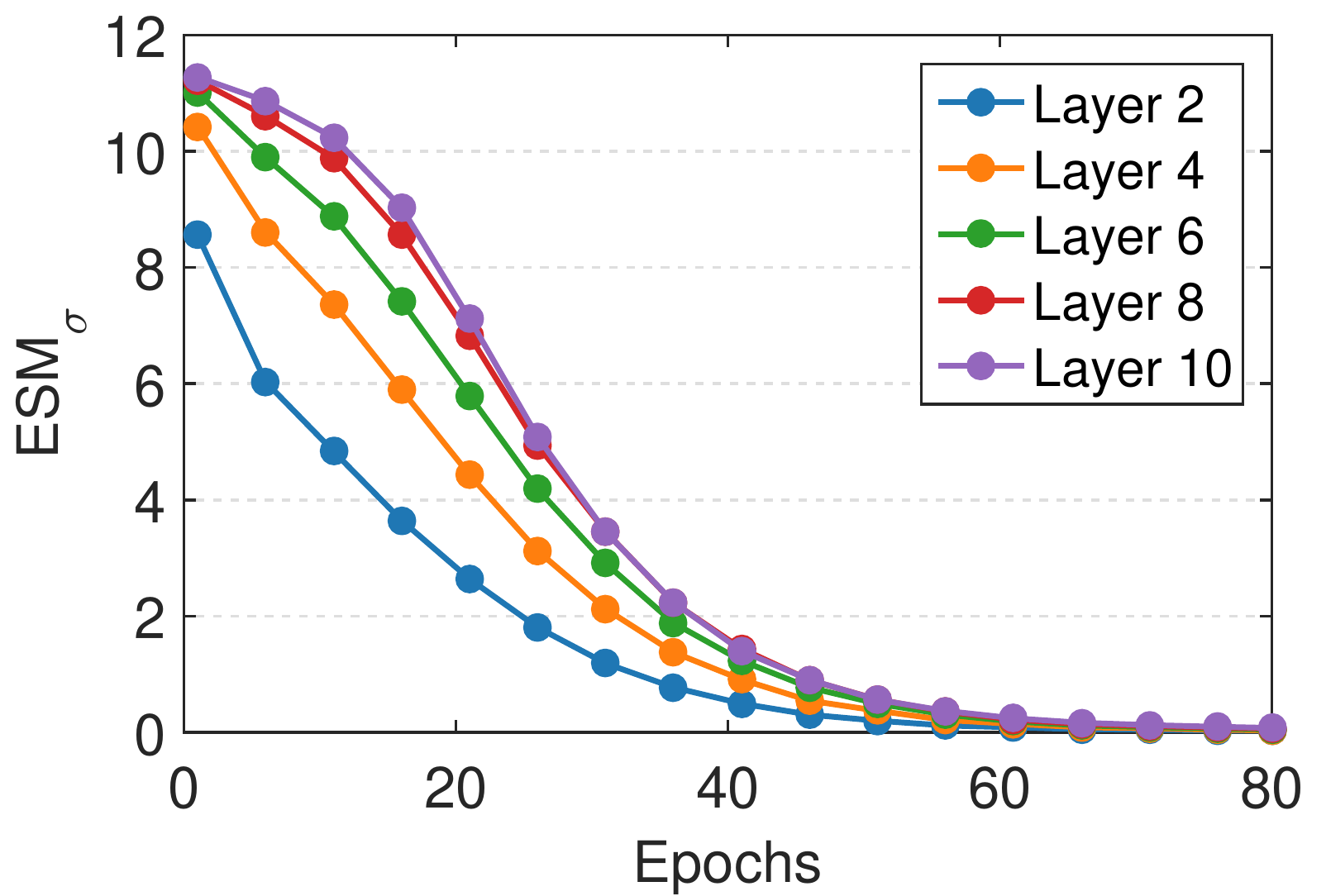}
		\end{minipage}
	}
	%\vspace{-0.08in}
	\caption{Experiments with training set $\mathbf{S}$ equaling to the test set  $\mathbf{S'}$. We follow the same experimental setup as the one in Figure~\ref{fig:exp1_valval} of the paper, except that we train a 10-layer MLP. }
	\label{sup_fig:exp1_valval_d10}
	%	\vspace{-0.17in}
\end{figure*}

\begin{figure*}[htb]
	\centering
	%	\vspace{-0.08in}
	\hspace{-0.15in}	\subfloat[Training and test errors]{
		\begin{minipage}[c]{.30\linewidth}
			\centering
			\includegraphics[width=5.4cm]{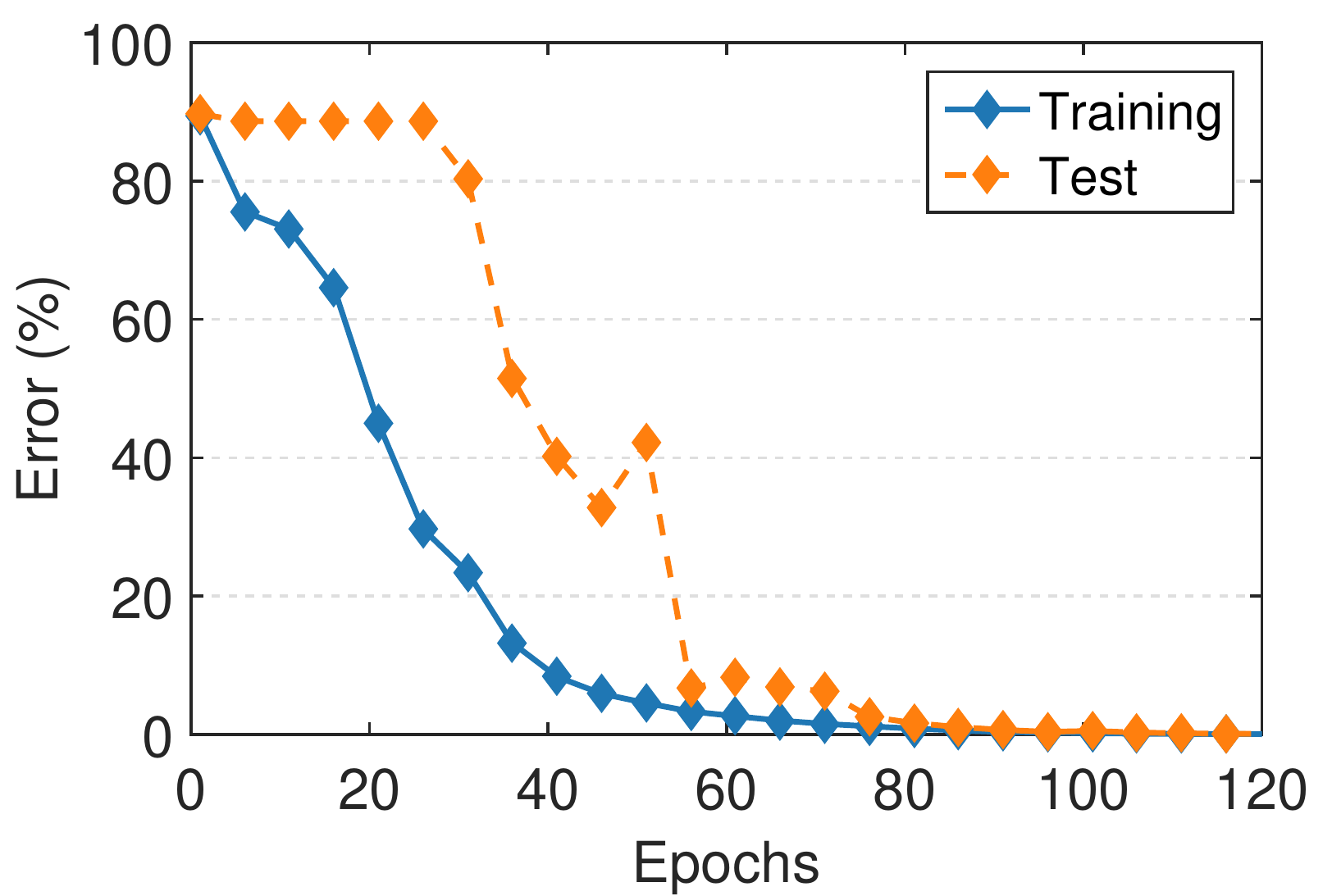}
		\end{minipage}
	}
	\hspace{0.15in}		\subfloat[ESM$_\mu$ of BN in different layers]{
		\begin{minipage}[c]{.30\linewidth}
			\centering
			\includegraphics[width=5.4cm]{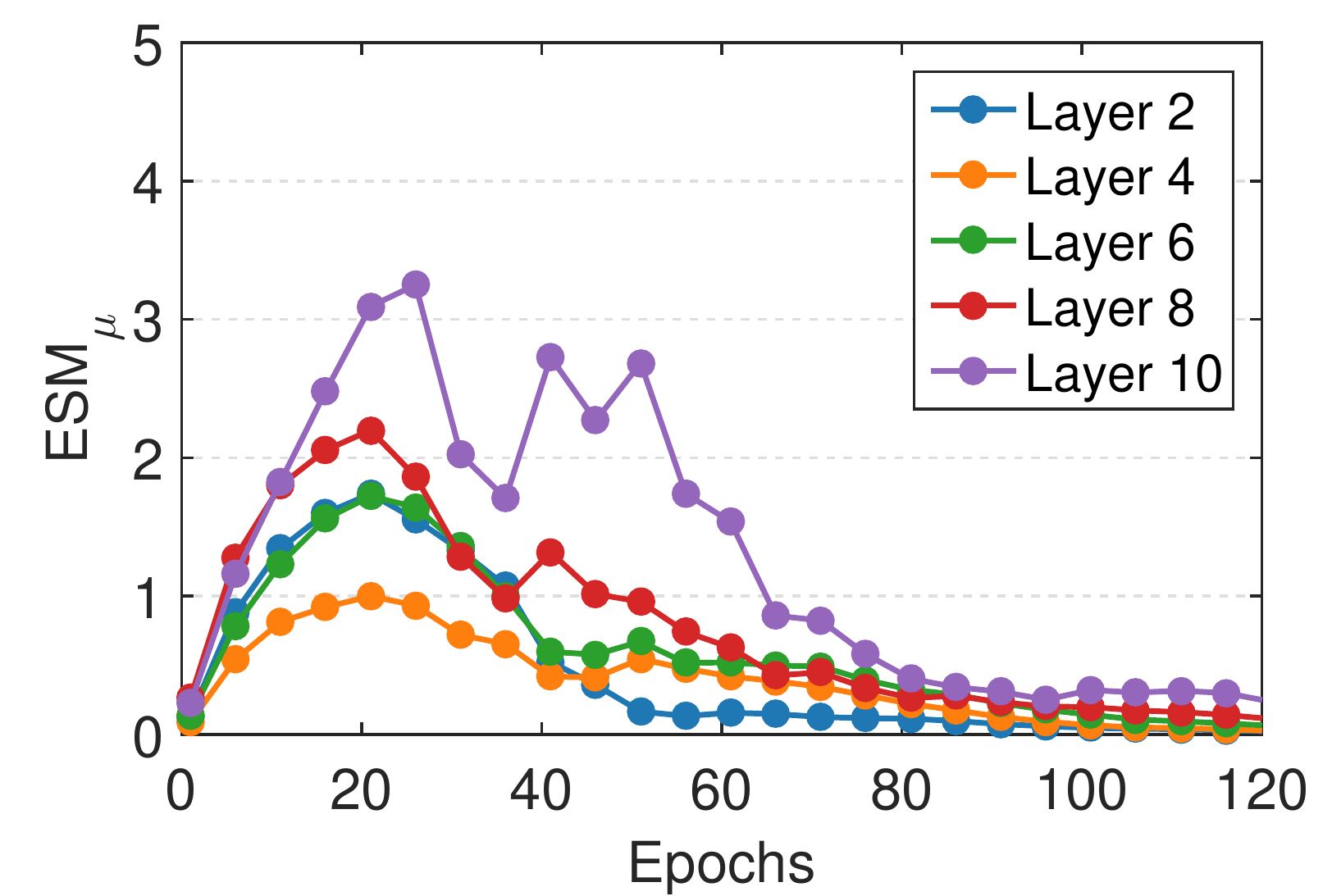}
		\end{minipage}
	}
	\hspace{0.15in}		\subfloat[ESM$_\sigma$ of BN in different layers]{
		\begin{minipage}[c]{.30\linewidth}
			\centering
			\includegraphics[width=5.4cm]{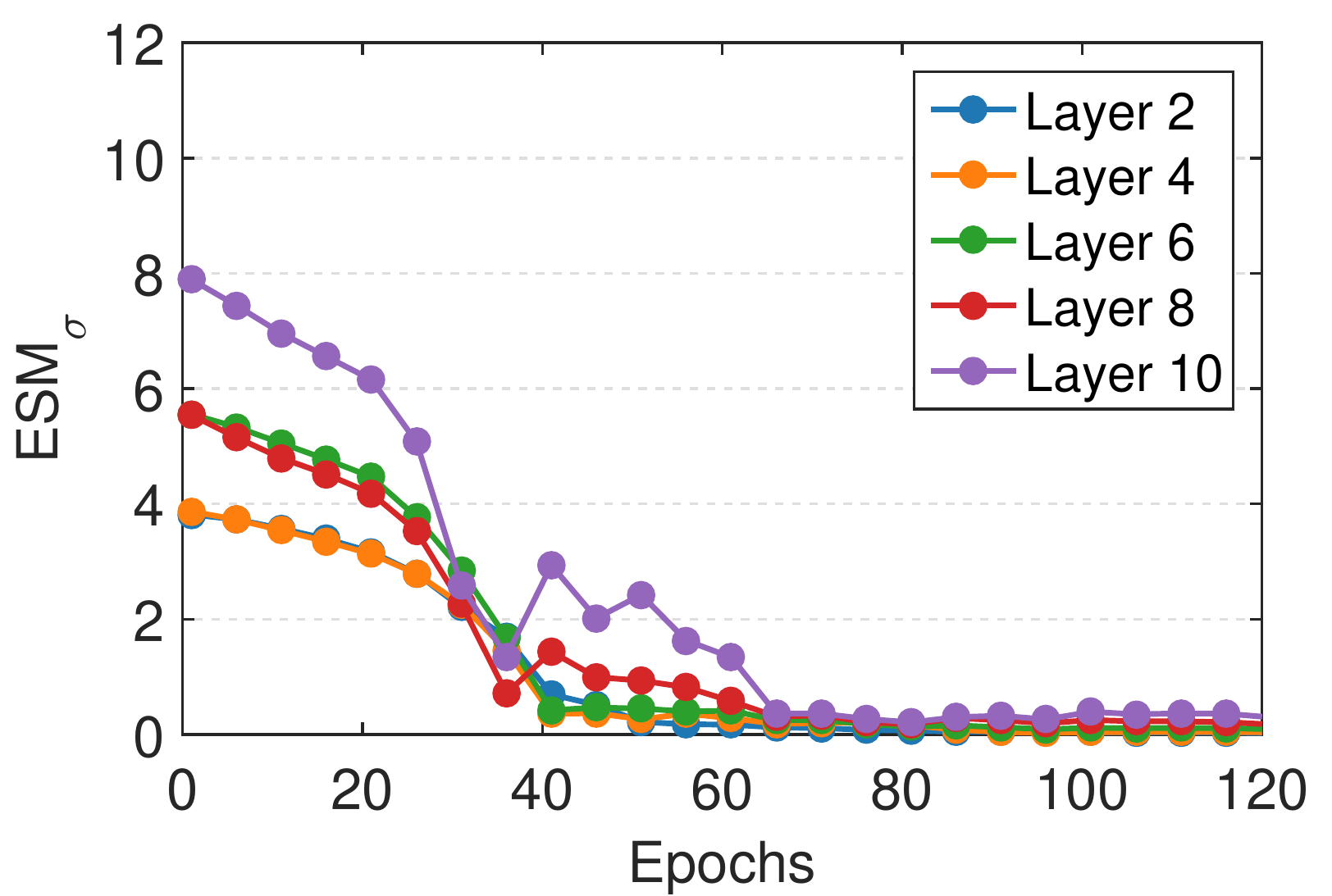}
		\end{minipage}
	}
	%\vspace{-0.08in}
	\caption{Experiments with training set $\mathbf{S}$ equaling to the test set  $\mathbf{S'}$. We train a 14-layer convolutional neural network (CNN) for MNIST classification. $\mathbf{S}$ and $\mathbf{S'}$ are the original test set of MNIST with 10,000 samples. We use full-batch gradient descent to train 120 epochs (iterations) with a learning rate of 0.1. The estimated population statistics of BN are calculated by the commonly used running 
		average with update factor $\alpha=0.9$.}
	\label{sup_fig:exp1_valval_CNN}
	%	\vspace{-0.17in}
\end{figure*}

Here, we provide the results under different configurations, including varying the learning rate (Figure~\ref{sup_fig:exp1_valval_LR05}), varying the update factor $\alpha$  (Figure~\ref{sup_fig:exp1_valval_MM01} and~\ref{sup_fig:exp1_valval_MM05}), varying the depth of the network (Figure~\ref{sup_fig:exp1_valval_d10}), and further experiments on convolutional neural networks (Figure~\ref{sup_fig:exp1_valval_CNN}).
We have the similar observations as  the ones shown in Figure~\ref{fig:exp1_valval} of the paper: 1) there are significant gaps between the training and test errors in the first dozens of  epochs, and these error gaps between training and test are mainly caused by the inaccurate estimation of the population statistics of batch normalization (BN)~\cite{2015_ICML_Ioffe}; 2) the $ESM_{\mu}$ and $ESM_{\sigma}$ of BN in deeper layers  have potentially  higher values during the first dozens of  epochs.

\begin{figure}[t]
	\centering
	%\vspace{-0.08in}
	\hspace{0.15in}	\subfloat[Training and test errors]{
		\begin{minipage}[c]{.48\linewidth}
			\centering
			\includegraphics[width=4.2cm]{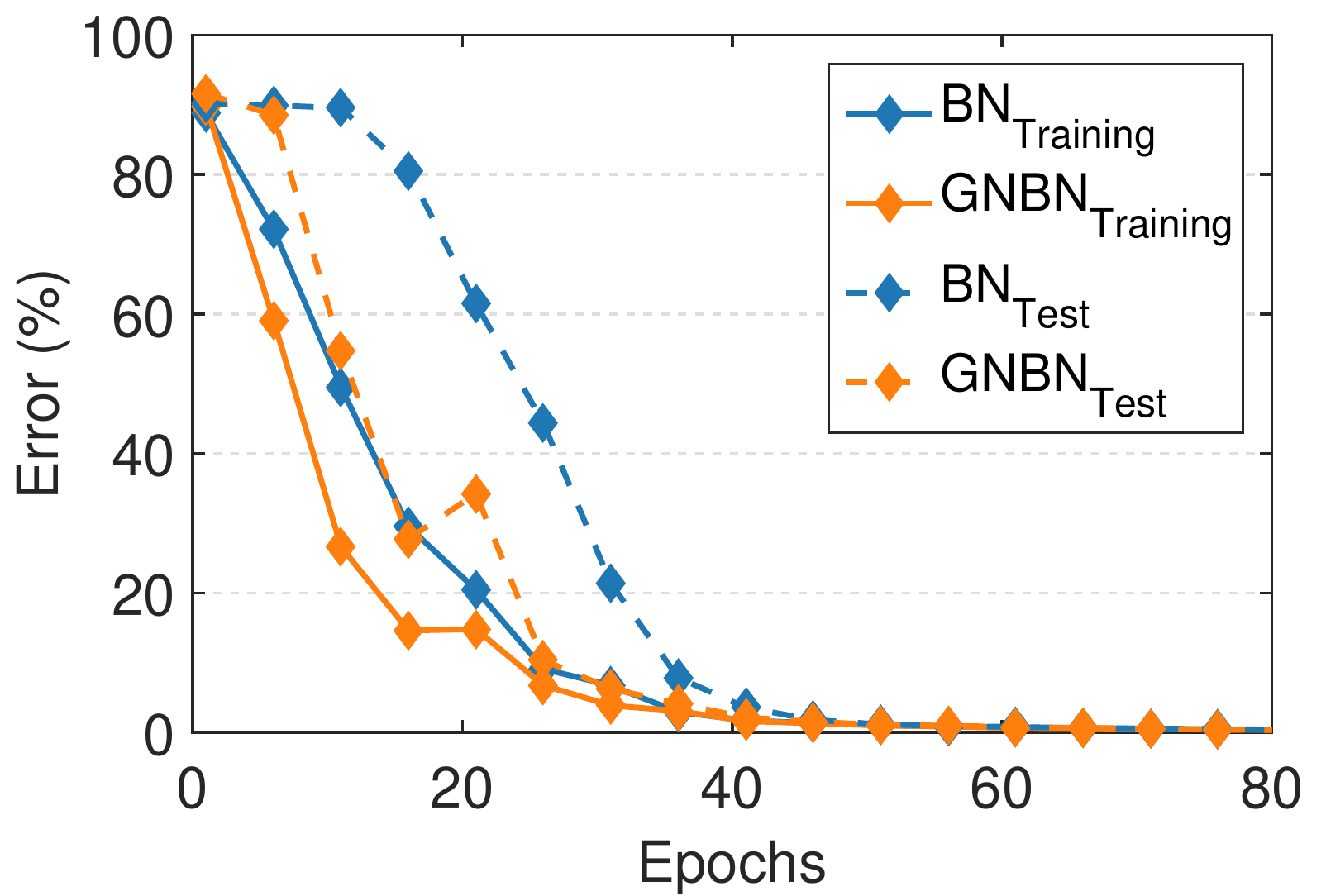}
		\end{minipage}
	}
	\subfloat[ESM$_\sigma$ of BN in different layers]{
		\begin{minipage}[c]{.48\linewidth}
			\centering
			\includegraphics[width=4.2cm]{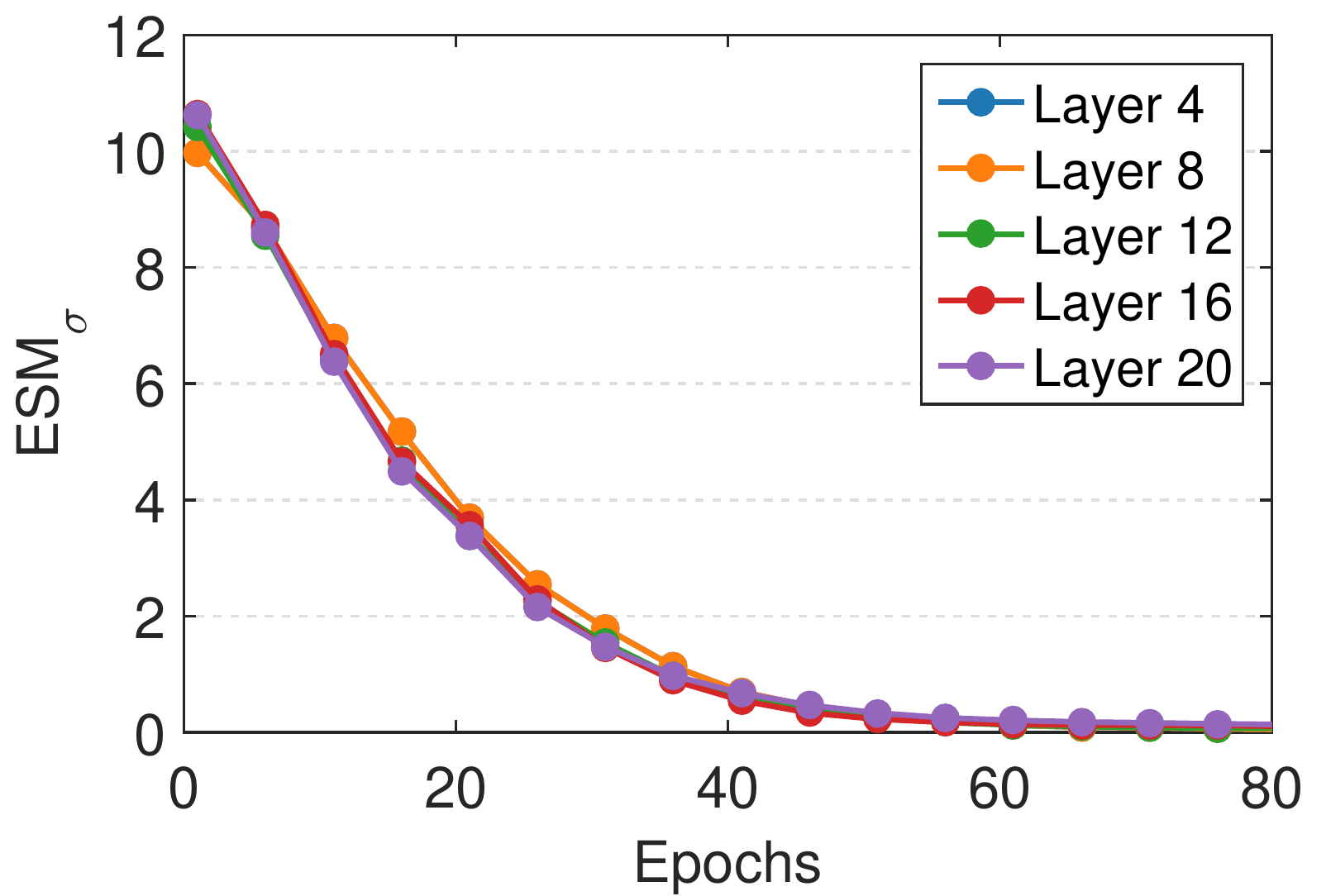}
		\end{minipage}
	}
	%\vspace{-0.08in}
	\caption{Estimation shift experiments on a network with BN and GN mixed (referred to as `GNBN'). We follow the same experimental setup as the one in Figure~\ref{fig:exp1_valval_GNvsBN} of the paper, except that we use GN with a group number of 1. }
	\label{sup_fig:exp1_valval_GNvsBN_NG1}
	%	\vspace{-0.17in}
\end{figure}

\begin{figure}[t]
	\centering
	%\vspace{-0.08in}
	\hspace{0.15in}	\subfloat[Training and test errors]{
		\begin{minipage}[c]{.48\linewidth}
			\centering
			\includegraphics[width=4.2cm]{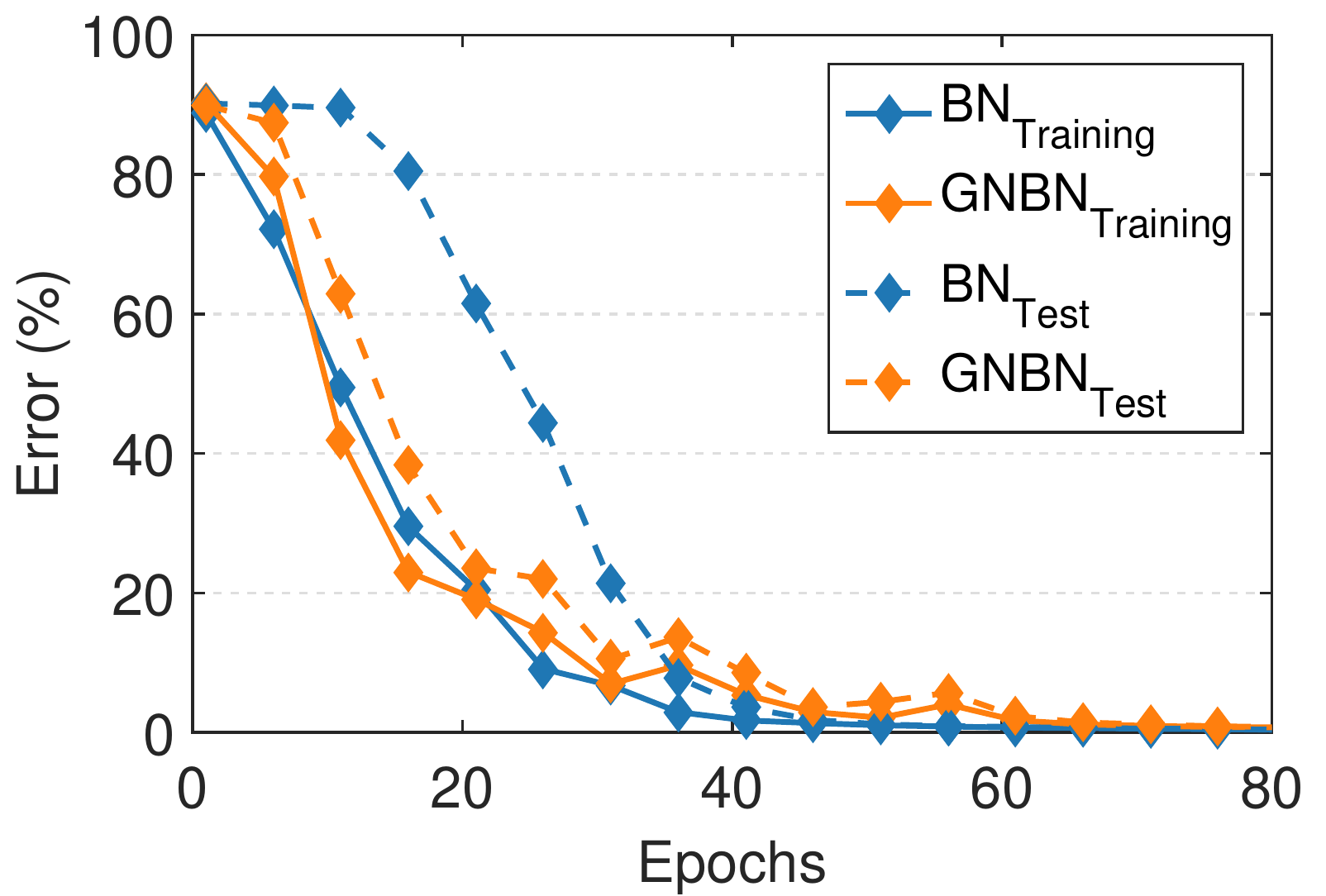}
		\end{minipage}
	}
	\subfloat[ESM$_\sigma$ of BN in different layers]{
		\begin{minipage}[c]{.48\linewidth}
			\centering
			\includegraphics[width=4.2cm]{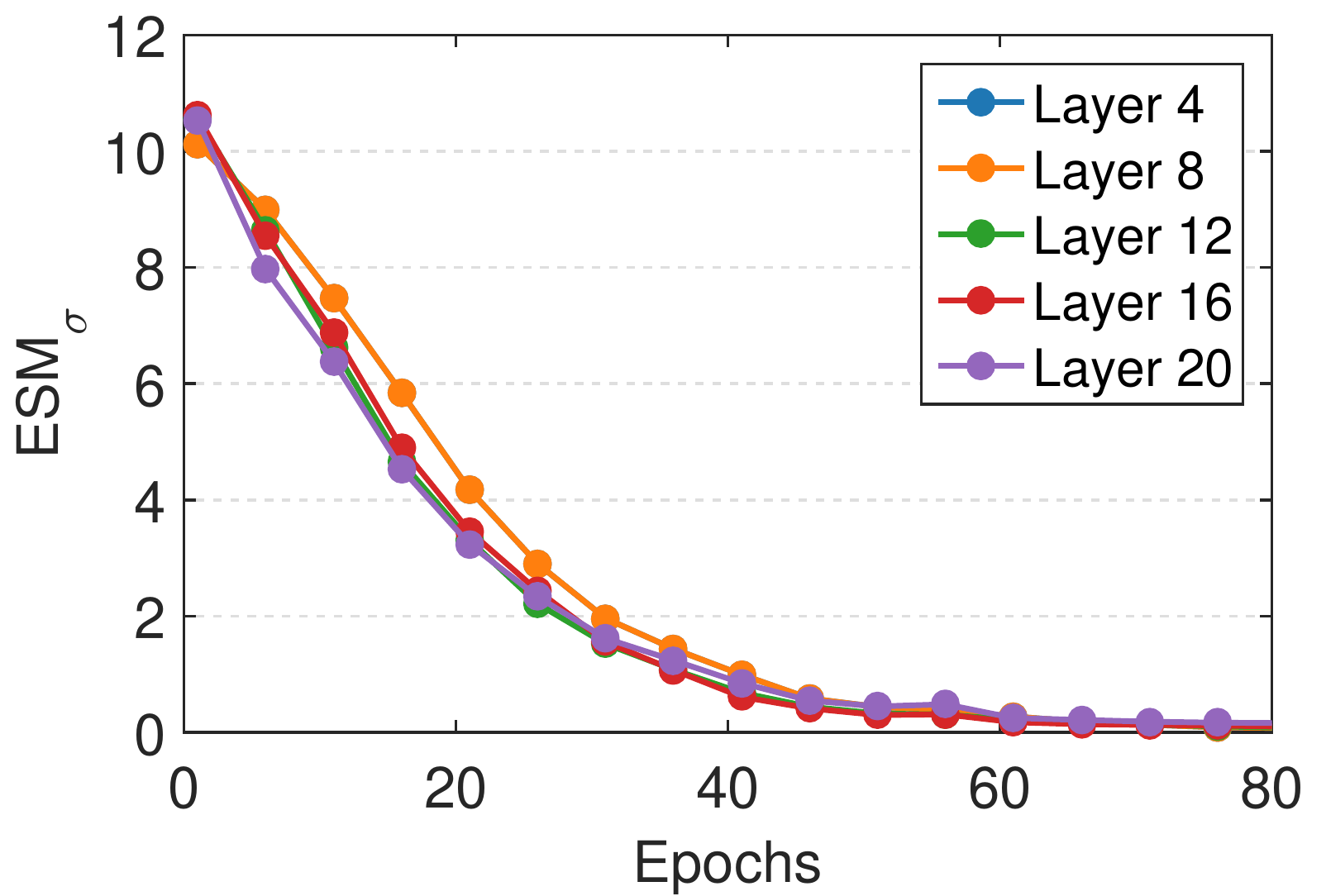}
		\end{minipage}
	}
	%\vspace{-0.08in}
	\caption{Estimation shift experiments on a network with BN and GN mixed (referred to as `GNBN').  We follow the same experimental setup as the one in Figure~\ref{fig:exp1_valval_GNvsBN}  of the paper, except that we use GN with a group number of 16. }
	\label{sup_fig:exp1_valval_GNvsBN_NG16}
	%	\vspace{-0.17in}
\end{figure}

\paragraph{Group normalization with different groups.}
In Figure~\ref{fig:exp1_valval_GNvsBN} of the paper, we show the results using group normalization (GN)~\cite{2018_ECCV_Wu} with group number $g=4$. Here, we provide results of GN with group numbers $g=1$ (Figure~\ref{sup_fig:exp1_valval_GNvsBN_NG1}) and $g=16$ (Figure~\ref{sup_fig:exp1_valval_GNvsBN_NG16}). Note that GN with $g=1$ is equivalent to layer normalization (LN)~\cite{2016_CoRR_Ba}. We have the similar observations  as  the ones shown in Figure~\ref{fig:exp1_valval_GNvsBN} of the paper:  1)  the gaps of training and test errors of `GNBN' are significantly reduced in the first 30 epochs; 2) the $ESM_{\sigma}$ of BNs in each layer of `GNBN' nearly are the same during training.

%\vspace{-0.05in}
\section{More Results on ImageNet Classification}
\vspace{-0.05in}
\label{sup_sec:experiments}
In this section, we provide more results on large-scale ImageNet classification~\cite{2015_IJCV_ImageNet}. 

\begin{figure}[t]
	\centering
	\vspace{-0.1in}
	\includegraphics[width=7.5cm]{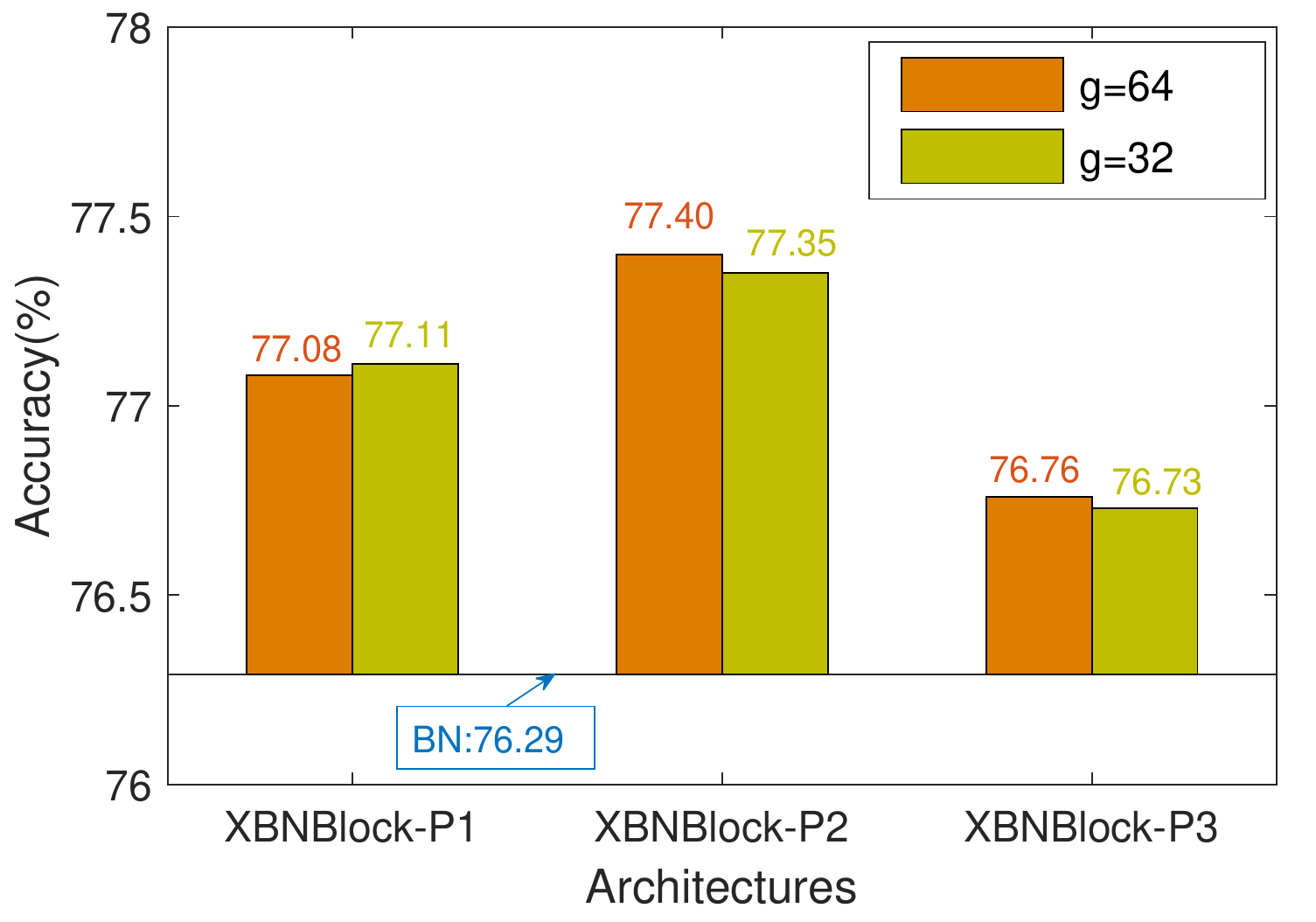}
	\vspace{-0.12in}
	\caption{Comparison of the results (top-1 validation accuracy) between GNs with $g=64$ and  $g=32$, under different positions where a GN is used in an XBNBlock. `XBNBlock-P1', `XBNBlock-P2' and `XBNBlock-P3' indicate that the first, second and third BN in the bottleneck are replaced with GN, respectively.}
	\label{sup_fig:XBNBlock_group}
	%	\vspace{-0.14in}
\end{figure}

\paragraph{Different group number.}
In Section~\ref{sec:exp_ablation} of the paper, we mention that we use GN with group number $g=64$ as BFN in the XBNBlock. Here, we compare the results between  GNs with $g=64$ and  $g=32$, under different positions where a GN is used in an XBNBlock. Figure~\ref{sup_fig:XBNBlock_group} shows the results. We observe that all the models with different group numbers outperform the baseline (BN) significantly. Besides, There are no remarkable differences in performance between the models using GN with $g=64$ and  $g=32$.

\begin{figure*}[t]
	\centering
	%	\vspace{-0.08in}
	\hspace{-0.15in}	\subfloat[]{
		\begin{minipage}[c]{.30\linewidth}
			\centering
			\includegraphics[width=5.6cm]{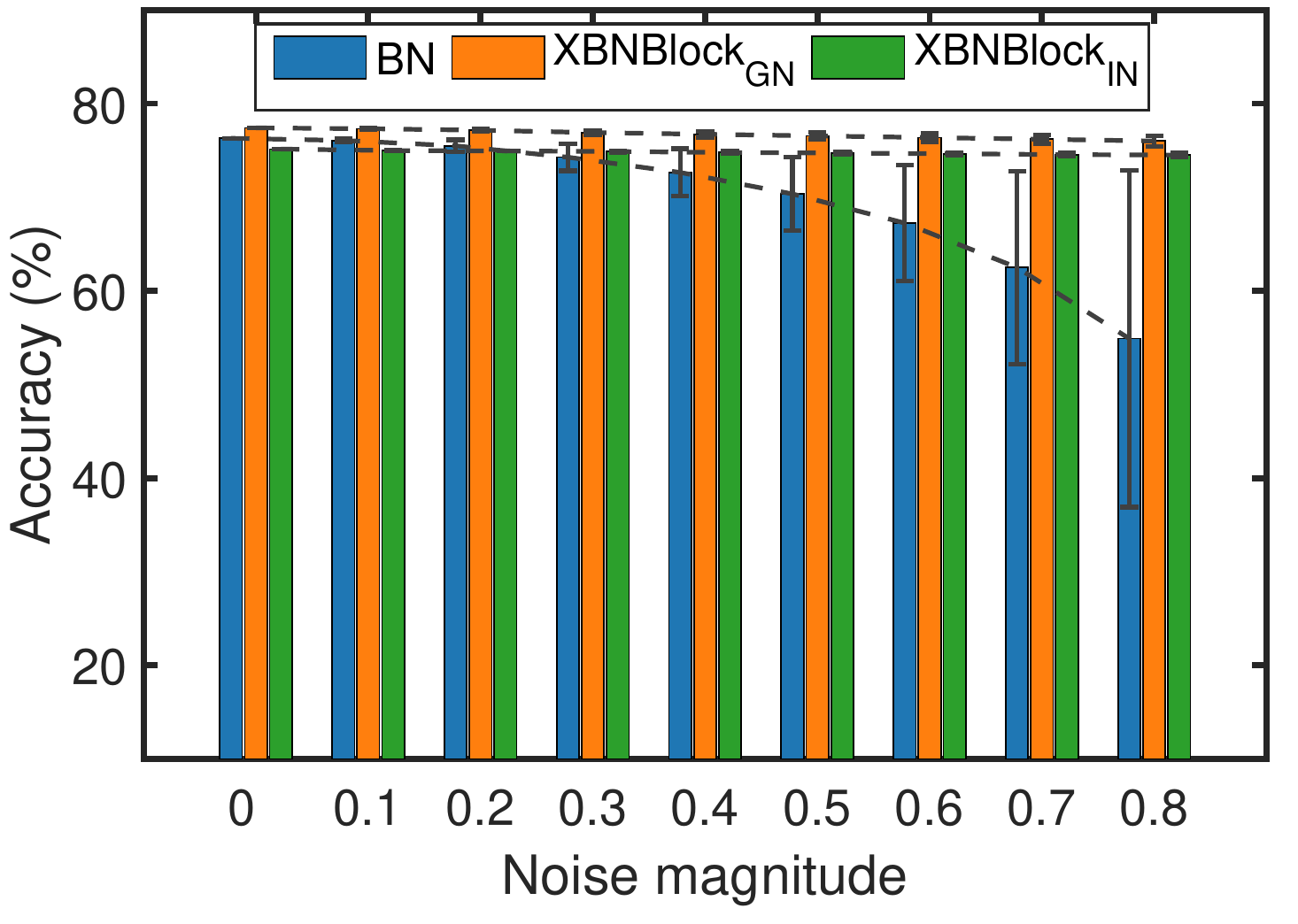}
		\end{minipage}
	}
	\hspace{0.15in}		\subfloat[]{
		\begin{minipage}[c]{.30\linewidth}
			\centering
			\includegraphics[width=5.6cm]{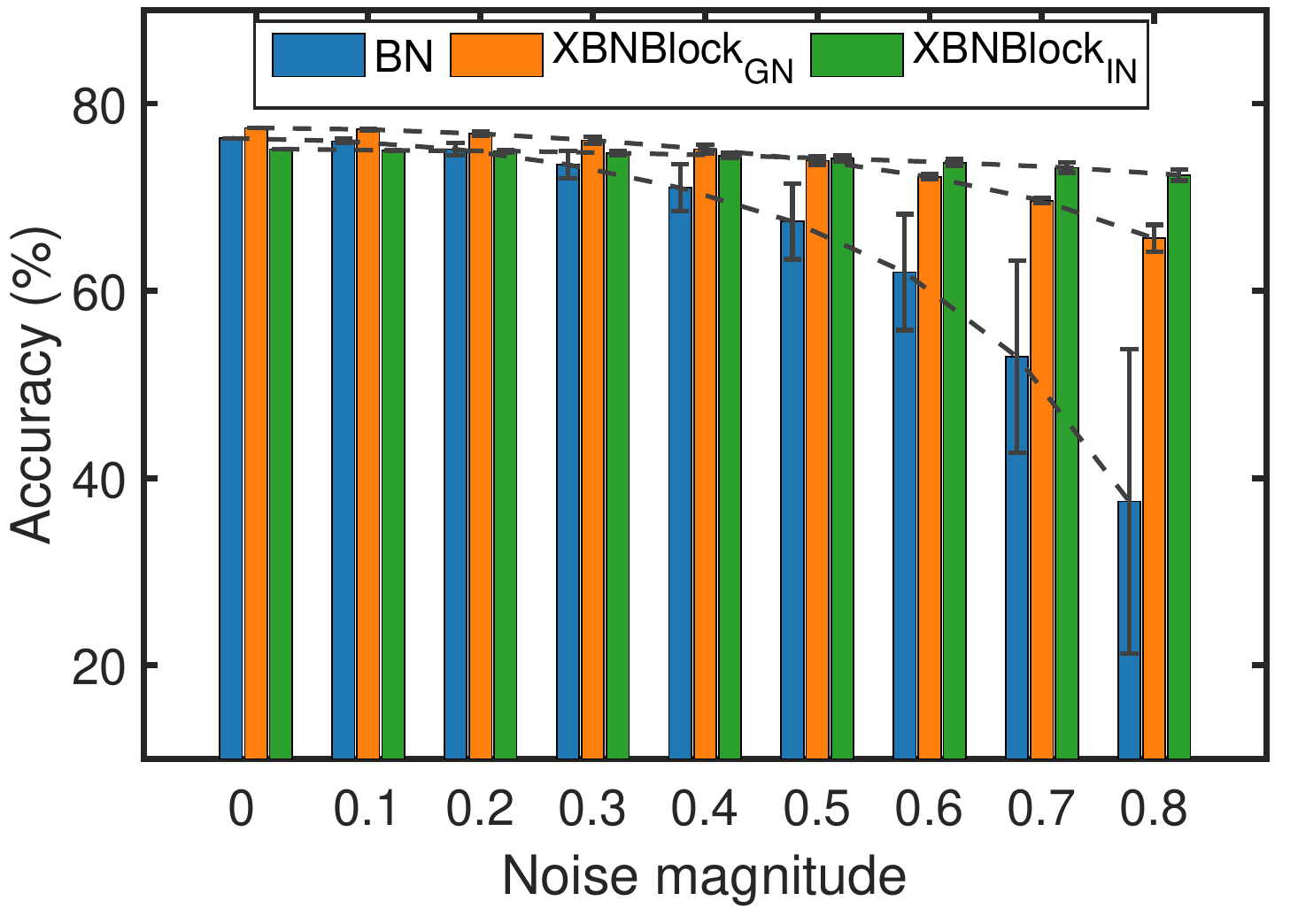}
		\end{minipage}
	}
	\hspace{0.15in}		\subfloat[]{
		\begin{minipage}[c]{.30\linewidth}
			\centering
			\includegraphics[width=5.6cm]{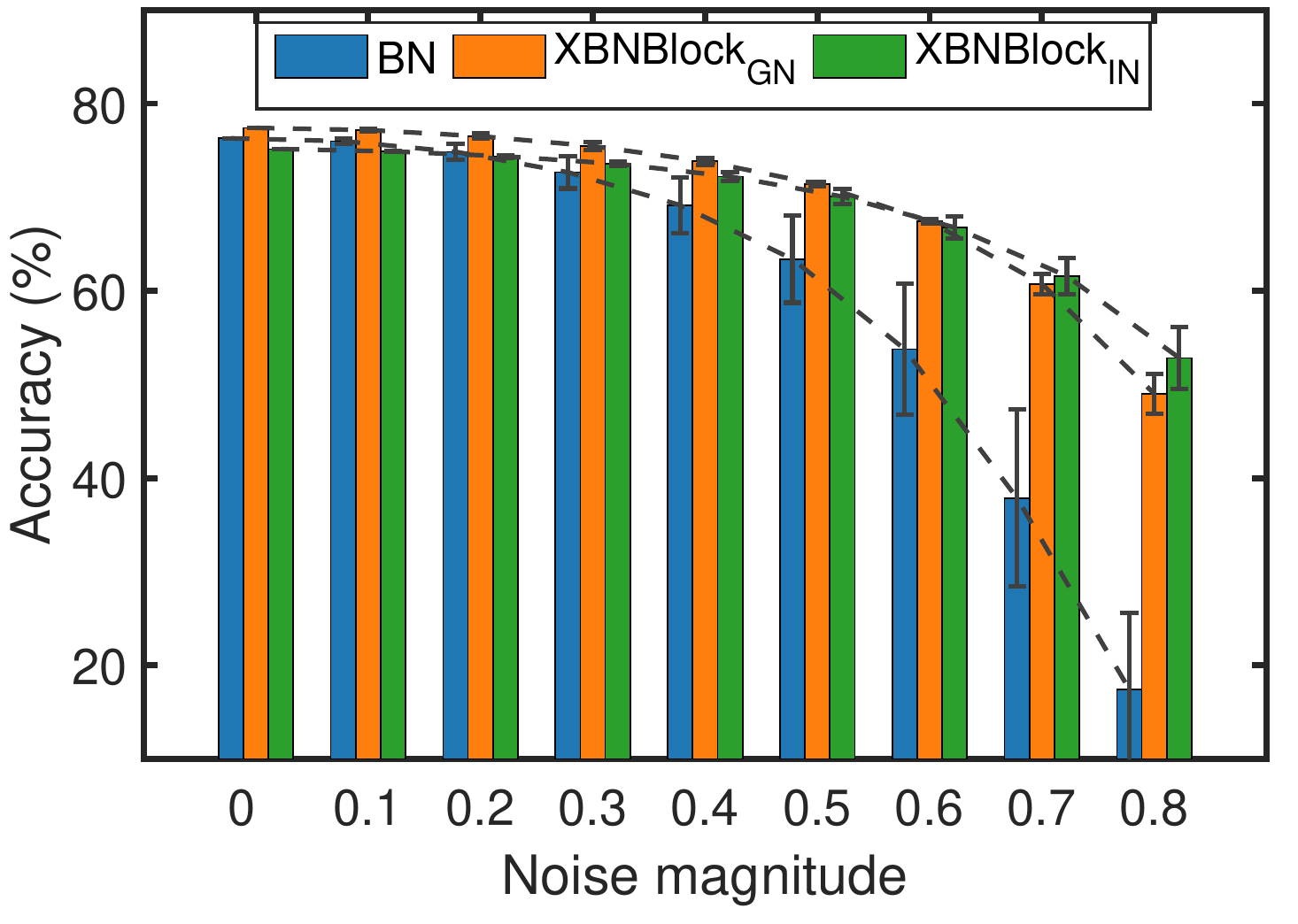}
		\end{minipage}
	}
	%\vspace{-0.08in}
	\caption{Top-1 validation accuracy with different noise magnitude imposed on the estimated population statistics. The results are averaged over 5 random seeds. We refer to a bottleneck/XBNBlock as `disturbed block' if its first BN uses  $\{ \hat{\mu}_{\delta}, \hat{\sigma}^2_{\delta}\}$ for normalization during inference. (a) the first  block of ResNet-50 is  `disturbed block'; (b) the first 6 blocks of ResNet-50 are `disturbed block'; (c) the first 13rd blocks of ResNet-50 are `disturbed block'. }
	\label{sup_fig:exp6_RandomPerturb}
	%	\vspace{-0.17in}
\end{figure*}

\paragraph{Robustness to distribution shift.}
In Figure~\ref{fig:exp6_RandomPerturb} of the paper, we conduct experiments on model robustness to distribution shift. We show the results on ResNet-50 with its first six blocks being `disturbed block'. Here, we provide more results on ResNet-50 with different blocks being `disturbed block', and the results are shown in Figure~\ref{sup_fig:exp6_RandomPerturb}. We obtain similar observations as the ones in Figure~\ref{fig:exp6_RandomPerturb} of the paper. Besides, we observe that BN has significant performance degeneration, even though only one BN's population statistics are disturbed (Figure~\ref{sup_fig:exp6_RandomPerturb} (a)), while XBNBlock$_{GN}$ and XBNBlock$_{IN}$ have no remarkable performance degeneration in this case (Figure~\ref{sup_fig:exp6_RandomPerturb} (a)). 

%
%\vspace{-0.05in}
%\subsubsection{Validation on Larger Models}
%\vspace{-0.05in}
%In this section, we validate the effectiveness of XBNBlock on ResNet-101~\cite{2015_CVPR_He}, ResNeXt-50 and ResNeXt-101~\cite{2017_CVPR_Xie}. 
%Our baselines are the original networks trained with BN, and we also train the models with GN.
%The results are shown in Table~\ref{table:ImageNet-Res-Step}. We can see that our method consistently improves the baseline (BN) by a significant margin over all architectures. Our method obtains comparable performance to IBN-Net~\cite{2018_ECCV_Pan}. Note that IBN-Net finely designs the position of a IN layer and its channel number, while the design of our XBNBlock is simple. We argue  our claims that a BFN (\eg, IN) can block the accumulation of estimation shift also provide a reasonable explanation to the success of IBN-Net in its test  performance, especially in the scenrios with distribution shift (\eg, domain adaptations and transfer learning tasks.)

\paragraph{Advanced training strategies.}
In Section~\ref{sec:larger-model} of the paper,  we conduct experiments using more advanced training strategies and  show the results on ResNet-50~\cite{2015_CVPR_He}. Here, we provide the results on ResNet-101~\cite{2015_CVPR_He} and ResNext-50~\cite{2017_CVPR_Xie} (Table~\ref{sup_table:Advanced-Strategy}).  We also observe that XBNBlock  consistently outperforms the baseline by a remarkable margin. 
\begin{table*}[b]
	\centering
	\vspace{-0.1in}
	%\begin{footnotesize}
	\begin{tabular}{c|cc|cc}
		\bottomrule[1pt]
		& \multicolumn{2}{c| }{ResNet-101} &    \multicolumn{2}{c}{ResNext-50}  \\
		Training strategies & Baseline (BN)    & XBNBlock$_{GN}$ & Baseline (BN)    & XBNBlock$_{GN}$        \\
		\hline
		label smooth (LS)  &78.25 &\textbf{78.85} &77.83 &\textbf{78.46} \\
		MixUp &78.67 &\textbf{79.14}  &78.20 &\textbf{78.73} \\
		cosine learning (COS) &78.51 &\textbf{78.91}  &77.91 &\textbf{78.31} \\
		LS~+~MixUP~+~COS &79.10 &\textbf{79.41}  &78.84 &\textbf{79.27} \\
		\toprule[1pt]
	\end{tabular}
	\caption{Top-1 accuracy ($\%$) on ResNet-101 and ResNeXt-50 using advanced training strategies.}
	%	\vspace{-0.05in}
	\label{sup_table:Advanced-Strategy}
\end{table*}

\paragraph{Towards whitening.} 
We also apply the recently proposed group whitening (GW)~\cite{2021_CVPR_Huang} as a BFN in our XBNBlock, referred to as XBNBlock$_{GW}$. We use the released code provided in ~\cite{2021_CVPR_Huang}. The results are shown in Table~\ref{sup_table:whitening}. By applying GW in our design, our XBNBlock outperforms the state-of-the-art whitening methods. \Eg, our method `XBNBlock$_{GW}$-D4' obtains $79.18\%$ top-1 accuracy on ResNet-101. Note that XBNBlock$_{GW}$-D4  has only  $7\%$ additional time cost. 
% to form a more beneficial results. By using group whitening into our design, our method can obtain comparable performance to the state-of-art hwitening methods,  

\begin{table}[b]
	\centering
	%	\vspace{-0.1in}
	%	\begin{footnotesize}
		\begin{tabular}{c|cc}
			\bottomrule[1pt]
			Method     & ResNet-50   & ResNet-101    \\
			\hline
			Baseline (BN)~\cite{2015_ICML_Ioffe}  &76.29 &77.65  \\
			%	GN~\cite{2018_ECCV_Wu}  &75.73 &77.18  \\
			BW~\cite{2020_CVPR_Huang}  &77.21 &78.27  \\
			SW~\cite{2019_ICCV_Pan} &\textbf{77.93}&79.13  \\
			GW~\cite{2021_CVPR_Huang}/XBNBlock$_{GW}$ &77.72 &78.71  \\
			XBNBlock$_{GW}$-D2 (ours) &77.89&79.12  \\
			XBNBlock$_{GW}$-D4 (ours) &77.56&\textbf{79.18}  \\
			\toprule[1pt]
		\end{tabular}
		\vspace{-0.13in}
		\caption{Comparison of top-1 validation accuracy (\%) between methods with whitening module. We compare our method to batch whitening (BW)~\cite{2020_CVPR_Huang}, switchable whitening (SW) ~\cite{2019_ICCV_Pan} and group whitening (GW) ~\cite{2021_CVPR_Huang}. Note that the ResNet-50 (ResNet-101) with GW used in paper ~\cite{2021_CVPR_Huang} is equivalent to the ResNet-50 (ResNet-101) with our XBNBlock$_{GW}$ used.}
		%	\vspace{-0.17in}
		\label{sup_table:whitening}
		%	\end{footnotesize}
\end{table}

\begin{table}[t]
	\centering
	%	\vspace{-0.1in}
	%	\begin{footnotesize}
		\begin{tabular}{c|cc}
			\bottomrule[1pt]
			Method     & Standard training   &  Advanced training   \\
			\hline
			Baseline (BN)  &66.59 &70.83  \\
			XBNBlock$_{GN}$ &\textbf{70.81}&\textbf{72.69}  \\
			\toprule[1pt]
		\end{tabular}
		\vspace{-0.13in}
		\caption{Top-1 accuracy (\%)  on MobileNet-V2  for ImageNet classification.}
		%\vspace{-0.17in}
		\label{sup_table:MobileNet}
		%	\end{footnotesize}
\end{table}

\subsection{Experiments on other Architectures}
\label{supsec:other-arch}
\begin{figure}[]
	\centering
	\vspace{-0.1in}
	\hspace{-0.15in}	\subfloat[Inverted residual block]{
		\begin{minipage}[c]{.42\linewidth}
			\centering
			\includegraphics[width=3cm]{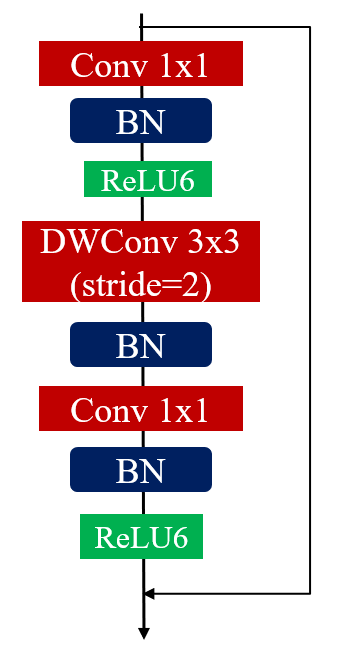}
		\end{minipage}
	}
	\hspace{0.15in}		\subfloat[XBNBlock-P2]{
		\begin{minipage}[c]{.42\linewidth}
			\centering
			\includegraphics[width=3cm]{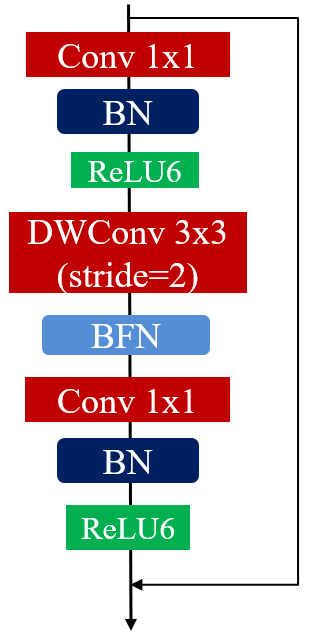}
		\end{minipage}
	}
	\vspace{-0.1in}
	\caption{Inverted residual block of MobileNet-V2 \vs~our XBNBlock-P2 that replaces its second BN with a  BFN. Note that `DWconv' indicates the depth-wise convolutions.}
	\label{sup_fig:MoblieNet}
	%	\vspace{-0.1in}
\end{figure}

\begin{figure}[]
	\centering
	\vspace{-0.1in}
	\hspace{-0.15in}	\subfloat[Basic block of ShuffleNet-V2]{
		\begin{minipage}[c]{.42\linewidth}
			\centering
			\includegraphics[width=3cm]{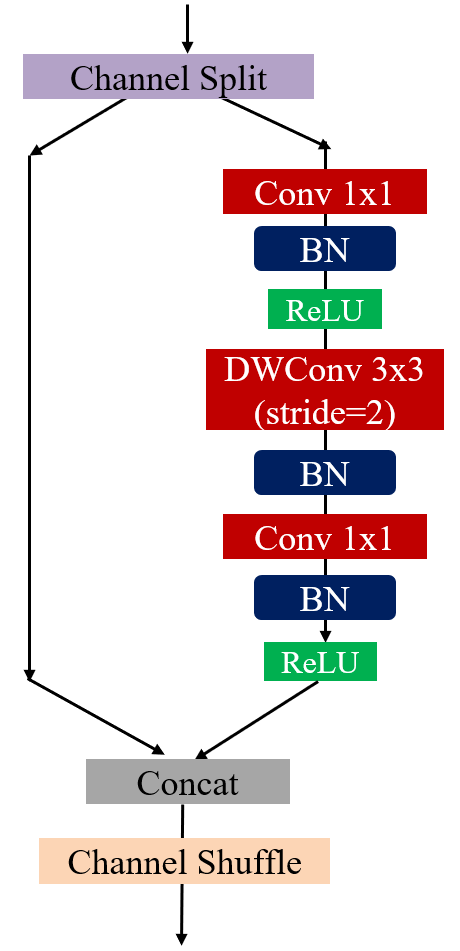}
		\end{minipage}
	}
	\hspace{0.15in}		\subfloat[XBNBlock-P2]{
		\begin{minipage}[c]{.42\linewidth}
			\centering
			\includegraphics[width=3cm]{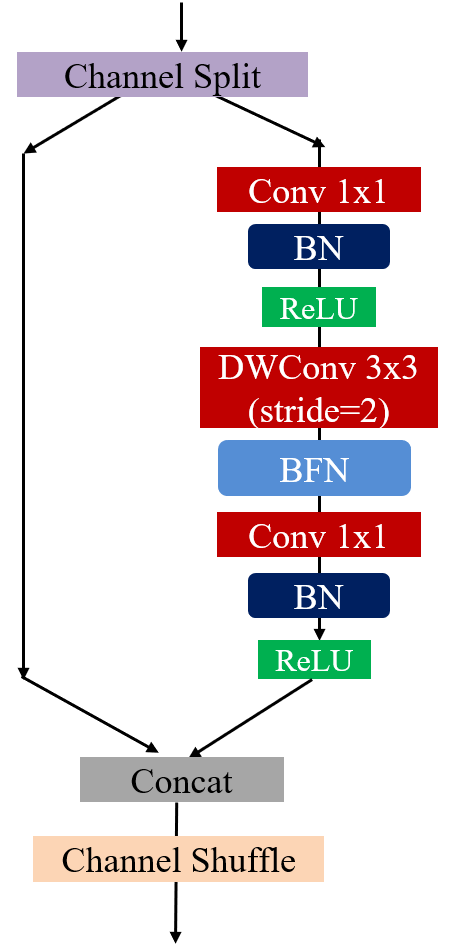}
		\end{minipage}
	}
	\vspace{-0.1in}
	\caption{Basic block of ShuffleNet-V2 \vs~our XBNBlock-P2 that replaces its second BN with a  BFN.}
	\label{sup_fig:ShuffleNet}
	%	\vspace{-0.1in}
\end{figure}

In Section~\ref{subsec_imagenet} of the paper, we show the results experimented on ResNet and ResNeXt architectures. 
Here, we  provide the results on MobileNet-V2~\cite{2018_CVPR_Sandler} and ShuffleNet-V2~\cite{2018_ECCV_Ma} architectures which are designed for more efficient computations.
We conduct experiments using  XBNBlock-P2 that replace the second BN of the original block of MobileNet-V2 (Figure~\ref{sup_fig:MoblieNet}) and ShuffleNet-V2 (Figure~\ref{sup_fig:ShuffleNet}) with a BFN. We again use GN as BFN. 

% Besides, we provide preliminary experiments on vision transformer\TODO{cite}. 

\subsubsection{MobileNet-V2}
Following the experimental setup shown in the paper, we consider two training protocols:

(1) \textbf{Standard training protocol}: We apply stochastic gradient descent (SGD) using a mini-batch size of 256, momentum of 0.9 and weight decay of 0.0001. We train over 100 epochs. The initial learning rate is set to 0.1 and divided by 10 at 30, 60  and 90 epochs. 

(2) \textbf{Advanced training protocol}: We train 150 epochs with cosine learning rate decay, and use a weight decay of 0.00004, under which the baseline model (MobileNet) obtains a better performance. 

The results are shown in Table~\ref{sup_table:MobileNet}.  We observe that the proposed  XBNBlock consistently improves the performance of the original MobileNet-V2 architecture.
%Note the results we run on Moblinest do not match the peformance of Mobilenet the papers, since they use a more sophised ate trianing protocles. Here, we only show that our method can imporve the 

\begin{figure*}[h]
	\centering
	\vspace{-0.1in}
	\hspace{-0.15in}	\subfloat[Standard training protocol]{
		\begin{minipage}[c]{.42\linewidth}
			\centering
			\includegraphics[width=6.2cm]{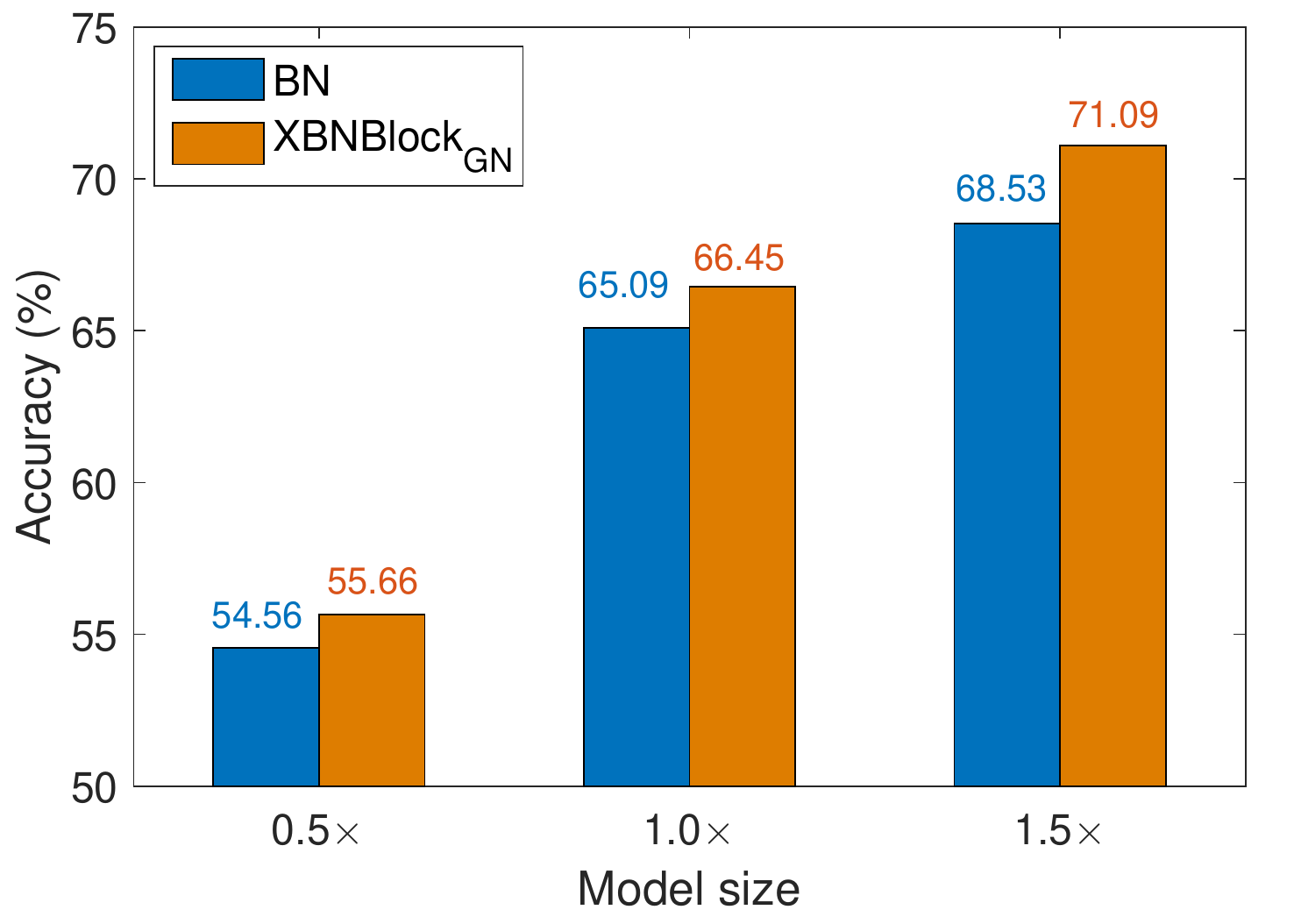}
		\end{minipage}
	}
	\hspace{0.15in}		\subfloat[Advanced training protocol]{
		\begin{minipage}[c]{.42\linewidth}
			\centering
			\includegraphics[width=6.2cm]{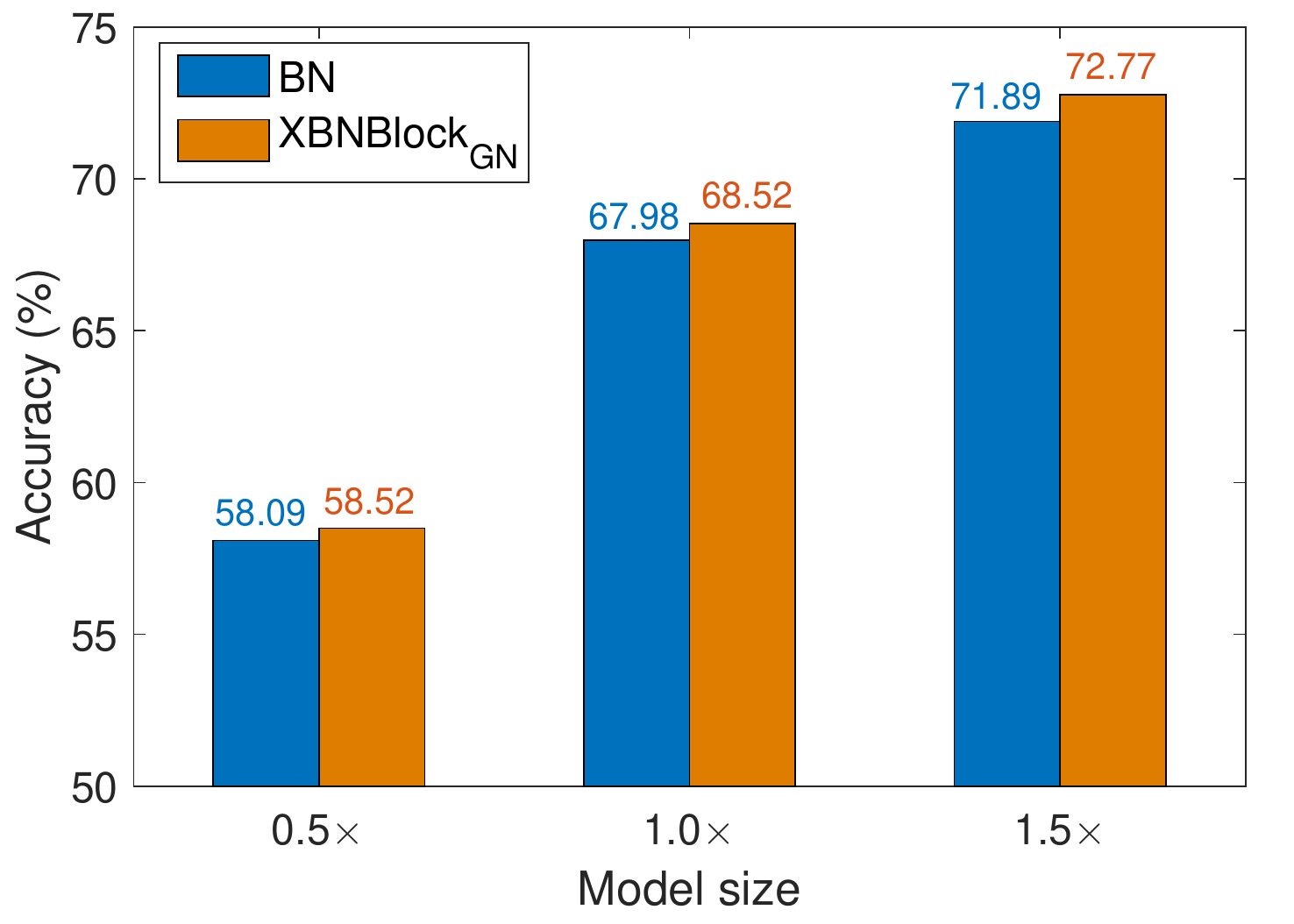}
		\end{minipage}
	}
	\vspace{-0.1in}
	\caption{Top-1 accuracy (\%)  on ShuffleNet-V2  for ImageNet classification.}
	\label{sup_fig:Result_shuffleNet}
	\vspace{-0.1in}
\end{figure*}

\subsubsection{ShuffleNet-V2}
%\TODO{describe ShuffleNet}

Here, we also consider two training protocols:

(1) \textbf{Standard training protocol}: We apply SGD using a mini-batch size of 256, momentum of 0.9 and weight decay of 0.0001. We train over 100 epochs. The initial learning rate is set to 0.1 and divided by 10 at 30, 60  and 90 epochs. 

(2) \textbf{Advanced training protocol}: We train 150 epochs with cosine learning rate decay, and use a weight decay of 0.00004. We also further use the label smoothing with a  smoothing factor of 0.1. The baseline (ShuffleNet-V2)  has a better performance under this training protocol. 

Table~\ref{sup_table:MobileNet} show the results of ShuffleNet-V2 with different model sizes, including the  `$0.5\times$', `$1.0\times$'  and `$1.5\times$'~\cite{2018_ECCV_Ma}.  We observe that the proposed  XBNBlock consistently improve the performance of the original ShuffleNet-V2 architecture, under different training protocols and model sizes.

\section{More Results of Detection and Segmentation on COCO}
In Section~\ref{sec_Coco} the paper, we only report the average precision (AP) for bounding box detection (AP$^{bbox}$) and instance segmentation (AP$^{mask}$)~\cite{2014_ECCV_COCO}, due to space limit. Here, we report more COCO metrics including the results at different scales (AP$_{50}$, AP$_{75}$, AP$_{s}$, AP$_{m}$ and AP$_{l}$) for both bounding box detection and instance segmentation. Table~\ref{sup-table:Mask-R50} and Table~\ref{sup-table:Mask-X101} show the results using ResNet-50 and ResNeXt-101 respectively as the backbone.

\begin{table*}[t]
	\centering
	\begin{scriptsize}
		\begin{tabular}{c|llllll|llllll}
			\bottomrule[1pt]
			\multicolumn{12}{c}{2fc head box}  \\
			\hline
			Method     & AP$^{bbox}$   & AP$^{bbox}_{50}$ & AP$^{bbox}_{75}$  & AP$^{bbox}_{s}$ & AP$^{bbox}_{m}$ & AP$^{bbox}_{l}$ & AP$^{mask}$ & AP$^{mask}_{50}$ & AP$^{mask}_{75}$  & AP$^{mask}_{s}$ & AP$^{mask}_{m}$ & AP$^{mask}_{l}$\\
			\hline
			$BN^{\dag}$  &  37.40 & 59.01 & 40.43 & 21.87 & 40.91 & 48.17  &  34.01 & 55.72 & 35.9 & 15.56 & 37.05 & 49.86  \\
			GN  &  37.55 & 59.36 & 40.87 & 21.99 & 40.49 & 48.43  &  34.06 & 55.97 & 35.76 & 15.58 & 36.86 & 49.61  \\
			XBNBlock$_{GN}$ & \textbf{38.19} & \textbf{60.09} & \textbf{41.65} & \textbf{22.48} & \textbf{41.50} & \textbf{48.73}  &  \textbf{34.57} & \textbf{56.79} & \textbf{36.58} & \textbf{16.14} & \textbf{37.49} & \textbf{50.58}  \\
			%	BN  &  --  & -- & -- & -- & -- & --  \\
			\toprule[1pt]
			\multicolumn{12}{c}{4conv1fc head box}  \\
			\hline
			Method     & AP$^{bbox}$   & AP$^{bbox}_{50}$ & AP$^{bbox}_{75}$  & AP$^{bbox}_{s}$ & AP$^{bbox}_{m}$ & AP$^{bbox}_{l}$ & AP$^{mask}$ & AP$^{mask}_{50}$ & AP$^{mask}_{75}$  & AP$^{mask}_{s}$ & AP$^{mask}_{m}$ & AP$^{mask}_{l}$\\
			\hline
			$BN^{\dag}$  &  37.51 & 58.26 & 40.65 & 21.88 & 40.72 & 48.00  &  33.68 & 54.90 & 35.62 & 15.44 & 36.60 & 48.88  \\
			GN  &  39.02 & 59.87 & 42.71 & 23.29 & 42.14 & 50.48  &  34.37 & 56.56 & 36.47 & 15.86 & 37.16 & 50.17  \\
			XBNBlock$_{GN}$ & \textbf{39.57} & \textbf{60.64} & \textbf{42.93} & \textbf{24.15} & \textbf{42.32} & \textbf{51.12}  &  \textbf{34.86} & \textbf{57.06} & \textbf{36.95} & \textbf{17.12} & \textbf{37.41} & \textbf{50.65}  \\
			%	BN  &  --  & -- & -- & -- & -- & --  \\
			\toprule[1pt]		
		\end{tabular}
		\caption{Detection and segmentation results ($\%$) on COCO using the Mask R-CNN framework implemented in~\cite{massa2018mrcnn}.  We use ResNet-50 as the backbone, combined with FPN. All models are trained by 1x lr scheduling (90k iterations), with a batch size of 16 on eight GPUs.}
		\vspace{-0.1in}
		\label{sup-table:Mask-R50}
		%	\vspace{0.1in}
	\end{scriptsize}
	\vspace{-0.1in}
\end{table*}

\begin{table*}[t]
	\centering
	\begin{scriptsize}
		\begin{tabular}{c|llllll|llllll}
			\bottomrule[1pt]
			\multicolumn{12}{c}{2fc head box}  \\
			\hline
			Method     & AP$^{bbox}$   & AP$^{bbox}_{50}$ & AP$^{bbox}_{75}$  & AP$^{bbox}_{s}$ & AP$^{bbox}_{m}$ & AP$^{bbox}_{l}$ & AP$^{mask}$ & AP$^{mask}_{50}$ & AP$^{mask}_{75}$  & AP$^{mask}_{s}$ & AP$^{mask}_{m}$ & AP$^{mask}_{l}$\\
			\hline
			$BN^{\dag}$  &  42.13 & 63.98 & 46.35 & 24.94 & 45.98 & 54.87  &  37.78 & 60.37 & 40.34 & 17.74 & 40.69 & 55.43  \\
			GN  & 41.47 & 63.54 & 44.76 & 25.36 & 45.33 & 53.20  &  37.17 & 60.23 & 39.31 & 17.92 & 40.24 & 54.12  \\
			XBNBlock$_{GN}$ & \textbf{42.69} & \textbf{64.98} & \textbf{46.20} & \textbf{25.39} & \textbf{46.72} & \textbf{55.03}  &  \textbf{38.00} & \textbf{61.03} & \textbf{40.39} & \textbf{17.99} & \textbf{40.95} & \textbf{55.52}  \\
			%	BN  &  --  & -- & -- & -- & -- & --  \\
			\toprule[1pt]
			\multicolumn{12}{c}{4conv1fc head box}  \\
			\hline
			Method     & AP$^{bbox}$   & AP$^{bbox}_{50}$ & AP$^{bbox}_{75}$  & AP$^{bbox}_{s}$ & AP$^{bbox}_{m}$ & AP$^{bbox}_{l}$ & AP$^{mask}$ & AP$^{mask}_{50}$ & AP$^{mask}_{75}$  & AP$^{mask}_{s}$ & AP$^{mask}_{m}$ & AP$^{mask}_{l}$\\
			\hline
			$BN^{\dag}$  &  42.18 & 63.22 & 46.00 & 25.01 & 45.60 & 54.90  &  37.53 & 60.18 & 39.99 & 17.80 & 40.49 & 55.04  \\
			GN  & 42.24 & 63.00 & 46.19 & 25.27 & 45.76 & 54.94  &  37.53 & 59.82 & 39.96 & 18.00 & 40.42 & 54.52  \\
			XBNBlock$_{GN}$ & \textbf{43.43} & \textbf{64.56} & \textbf{47.51} & \textbf{25.89} & \textbf{46.65} & \textbf{56.65}  &  \textbf{38.68} & \textbf{61.62} & \textbf{41.34} & \textbf{18.52} & \textbf{41.60} & \textbf{56.68}  \\
			\toprule[1pt]		
		\end{tabular}
		\caption{Detection and segmentation results ($\%$) on COCO using the Mask R-CNN framework and using ResNeXt-101 as the backbone, combined with FPN. All models are trained by 1x lr scheduling (180k iterations), with a batch size of 8 on eight GPUs.}
		\vspace{-0.1in}
		\label{sup-table:Mask-X101}
		%	\vspace{0.1in}
	\end{scriptsize}
	\vspace{-0.1in}
\end{table*}

\end{document}